%% file: 2017_TSP_L0TV_arxiv.tex
\documentclass[letterpaper]{article}
\usepackage[margin=1in]{geometry}
\usepackage[utf8]{inputenc}
\usepackage[cmex10]{amsmath}
\usepackage{amssymb}
\usepackage[hang,small]{caption}
\usepackage{graphicx}
\usepackage{color}
\usepackage{subfig}
\usepackage{theorem}
\usepackage{bbold}
\usepackage{epstopdf}
\usepackage{rotating}
\usepackage{url}
\usepackage{multirow}

\newtheorem{proposition}{Proposition}[section]
\newtheorem{lemma}{Lemma}[section]
\newtheorem{remark}{Remark}

\newtheorem{problem}{Problem}[section]
\usepackage{hhline}
\input notations.tex

\newcommand{\RR}{\mathbb{R}}

\newcommand{\Argmin}{\ensuremath{\mathrm{Argmin}}}

\usepackage{algorithmic}
\usepackage{algorithm}
\usepackage{color}
\usepackage{cite}

\newcommand{\xtrue}{\overline{x}}
\newcommand{\xvar}{x}

\title{Bayesian selection for the $\ell_2$-Potts model regularization parameter:\\$1$D piecewise constant signal denoising}
\author{Jordan~Frecon\thanks{J. Frecon, N. Pustelnik (Corresponding author) and P. Abry are with Univ Lyon, Ens de Lyon, Univ Claude Bernard, CNRS, Laboratoire de Physique, F-69342 Lyon, France (email: firstname.lastname@ens-lyon.fr).}, Nelly~Pustelnik$^*$, \\Nicolas~Dobigeon\thanks{N. Dobigeon and H. Wendt are with IRIT/INP-ENSEEIHT, University of Toulouse, France (email: firstname.lastname@irit.fr).}, Herwig~Wendt$^\dag$, and Patrice~Abry$^*$}
\date{}

\begin{document}
\maketitle

\begin{abstract}
Piecewise constant denoising can be solved either by deterministic optimization approaches, based on  the Potts model, or by stochastic Bayesian procedures. The former lead to low computational time but require the selection of a regularization parameter, whose value significantly impacts the achieved solution, and whose automated selection remains an involved and challenging problem.
Conversely, fully Bayesian formalisms encapsulate the regularization parameter selection into hierarchical models, at the price of high computational costs.
This contribution proposes an operational strategy that combines hierarchical Bayesian and Potts model formulations, with the double aim of automatically tuning the regularization parameter and of maintaining computational efficiency.
The proposed procedure relies on formally connecting a Bayesian framework to a $\ell_2$-Potts functional.
Behaviors and performance for the proposed piecewise constant denoising and regularization parameter tuning techniques are studied qualitatively and assessed quantitatively, and shown to compare favorably against those of a fully Bayesian hierarchical procedure, both in accuracy and in computational load.
\end{abstract}

\clearpage
\section{Introduction}

\noindent {\bf Piecewise constant denoising.}  Piecewise constant denoising (tightly related to change-point detection) is of considerable potential interest in numerous signal processing applications including, e.g., econometrics and biomedical analysis (see \cite{Basseville_M_1993_book_det_act,Little_M_2011_p-r-soc-a_gen_msn}, for an overview and \cite{Sowa_Y_2005_j-nat_dir_osr} for an interesting application in biology).
An archetypal and most encountered in the literature formulation considers noisy observations as resulting from the additive mixture of  a piecewise constant signal $\boldsymbol{\xtrue}\in\mathbb{R}^N$ with a Gaussian noise $\boldsymbol \epsilon \in \mathcal{N}(0,\sigma^2 \boldsymbol{I}_N)$
\begin{equation}
\label{eq:obs}
\boldsymbol{y} = \boldsymbol{\xtrue} + {\boldsymbol \epsilon}.
\end{equation}
Detecting change-points or denoising the piecewise constant information has been addressed by several strategies, such as Cusum procedures \cite{Basseville_M_1993_book_det_act}, {hierarchical Bayesian inference frameworks} \cite{Lavielle2001,Dobigeon_N_2007_j-ieee-tsp_joi_sma}, or functional optimization formulations, involving either the $\ell_1$-norm \cite{Rudin_L_1992_j-phys-d_tv_atvmaopiip,Kim2009,Dumbgen_L_2009_j-ejs_ext_sts,Little2010,Selesnick2010} or the $\ell_0$-pseudo-norm of the first differences of the signal \cite{Geman_S_1984_j-ieee-tpami_sto_rgd, Yao_Y_1984_j-ann-stat_est_nst, Mumford_D_1989_j-comm-pure-appl-math_optimal_aps,Winkler_G_2002_sds}.

This latter class frames the present contribution. Formally, it amounts to recovering the solution of a $\ell_2$-Potts model, namely,
\begin{equation}
\widehat{\boldsymbol{\xvar}}_\lambda = \arg\underset{\boldsymbol{\xvar} \in\mathbb{R}^N}{\min} \frac{1}{2}\|\boldsymbol{y}-\boldsymbol{\xvar}\|_2^2 + \lambda \Vert \bsL\boldsymbol{\xvar}\Vert_0,
\label{eq:minl0}
\end{equation}
where $\bsL\in \mathbb{R}^{(N-1)\times N}$ models the first difference operator, i.e., $\bsL\boldsymbol{\xvar}= (x_{i+1}-x_{i})_{1\leq i \leq N-1}$, the $\ell_0$-pseudo-norm counts the non-zero elements in $\bsL\boldsymbol{\xvar}$, and $\lambda >0$ denotes the regularization parameter which adjusts the respective contributions of the data-fitting and penalization terms in the
objective function.

Such a formulation however suffers from a major limitation: its actual solution depends drastically on the regularization parameter $\lambda$. The challenging question of automatically estimating $\lambda$ from data constitutes the core issue addressed in the present contribution.\\

\noindent {\bf Related works.}
Bayesian hierarchical inference frameworks have received considerable interests for addressing change-point or piecewise denoising problems\cite{Lavielle2001,Dobigeon_N_2007_j-ieee-tsp_joi_sma}.
This mostly results from their ability to include the hyperparameters within the Bayesian modeling and to jointly estimate them with the parameters of interest. In return for this intrinsic flexibility, approximating the Bayesian estimators associated with this hierarchical model generally requires the use of Markov chain Monte Carlo (MCMC) algorithms, which are often known as excessively demanding in terms of computational burden.

Remaining in the class of deterministic functional minimization, the non-convexity of the objective function underlying \eqref{eq:minl0} has sometimes been alleviated by a convex relaxation, i.e., the $\ell_0$-pseudo-norm is replaced by the $\ell_1$-norm
\begin{equation}
\widetilde{\boldsymbol{\xvar}}_\tau = \arg\underset{\boldsymbol{\xvar} \in\mathbb{R}^N}{\mathrm{min}} \frac{1}{2}\|\boldsymbol{y}-\boldsymbol{\xvar}\|_2^2 + \tau \Vert \bsL\boldsymbol{\xvar}\Vert_1.
\label{eq:minl1}
\end{equation}
In essence, such an approach preserves the same intuition of piecewise constant denoising, at the price of a shrinkage effect, but with the noticeable advantage of ensuring convexity of the resulting function to be minimized (see, e.g., \cite{Dumbgen_L_2009_j-ejs_ext_sts} for an intermediate solution where the penalization term is not convex but where the global criterion stays convex). This ensures the convergence of the minimization algorithms \cite{Vogel_C_1996_j-sc-comp_ite_mtv, Chambolle_A_2004_j-math-imaging-vis_alg_tvm, Barbero_A_2011_p-icml_fast_ntm, Bauschke_H_2011_book_con_amo} or straightforward computations \cite{Condat_L_2013_j-ieee-spl_dir_atv,Davies_P_2001_j-annals-statistics_le_rsm}.
The formulation proposed in \eqref{eq:minl1} has received considerable interest, because, besides the existence and performance of sound algorithmic resolution procedures, it can offer some convenient ways to handle the automated tuning of the regularization parameter  $\tau>0$. For instance, the \emph{Stein unbiased risk estimate} (SURE) \cite{Stein_C_1981_j-annals-statistics_est_mmn,Deledalle_C_2012_p-icip_unb_res} aims at producing an unbiased estimator that minimizes the mean squared error between $\boldsymbol{\xtrue}$ and $\widetilde{\boldsymbol{\xvar}}_\tau $.
While practically effective, implementing SURE requires the prior knowledge of the variance $\sigma^2$ of the residual error ${\boldsymbol \epsilon}$, often unavailable a priori (see, a contrario, \cite{Benazza_A_2005_j-ieee-tip_bui_twe,Pesquet_J_2009_j-ieee-tsp_sure_ads} for hyperspectral denoising or image deconvolution involving frames where $\sigma^2$ has been estimated).
In \cite{Selesnick_I_2015_j-ieee-spl_con_1tv}, the regularization parameter $\tau$ is selected according to an heuristic rule, namely  $\tau=0.25\sqrt{N}\sigma$, derived in \cite{Dumbgen_L_2009_j-ejs_ext_sts}.

Alternatively, again, the problem in \eqref{eq:minl1} can be tackled within a fully Bayesian framework, relying on the formulation of~\eqref{eq:minl1} as a statistical inference problem. Indeed, in the right-hand side of \eqref{eq:minl1}, the first term can be straightforwardly associated with a negative log-likelihood function by assuming an additive white Gaussian noise sequence ${\boldsymbol \epsilon}$, i.e., $\boldsymbol{y}\vert \boldsymbol{\xvar}$ is distributed according to the Gaussian distribution $\mathcal{N}(\boldsymbol{\xvar},\sigma^2 {\boldsymbol{I}_N})$. Further, the second term refers to a Laplace prior distribution for the first difference {$\bsL\boldsymbol{x}$} of the unobserved signal. Under such Bayesian modeling, the corresponding maximum a posteriori (MAP) criterion reads
\begin{equation}
\underset{\boldsymbol{\xvar} \in\mathbb{R}^N}{\textrm{{maximize}}}\left\{\!\left(\frac{1}{2\pi \sigma^2}\right)^{N/2}\! \!\!\!\!\!\!\!e^{-\frac{1}{2\sigma^2}\Vert \boldsymbol{y} -\boldsymbol{\xvar} \Vert_2^2} \frac{1}{Z(\tau/\sigma^2)}e^{-\frac{\tau}{\sigma^2} \Vert \bsL\boldsymbol{\xvar} \Vert_1}\!\right\}
\label{eq:minl1bayes}
\end{equation}
whose resolution leads to the solution \eqref{eq:minl1} and where $Z(\tau/\sigma^2)$ is the normalizing constant associated with the prior distribution. Following a hierarchical strategy, the hyperparameters $\tau$ and $\sigma^2$ could be included into the Bayesian model to be jointly estimated with $\boldsymbol{\xvar}$. However, in the specific case of~\eqref{eq:minl1}, the prior distribution related to the penalization is not separable with respect to (w.r.t.) the individual components of $\boldsymbol{\xvar}$:
The partition function $Z(\tau/\sigma^2)$ can hence not be expressed analytically.
Therefore, estimating $\tau$ within a hierarchical Bayesian framework would require either to choose a heuristic prior for $\tau$
as proposed in \cite{Oliveira_JP_2009_j-sp_atvidmma,Chaari_L_2011_p-ssp_parameter_ehw,Pereyra_M_2015_p-eusipco_mapeurp}
or to conduct intensive approximate Bayesian computations as in \cite{Pereyra2013}.\\

\noindent {\bf Goals, contributions and outline.}  Departing from the Bayesian interpretation of \eqref{eq:minl1}, the present contribution chooses to stick to the original $\ell_2$-Potts model~\eqref{eq:minl0}. Capitalizing on efficient dynamic programming algorithms \cite{Winkler_G_2002_sds,Winkler_G_2005_book_pstspm,Friedrich_F_2008_j-cgs_cpme,Storath_M_2014_j-ieee-tsp_jssrupf,Storath_M_2016_j-sc-comp_eal1tvrrvcvs} which allow $\widehat{\boldsymbol{\xvar}}_\lambda$ to be recovered for a predefined value of $\lambda$, the main objective of this work resides in the joint estimation of the denoised signal and the \emph{optimal} hyperparameter $\lambda$, without assuming any additional prior knowledge regarding the residual variance $\sigma^2$. Formally, this problem can be formulated as an extended counterpart of \eqref{eq:minl0}, i.e.,  a minimization procedure involving $\boldsymbol{\xvar}$, $\lambda$, and $\sigma^2$ as stated in what follows.
\begin{problem}
\label{pb:paramx}Let $\boldsymbol{y}\in \RR^N$ and $\phi\colon\RR_+\times \RR_+ \to \RR$. The problem under consideration is\footnote{Note that \eqref{eq:minl0lambda} could have been normalized differently without changing the minimization problem. A usual formulation aims at multiplying all terms by $\sigma^2$. For our study, formulation (5) is adopted for convenience.}
\begin{equation}
\underset{\boldsymbol{\xvar} \in\mathbb{R}^N, \lambda>0, \sigma^2>0}{\mathrm{minimize}}\;\frac{1}{2\sigma^2}\|\boldsymbol{y}-\boldsymbol{\xvar}\|_2^2 + \frac{\lambda}{\sigma^2} \Vert \bsL\boldsymbol{\xvar}\Vert_0 + \phi(\lambda,\sigma^2).
\label{eq:minl0lambda}
\end{equation}
\end{problem}
The main challenge for handling Problem \ref{pb:paramx} lies in the design of an appropriate function $\phi$ that leads to a relevant penalization of the overall criterion w.r.t. the set of nuisance parameters $(\lambda,\sigma^2)$.
To that end, Section~\ref{s:pb} provides a natural parametrization of $\boldsymbol{\xvar}$ and a reformulation of Problem~\ref{pb:paramx}.
In Section~\ref{s:Bayesian_framework}, a closed-form expression of $\phi(\cdot,\cdot)$ (cf.~\eqref{eq:lambdap}) and an interpretation of $\lambda$  will be derived from a relevant hierarchical Bayesian inference framework. For this particular choice of the function $\phi(\cdot,\cdot)$, Section \ref {s:algo} proposes an efficient algorithmic strategy to approximate a solution of the Problem \ref{pb:paramx}.
In Section~\ref{s:results}, the relevance and performance of this procedure are qualitatively illustrated and quantitatively assessed, and shown to compare favorably against the Markov chain Monte Carlo algorithm resulting from the hierarchical Bayesian counterpart of \eqref{eq:minl0lambda}, both in terms of accuracy and in computational load.

\section{Problem parametrization}
\label{s:pb}

\label{s:paramxvar}

Following \cite{Lavielle_M_1998_j-ieee-tsp_opt_srp,Lavielle2001,Dobigeon_N_2007_j-ieee-tsp_joi_sma,Dobigeon2007b,Dobigeon2007c}, piecewise constant signals ${\boldsymbol{\xvar}}\in\mathbb{R}^N$ can be explicitly parametrized via change-point locations $\boldsymbol{r}$ and amplitudes of piecewise constant segments~$\boldsymbol{\mu}$. These reparametrizations are derived in Sections \ref{p:r} and \ref{p:mu}. They are in turn used in Section \ref{p:reformulation} to bring Problem \ref{pb:paramx} in a form more amenable for explicit connection to a hierarchical Bayesian model.

\subsection{Change-point location parametrization $\boldsymbol{r}$}
\label{p:r}
To locate the time instants of the change-points in the denoised signal $\boldsymbol{\xvar}$, an indicator vector $\boldsymbol{r}=\big(r_{i}\big)_{1\leq i \leq N} \in \{0,1\}^N$ is introduced as follows
\begin{equation}
r_{i} =\left\{
              \begin{array}{ll}
                1, & \hbox{if there is a change-point at time instant $i$,} \\
                0, & \hbox{otherwise.}
              \end{array}
            \right.
\end{equation}
By convention, $r_{i}=1$ indicates that ${\xvar}_{i}$ is the last sample belonging to the current segment, and thus that ${\xvar}_{i+1}$ belongs to the next segment. Moreover, stating $r_{N} = 1$ ensures that the number $K$ of segments is equal to the number of change-points, i.e., $K=\sum_{i=1}^{N} r_{i}$.

For each segment index $k\in\{1,\dots,K\}$, the set $\mathcal{R}_{k} \subset\{1,\ldots,N\}$ is used to denote the set of time indices associated with the $k$-th segment. In particular, it is worthy to note that $\mathcal{R}_{k}\cap\mathcal{R}_{k'}=\emptyset$ for $k\neq k'$ and $\cup_{k=1}^K\mathcal{R}_{k}=\{1,\cdots,N\}$.
Hereafter, the notation $K_{\boldsymbol{r}}$ will be adopted to emphasize the dependence of the number $K$ of segments on the indicator vector $\boldsymbol{r}$, i.e., $K = \Vert \boldsymbol{r} \Vert_0$.

\subsection{Segment amplitude parametrization $\boldsymbol{\mu}$}
\label{p:mu}
The amplitudes of each segment of the piecewise constant signal can be encoded by introducing the vector $\boldsymbol{\mu} = (\mu_k)_{1\leq k \leq K_{\boldsymbol{r}}}$ such that
\begin{equation}
(\forall k\in \{1,\ldots,K_{\boldsymbol{r}}\})(\forall i\in \mathcal{R}_k) \quad {\xvar}_i = \mu_k.
\end{equation}

\subsection{Reformulation of Problem~\ref{pb:paramx}}
\label{p:reformulation}
In place of $\boldsymbol{\xvar}$, the parameter vector $\boldsymbol{\theta}=\{\boldsymbol{r},\boldsymbol{\mu}\}$ will now be used to fully specify the piecewise constant signal $\boldsymbol{\xvar}$. An important issue intrinsic to the $\ell_2$-Potts model and thus to this formulation stems from the fact that the unknown parameter $\boldsymbol{\theta}$ belongs to the space $$\mathcal{S} = \{\{0,1\}^N\times\mathbb{R}^{K_{\boldsymbol{r}}} : K_r = \{1,...,N-1\}\}$$ whose dimension is a priori unknown, as it depends on the number $K_{\boldsymbol{r}}$ of change-points. Moreover, this parametrization leads to the following lemma.
\begin{lemma}
\label{pb:minthetlam}
Let $\boldsymbol{y}\in \RR^N$ and $\phi\colon\RR_+\times \RR_+ \to \RR$. Problem~\ref{pb:paramx} is equivalent to
\begin{equation}
\underset{\substack{\boldsymbol{\theta} =\{\boldsymbol{r},\boldsymbol{\mu}\}\in \mathcal{S}\\\lambda>0,\sigma^2>0}}{\mathrm{minimize}} \;\frac{1}{2\sigma^2}\sum_{k=1}^{K_{\boldsymbol{r}}}\sum_{i\in \mathcal{R}_k} (y_i - \mu_k)^2 + \frac{\lambda}{\sigma^2} (K_{\boldsymbol{r}}-1) + \phi(\lambda,\sigma^2)
\label{eq:minl02_ext}
\end{equation}
where  $(\mathcal{R}_k)_{1\leq k \leq K_{\boldsymbol{r}}}$ is related to $\boldsymbol{r}$ as indicated in Section~\ref{p:r}.
\end{lemma}
Indeed, the data fidelity term in the minimization Problem~\ref{pb:paramx}
can be equivalently written as
\begin{equation}
\Vert \boldsymbol{y} - \boldsymbol{\xvar}\Vert^2 = \sum_{k=1}^{K_{\boldsymbol{r}}}\sum_{i\in \mathcal{R}_k} (y_i - \xvar_i)^2 =\sum_{k=1}^{K_{\boldsymbol{r}}}\sum_{i\in \mathcal{R}_k} ( y_i - \mu_k )^2.
\label{eq:minl0Kr}
\end{equation}
Moreover, the penalization can be rewritten as
\begin{equation}
\Vert \bsL\boldsymbol{x}\Vert_0 = \Vert \boldsymbol{r} \Vert_0 - 1= K_{\boldsymbol{r}}-1.
\label{eq:l0Kr}
\end{equation}

Lemma~\ref{pb:minthetlam} implies that estimating the piecewise constant signal ${\boldsymbol \xvar}$ can be equivalently formulated as estimating the parameter vector $\boldsymbol{\theta}$.

\section{Bayesian derivation of $\phi$}
\label{s:Bayesian_framework}

Assisted by the reformulation of Problem~\ref{pb:paramx} and a hierarchical Bayesian framework detailed in Section \ref{ss:hbm}, a relevant penalization function $\phi$ will be derived in Section \ref{ssec:map}.

\subsection{Hierarchical Bayesian model}
\label{ss:hbm}

In~\cite{Lavielle2001,Dobigeon_N_2007_j-ieee-tsp_joi_sma}, the problem of detecting change-points in a stationary sequence has been addressed following a Bayesian inference procedure which aims at deriving the posterior distribution of the parameter vector $\boldsymbol{\theta}$ from the likelihood function associated with the observation model and the prior distributions chosen for the unknown parameters. In what follows, a similar approach is proposed to produce a hierarchical Bayesian model that can be tightly related to the Problem \ref{pb:paramx} under a joint MAP paradigm.

First, the noise samples $({\boldsymbol \epsilon}_i)_{1\leq i \leq N}$ are assumed to be independent and identically distributed (i.i.d.) zero mean Gaussian variables with common but unknown variance $\sigma^2$, i.e., ${\boldsymbol \epsilon} |\sigma^2 \sim \mathcal{N}\left(\boldsymbol{0}, \sigma^2 {\boldsymbol{I}_N}\right)$.
The resulting joint likelihood function of the observations $ \boldsymbol{y} $ given the piecewise constant model $\{\boldsymbol{r}, \boldsymbol{\mu}\}$ and the noise variance $\sigma^2$ reads
\begin{equation}
\label{eq:likelihood}
f\!\left(\boldsymbol{y}|\boldsymbol{r}, \boldsymbol{\mu},\sigma^2 \right) \!\!=\! \prod_{k=1}^{K_{\boldsymbol{r}}} \prod_{i\in \mathcal{R}_{k}}\!\!\frac{1}{\sqrt{2\pi\sigma^2}}\exp\left(-\frac{\left(\mu_{k}-y_i \right)^2}{2\sigma^2}\right)\!.
\end{equation}
%

{Then} to derive the posterior distribution, prior distributions are elected for the parameters $\boldsymbol{r}$ and $\boldsymbol{\mu}$, assumed to be a priori independent.
Following well-admitted choices such as those in \cite{Lavielle_M_1998_j-ieee-tsp_opt_srp,Lavielle2001,Punskaya2002,Dobigeon_N_2007_j-ieee-tsp_joi_sma,Dobigeon2007b}, the  components $r_i$ of the indicator vector $\boldsymbol{r} $ are assumed to be a priori {independent and identically distributed (i.i.d.)} according to a Bernoulli distribution of hyperparameter $p$
\begin{equation}
\label{eq:prior_ind}
f(\boldsymbol{r} | p )  = \prod_{i=1}^{N-1} p^{r_i}(1-p)^{1-r_i}
		 	= p^{\sum_{i=1}^{N-1} r_i}(1-p)^{(N-1 -\sum_{i=1}^{N-1} r_i)}
		 	= \Big(\frac{p}{1-p}\Big)^{(K_{\boldsymbol{r}}-1)}(1-p)^{(N-1)}.
\end{equation}
The prior independence between the indicator components $r_i$ ($i=1,\ldots,N-1$) implicitly assumes that the occurrence of a change at a given time index does not depend on the occurrence of change at any other time index. Moreover, the hyperparameter $p$ stands for the prior probability of occurrence of a change, which is assumed to be independent of the location. Obviously, for particular applications, alternative and more specific choices can be adopted relying, e.g., on hard constraints \cite{Kail_IEEE_Trans_SP_2012} or Markovian models \cite{Dobigeon_EUSIPCO_2006}, for instance to prevent solutions composed of too short segments. %

From a Bayesian perspective, a natural choice for $  f(\boldsymbol{\mu}\vert \boldsymbol{r})$ consists in electing independent conjugate Gaussian prior distributions $\mathcal{N}\left(\mu_0,\sigma_0^2\right)$ for the segment amplitudes $\mu_k$ ($k=1,\ldots,K_{\boldsymbol{r}}$), i.e.,
\begin{equation}\label{eq:prior_mu}
f(\boldsymbol{\mu}\vert \boldsymbol{r}) = \prod_{k=1}^{K_{\boldsymbol{r}}} \frac{1}{\sqrt{2\pi \sigma_0^2}} e^{\frac{-(\mu_k - \mu_0)^2}{2\sigma_0^2}}.
\end{equation}
Indeed, this set of conjugate priors ensures that the conditional posterior distributions of the segment amplitudes are still Gaussian distributions.

Moreover, within a hierarchical Bayesian paradigm, nuisance parameters, such as the noise variance, and other hyperparameters defining the prior distributions can be included within the model to be estimated jointly with $\boldsymbol{\theta}$ \cite{Lavielle2001,Dobigeon_N_2007_j-ieee-tsp_joi_sma}.
In particular, to account for the absence of prior knowledge on the noise variance ${\sigma}^2$, a non-informative Jeffreys prior is assigned to $\sigma^2$
\begin{equation}
\label{eq:prior_noisevar}
f\left(\sigma^2\right) \propto \frac{1}{\sigma^2}.
\end{equation}
Proposed in \cite{Jeffreys1946}, the use of this improper distribution has been widely advocated in the Bayesian literature for its invariance under reparametrization \cite[Chap. 3]{Robert2007} (see also \cite{Punskaya2002,Dobigeon2007b}).
Finally, as in \cite{Dobigeon_EUSIPCO_2006,Dobigeon_N_2007_j-ieee-tsp_joi_sma,Dobigeon2007b,Dobigeon2007c}, a conjugate Beta distribution $\mathcal{B}(\alpha_0, \alpha_1)$ is assigned to the unknown hyperparameter $p$, which is a natural choice to model a $(0, 1)$-constrained parameter
\begin{equation}
\label{eq:prior_proba}
f(p) = \frac{\Gamma\left(\alpha_0+\alpha_1\right)}{\Gamma\left(\alpha_0\right)\Gamma\left(\alpha_1\right)} p^{\alpha_1-1}(1-p)^{\alpha_0-1}.
\end{equation}
Note that a wide variety of distribution shapes can be obtained by tuning the two hyperparameters $\alpha_0$ and $\alpha_1$, while ensuring the parameter $p$ belongs to the set $(0,1)$, as required since $p$ stands for a probability \cite[App. A]{Robert2007}. In particular, when the hyperparameters are selected as $\alpha_0=\alpha_1=1$, the prior in \eqref{eq:prior_proba} reduces to the uniform distribution.

\subsection{Joint MAP criterion}
\label{ssec:map}
From the likelihood function~\eqref{eq:likelihood} and prior and hyper-prior distributions~\eqref{eq:prior_ind}--\eqref{eq:prior_proba} introduced above, the joint posterior distribution reads
\begin{equation}
\label{eq:posteriorDistribution}
f(\boldsymbol{\Theta}|\boldsymbol{y}) \propto\; f\left(\boldsymbol{y}|\boldsymbol{r}, \boldsymbol{\mu},\sigma^2 \right) f\left(\boldsymbol{\mu} | \boldsymbol{r}\right) f(\boldsymbol{r} | p ) f(p)  f\left(\sigma^2\right)
\end{equation}
with $\boldsymbol{\Theta} = \left\{\boldsymbol{r},\boldsymbol{\mu}, \sigma^2,p\right\}$. Deriving the Bayesian estimators, such as the minimum mean square error (MMSE) and MAP estimators associated with this posterior distribution is not straightforward, mainly due to the intrinsic combinatorial problem resulting from the dimension-varying parameter space $\{0,1\}^N\times\mathbb{R}^{K_{\boldsymbol{r}}}$. In particular, a MAP approach would consist in maximizing the joint posterior distribution \eqref{eq:posteriorDistribution}, which can be reformulated as the following minimization problem by taking the negative logarithm of \eqref{eq:posteriorDistribution}.
\begin{problem}
\label{p:pb2}
Let $\boldsymbol{y} = (y_i)_{1\leq i \leq N}\in \RR^N$ and let $\Vhyper=\left\{\alpha_0,\alpha_1,\sigma_0^2\right\}$ a set of hyperparameters. We aim to
\begin{align}
\label{eq:posteriorDistribution2}
\underset{\boldsymbol{\Theta} = \left\{\boldsymbol{r},\boldsymbol{\mu}, \sigma^2,p\right\}}{\mathrm{minimize}}\; &
 \frac{1}{2\sigma^2}\sum_{k=1}^{K_{\boldsymbol{r}}} \sum_{i\in \mathcal{R}_k} (  y_i - \mu_k )^2+ (K_{\boldsymbol{r}}-1) \Bigg(\log \left(\frac{1-p}{p}\right) + \frac{1}{2}\log(2\pi\sigma_0^2)\Bigg) \nonumber\\
& + \frac{N}{2}\log (2\pi\sigma^2)   - (N-1) \log (1-p)+ \log  \sigma^2- (\alpha_1 -1)\log p - (\alpha_0 -1)\log (1-p)\nonumber\\
& + \frac{1}{2\sigma_0^2}\sum_{k=1}^{K_{\boldsymbol{r}}}(\mu_k - \mu_0)^2
+  \frac{1}{2}\log(2\pi\sigma_0^2).
\end{align}
\end{problem}

Despite apparent differences in parametrization between Problem~\ref{pb:paramx} and Problem~\ref{p:pb2}, we prove hereafter that both are equivalent for specific choices of $\lambda$ and $\phi(\cdot,\cdot)$.
\begin{proposition}
\label{prop:equiv}
For $\sigma_0^2$ large enough, Problem~\ref{pb:paramx} with
\begin{equation}
\lambda = \sigma^2 \bigg(\log\left(\frac{1-p}{p}\right) + \frac{1}{2}\log(2\pi \sigma_0^2)\bigg)
\label{eq:parametrized_lambda}
\end{equation}
and
\begin{align}
\label{eq:lambdap}
\phi(\lambda,\sigma^2) =& \frac{N}{2}\log(2\pi\sigma^2) + \log(\sigma^2) - \frac{\lambda}{\sigma^2}(N+\alpha_0-2) + \frac{N + \alpha_0-1}{2}\log(2\pi \sigma_0^2) \nonumber\\
&+ (N+\alpha_0+\alpha_1-3)\log\left( 1 + \exp\Big(\frac{\lambda}{\sigma^2}- \frac{1}{2}\log(2\pi \sigma_0^2)\Big)\right)
\end{align}
matches Problem~\ref{p:pb2}.
\end{proposition}

\begin{figure}[h]\centering
\includegraphics[width=.4\linewidth]{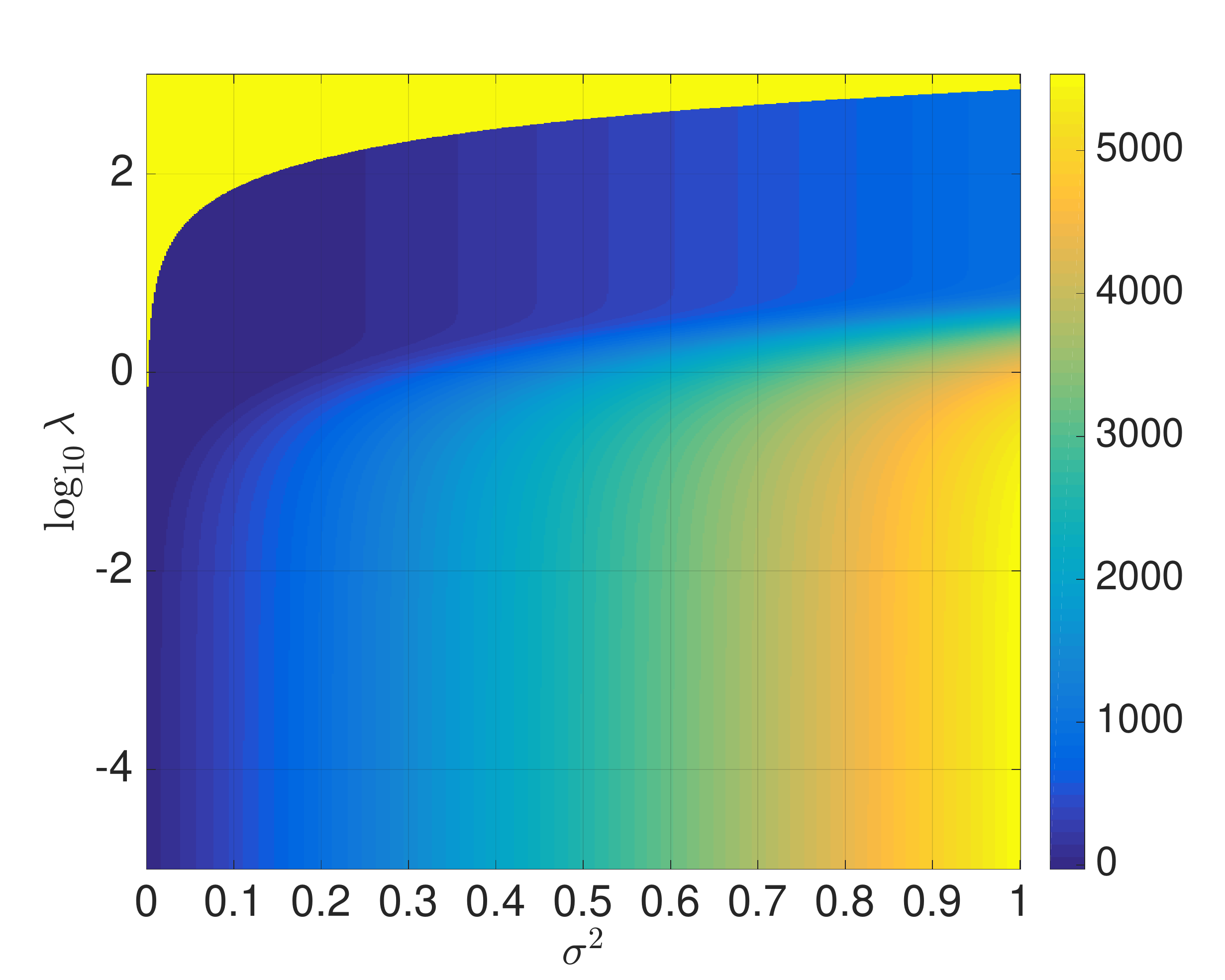}\;\;\includegraphics[width=.4\linewidth]{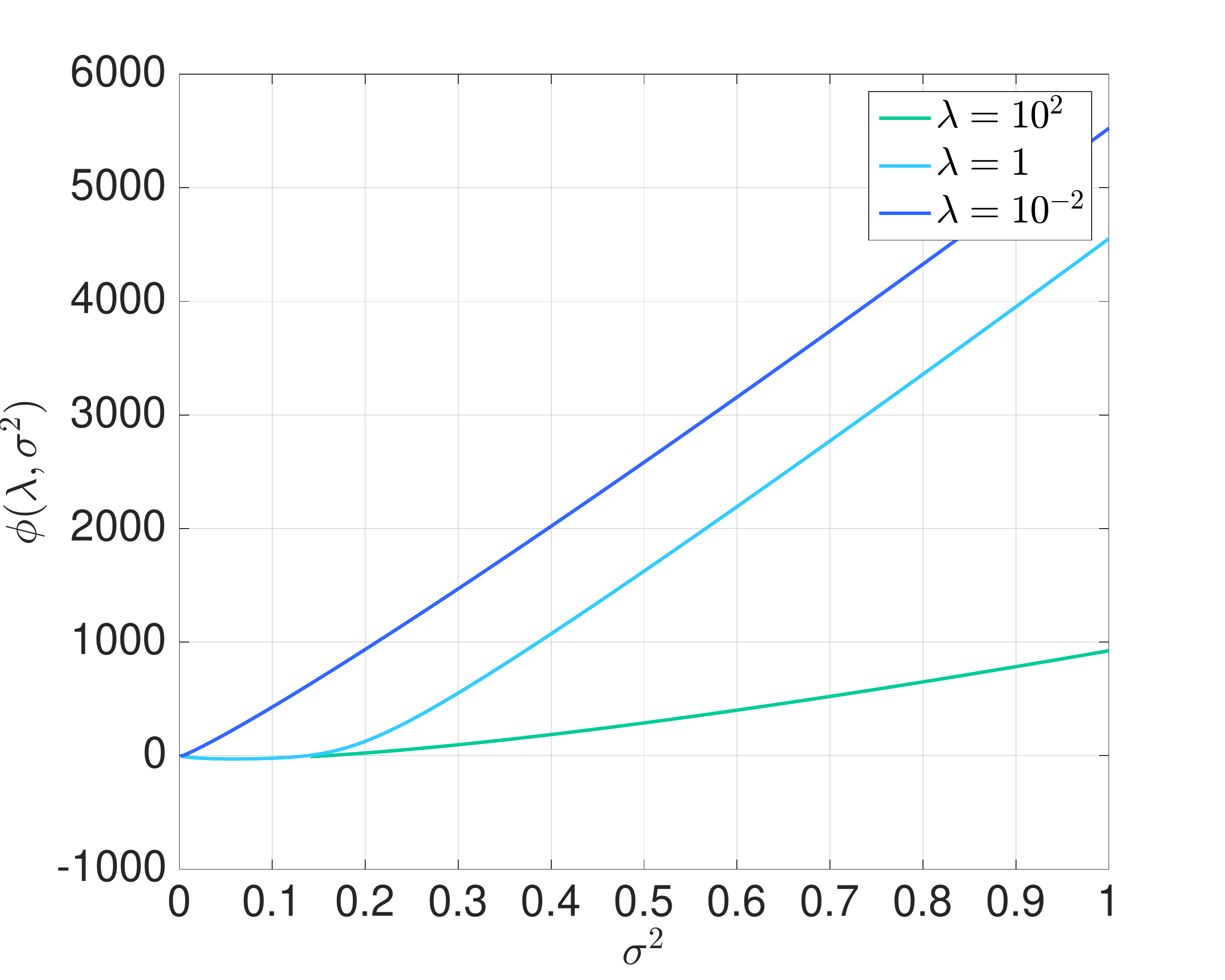}
\caption{\textbf{Illustration of $\phi(\lambda,\sigma^2)$} for the hyperparameter setting  $\alpha_0=\alpha_1=1$ and $2\pi\sigma_0^2 = 10^4$. }
\label{fig:displayphi}
\end{figure}

The sketch of the proof consists in identifying the three terms of the expression in \eqref{eq:minl02_ext} in the criterion~\eqref{eq:posteriorDistribution2}: the data fidelity term~\eqref{eq:minl0Kr}, a term proportional to the regularization~\eqref{eq:l0Kr}, and a third term $\phi(\lambda,\sigma^2)$ that is independent of the indicator vector $\boldsymbol{r}$.
Identification is possible under the condition that
the term $\frac{1}{2\sigma_0^2}\sum_{k=1}^{K_{\boldsymbol{r}}}(\mu_k - \mu_0)^2$ which explicitly depends on $\boldsymbol{r}$ through $K_{\boldsymbol{r}}$ can be neglected.
Thus, choosing $\sigma_0^2$ sufficiently large
\begin{equation}\label{eq:neglectsigma02}
\frac{1}{2\sigma_0^2}\sum_{k=1}^{K_{\boldsymbol{r}}}(\mu_k - \mu_0)^2 \ll \frac{1}{2}\log(2\pi\sigma_0^2),
\end{equation}
permits to equate Problem~\ref{pb:paramx} and Problem~\ref{p:pb2} with the choices $\lambda$ and $\phi(\lambda,\sigma^2)$  as defined in Proposition~\ref{prop:equiv}.
As an illustration of its complex shape, Fig.~\ref{fig:displayphi} represents $\phi(\lambda,\sigma^2)$ as function of  $\lambda$ and $\sigma^2$.

\begin{remark} The tuning of $\sigma_0^2$ requires some discussion:
\label{rem1}
\begin{itemize}
\item As  $\frac{1}{2\sigma_0^2}\sum_{k=1}^{K_{\boldsymbol{r}}}(\mu_k - \mu_0)^2$ is of the order of $ \frac{1}{2\sigma_0^2} {K_{\boldsymbol{r}}}\sigma_0^2 \approx \frac{pN}{2}$,
a sufficient condition for~\eqref{eq:neglectsigma02} to hold reads:
\begin{equation}\label{eq:sufficientCondNeglect}
\frac{pN}{2} \ll \frac{1}{2}\log(2\pi\sigma_0^2).
\end{equation}
\item However a careful examination of \eqref{eq:posteriorDistribution2} and \eqref{eq:parametrized_lambda} also leads to conclude that parameter $p$ actually controls $\lambda$ provided that $\log\left(\frac{1-p}{p}\right) $ is not totally neglectable when compared to $\frac{1}{2}\log(2\pi \sigma_0^2)$, thus implying an upper bound of the form:
\begin{equation}\label{eq:sufficientCondNeglect}
\frac{1}{2}\log(2\pi \sigma_0^2) \leq  \log \frac{(1-p)}{p} + \log \zeta
\end{equation}
where $\zeta>0$ is an arbitrary constant, whose magnitude will be precisely addressed in Section~\ref{ss:sigma0}.

The tuning of $\log(2\pi \sigma_0^2) $, of paramount practical importance, is hence not intricate and will be further discussed Section~\ref{ss:sigma0} from numerical experimentations.
\end{itemize}
\end{remark}

\section{Algorithmic solution}
\label{s:algo}

Thanks to Proposition~\ref{prop:equiv}, a function $\phi$ has been derived which allows the choice of the regularization parameter $\lambda$ to be penalized in Problem~\ref{pb:paramx}. In this section, an algorithmic solution is proposed to estimate $(\widehat{\boldsymbol{\xvar}}, \widehat{\lambda},\widehat{\sigma}^2)$, a solution of Problem~\ref{pb:paramx}.

An alternate minimization over $\boldsymbol{\xvar}$, $\lambda$ and $\sigma^2$ would not be efficient due to the non-convexity of the underlying criterion. To partly alleviate this problem, we propose to estimate $\lambda$ on a grid $\Lambda$. Therefore, a candidate solution can be obtained by solving $(\forall \lambda\in\Lambda)$
$$
(\widehat{\boldsymbol{x}}_{\lambda},\widehat{\sigma}^2_{\lambda}) \in\underset{\boldsymbol{x} \in\mathbb{R}^N,\sigma^2>0}{\Argmin}\;F(\boldsymbol{x},\lambda,\sigma^2)
$$
with
\begin{equation}
\label{eq:function_F}
F(\boldsymbol{x},\lambda,\sigma^2) = \frac{1}{2\sigma^2}\|\boldsymbol{y}-\boldsymbol{x}\|_2^2 + \frac{\lambda}{\sigma^2} \Vert L\boldsymbol{x}\Vert_0 + \phi(\lambda,\sigma^2).
\end{equation}
The minimization over $\boldsymbol{x}$ does not depend on $\sigma^2$, thus
\begin{equation}
\label{eq:l0TV_probAlgoApprox}
\widehat{\boldsymbol{x}}_\lambda = \arg\underset{\boldsymbol{x} \in\mathbb{R}^N}{\min}\;\frac{1}{2}\|\boldsymbol{y}-\boldsymbol{x}\|_2^2 + \lambda \Vert L\boldsymbol{x}\Vert_0,\\
\end{equation}
and we set
\begin{equation}
\label{eq:sigmaest}
\widehat{\sigma}^2_{\lambda} = \frac{\Vert \boldsymbol{y} - \widehat{\boldsymbol{x}}_\lambda\Vert^2}{N-1}.
\end{equation}
We finally select the triplet $(\widehat{\boldsymbol{x}}_{\widehat{\lambda}},\widehat{\lambda},\widehat{\sigma}^2_{\widehat{\lambda}})$ such that
\begin{equation}
\widehat{\lambda} = \arg\underset{\lambda\in\Lambda}{\min}\; F(\widehat{\boldsymbol{x}}_\lambda,\lambda,\widehat{\sigma}^2_\lambda).
\label{eq:l0TV_minl0lambdaP3}
\end{equation}
Note that the provided estimation amounts to using the solution of~\eqref{eq:minl0} for different $\lambda\in\Lambda$ to probe the space $(\boldsymbol{x},\sigma^2)\in\mathbb{R}^N\times \mathbb{R}_+$. Therefore, the iterations of the proposed {full algorithmic scheme} (reported in Algo.~\ref{algo:hybrid}) are very succinct and the overall algorithm complexity mainly depends on the ability to solve~\eqref{eq:minl0} efficiently for any $\lambda\in\Lambda$. In this work, we propose to resort to a dynamic programming algorithm developed in \cite{Winkler_G_2002_sds,Friedrich_F_2008_j-cgs_cpme} that allows \eqref{eq:l0TV_probAlgoApprox} to be solved exactly. We use its \emph{Pottslab} implementation~\cite{Storath_M_2014_j-ieee-tsp_jssrupf} augmented by a pruning strategy \cite{Storath_M_2014_j-siam-is_fast_pvv}.

\vspace{-0.3cm}

\textcolor{blue}{\begin{algorithm}[h!]
\small
\caption{Bayesian driven resolution of the $\ell_2$-Potts model\label{algo:hybrid}}
\begin{algorithmic}[1]
\REQUIRE Observed signal $\boldsymbol{y}\in\RR^N$.\\
$\quad\;$ The predefined set of regularization parameters $\Lambda$.\\
$\quad\;$ Hyperparameters $\Vhyper=\left\{\alpha_0,\alpha_1,\sigma_0^2\right\}$.\\
\vspace{0.1cm}
\hspace{-0.6cm}\textbf{Iterations:}
\FOR{$\lambda\in\Lambda$}
	\STATE Compute $\widehat{\boldsymbol{\xvar}}_{{\lambda}} =  \arg\underset{\boldsymbol{\xvar} \in\mathbb{R}^N}{\min} \frac{1}{2}\|\boldsymbol{y}-\boldsymbol{\xvar}\|_2^2 + \lambda \Vert \bsL\boldsymbol{\xvar}\Vert_0$.
	\STATE Compute $\widehat{\sigma}^2_\lambda = \Vert \boldsymbol{y} - \widehat{\boldsymbol{x}}_\lambda\Vert^2 / (N-1)$.
\ENDFOR
\vspace{0.1cm}
\ENSURE Solution $(\widehat{\boldsymbol{\xvar}}_{\widehat{\lambda}},\widehat{\lambda},\widehat{\sigma}^2_{\widehat{\lambda}})$ with $\widehat{\lambda} = \arg\underset{\lambda\in\Lambda}{\min}\; F(\widehat{x}_\lambda,\lambda,\widehat{\sigma}^2_\lambda)$
\end{algorithmic}
\end{algorithm}}

\vspace{-0.3cm}
\section{Automated selection of $\lambda$: Illustration and performance}
\label{s:results}
\subsection{Performance evaluation and hyperparameter settings}

\subsubsection{Synthetic data} The performance of the proposed automated selection of $\lambda$ are illustrated and assessed using Monte Carlo numerical simulation based on synthetic data $\boldsymbol{y}=\overline{\boldsymbol{x}}+\boldsymbol{\epsilon}$, where the noise $\boldsymbol{\epsilon} $ consists of i.i.d. samples drawn from $\boldsymbol{\epsilon}\sim\mathcal{N}(0,\sigma^2\boldsymbol{I}_N)$, and the signal $\overline{\boldsymbol{x}}$ is piecewise constant, with i.i.d. change-points, occurring with probability $p$, and i.i.d. amplitudes drawn from a uniform distribution (on the interval $[\overline{x}_{\min},\overline{x}_{\max}]$).

\subsubsection{Performance quantification} Performance are quantified by the relative mean square error (MSE) and the Jaccard error.
While the former evaluates performance in the overall (shape and amplitude) estimation $\widehat{\boldsymbol{x}}$ of $\overline{\boldsymbol{x}}$ such that
$$
\textrm{MSE}(\overline{\boldsymbol{x}},\widehat{\boldsymbol{x}}) = \frac{\Vert\overline{\boldsymbol{x}} -\widehat{\boldsymbol{x}} \Vert}{\Vert\overline{\boldsymbol{x}} \Vert},
$$
the latter focuses on the accuracy of change-point location estimation $\boldsymbol{r}$. The Jaccard error between the true change-point vector $\overline{\boldsymbol{r}}$ and its estimate $\widehat{\boldsymbol{r}}$ (both in $\{0,1\}^N$), is defined as \cite{Jaccard_P_1901_bsvsn_dfabdrv,Hamon_R_2016_jcn_rvansmcbs}
\begin{equation}
\small
J(\overline{\boldsymbol{r}},\widehat{\boldsymbol{r}}) = 1-\frac{\sum_{i=1}^N \min(\overline{r}_i, \widehat{r}_i)}{
\sum_{\substack{
   1 \leq i \leq N \\
   \overline{r}_i>0,
   \widehat{r}_i>0
  }} \frac{\overline{r}_i + \widehat{r}_i}{2} + \sum_{\substack{
   1 \leq i \leq N \\
   \widehat{r}_i=0
  }} \overline{r}_i
  +  \sum_{\substack{
   1 \leq i \leq N \\
   \overline{r}_i=0
  }} \widehat{r}_i
  }.
  \end{equation}
$J(\overline{\boldsymbol{r}},\widehat{\boldsymbol{r}}) $ ranges from $0$ when $\overline{\boldsymbol{r}} =\widehat{\boldsymbol{r}}$, to $1$, when $\overline{\boldsymbol{r}}\cap\widehat{\boldsymbol{r}}=\emptyset$. The Jaccard error is a demanding measure: when one half of non-zero values of  two given binary sequences coincides while the other half does not, then $J(\overline{\boldsymbol{r}},\widehat{\boldsymbol{r}}) =2/3$.

In the present study, to account for the fact that a solution with a change point position mismatch by a few time indices remains useful and of practical interest, the Jaccard error is computed between smoothed versions $\overline{\boldsymbol{r}} \ast \mathcal{G}$ and $\widehat{\boldsymbol{r}} \ast \mathcal{G}$ of the true and estimated sequences $\overline{\boldsymbol{r}}$ and $\widehat{\boldsymbol{r}}$.
The convolution kernel $\mathcal{G}$ is chosen here as a Gaussian filter (with a stansdard deviation of $0.5$) truncated to a $5$-sample support.

Performance are averaged over $50$ realizations, except for comparisons with the MCMC-approximated Bayesian estimators (see Section~\ref{ss:bayes}) where only $20$ realizations are used because of MCMC procedure's high computational cost.

\subsubsection{Hyperparameter setting}
The prior probability $p$ for change-point is chosen as a uniform distribution over $(0,1)$, obtained with hyperparameters  set to $\alpha_0=\alpha_1=1$. Indeed, for such a choice, the Beta distribution in~\eqref{eq:prior_proba} reduces to a uniform distribution, hence leading to a non-informative prior for the change-point probability.  This hyperparameter setting  leads to the following expression for the penalization function
\begin{equation}
\label{eq:lambdap}
\phi(\lambda,\sigma^2) = \log(\sigma^2)  + \frac{N}{2}\big(\log(2\pi\sigma^2) + \log(2\pi \sigma_0^2) \big) + (N-1)\Bigg(\log\left( 1 + \exp\Big(\frac{\lambda}{\sigma^2}- \frac{1}{2}\log(2\pi \sigma_0^2)\Big)\right)-\frac{\lambda}{\sigma^2}\Bigg).
\end{equation}%
Amplitudes $\boldsymbol{\mu}$ for $\overline{\boldsymbol{x}} $ are parametrized with $\sigma_0^2$, which according to Proposition~\ref{prop:equiv} should be chosen large enough.
For the time being, we set $2\pi\sigma_0^2 = 10^4$, and further discuss the impact of this choice in Section~\ref{ss:sigma0}.

\subsubsection{Discretization of $\Lambda$} For practical purposes, we make use of a discretized subset $\Lambda$ for $\lambda$ ($500$ values equally spaced, in a $\mathrm{log}_{10}$-scale, between $10^{-5}$ and $10^5$). Note that in the toolbox associated with this work, an option is called $\lambda-$shooting allowing to select the grid $\Lambda$ according to the strategy introduced in \cite{Friedrich_F_2008_j-cgs_cpme}.

\subsection{Illustration of the principle of the automated tuning of $\lambda$.}

\begin{figure*}[t]\centering
\begin{tabular}{ccc}
\includegraphics[width=.25\linewidth]{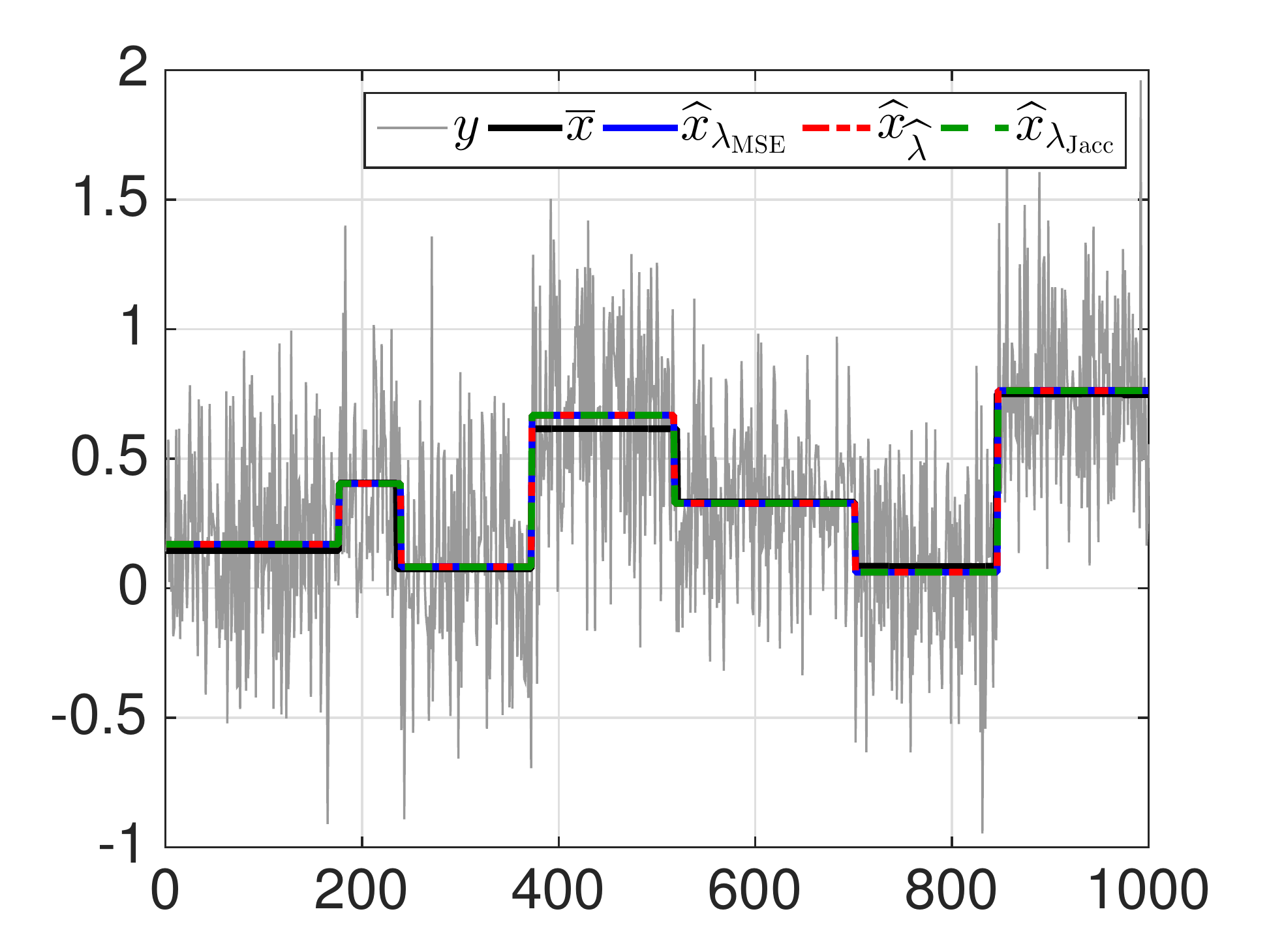}&
\includegraphics[width=.25\linewidth]{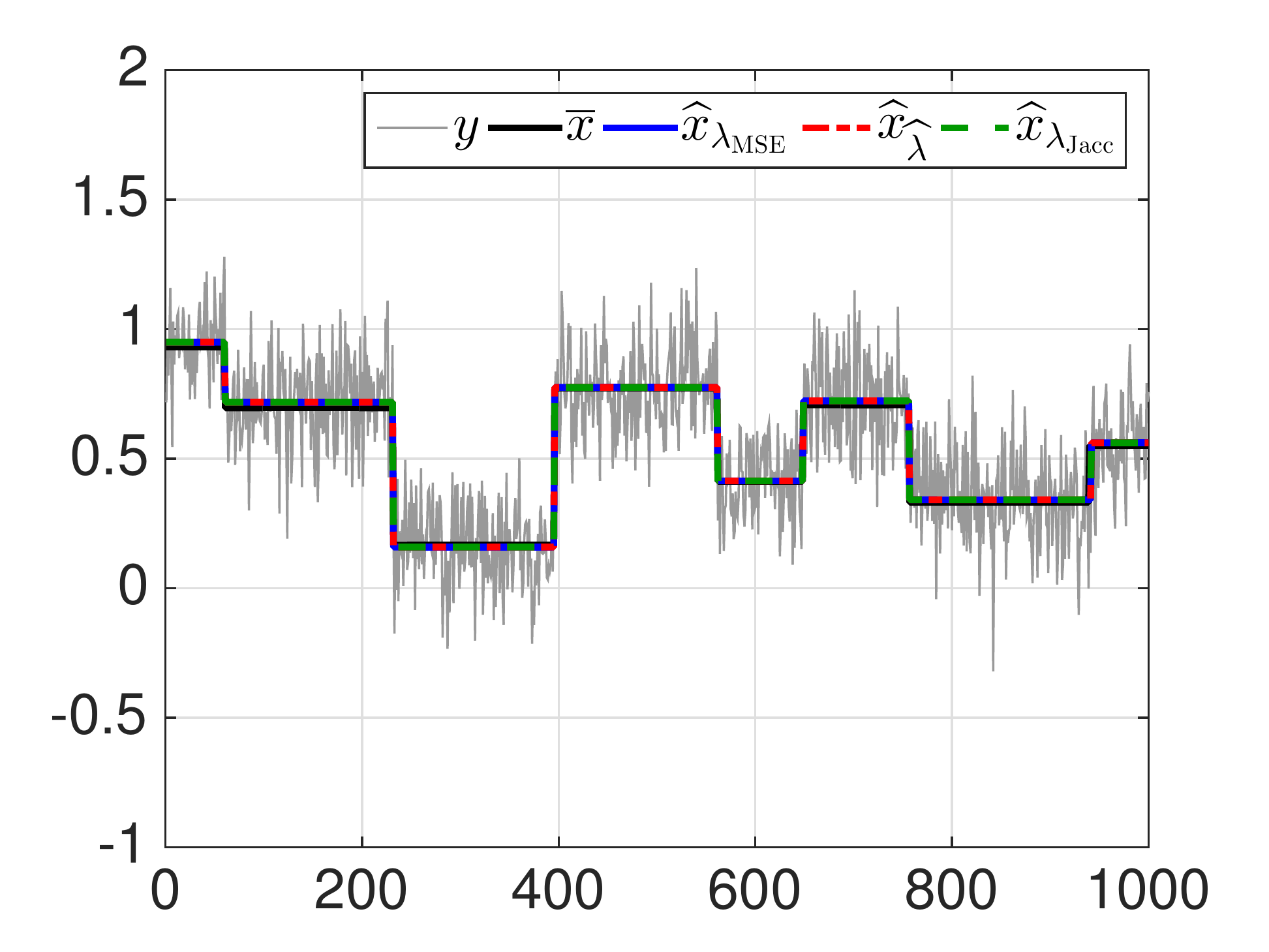}&
\includegraphics[width=.25\linewidth]{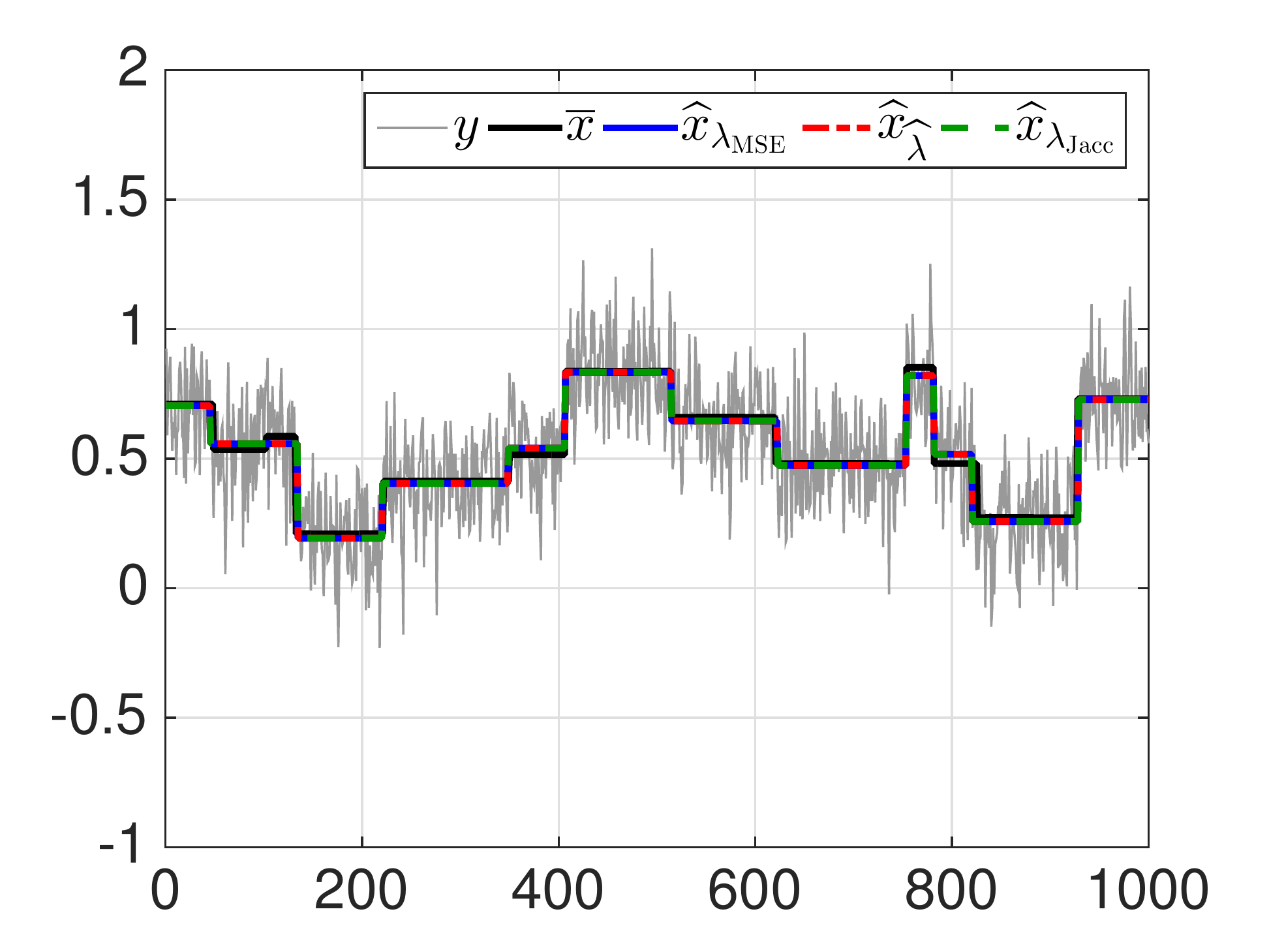}\\
\includegraphics[width=.25\linewidth,clip=true,trim=0cm 1cm 0cm 0cm]{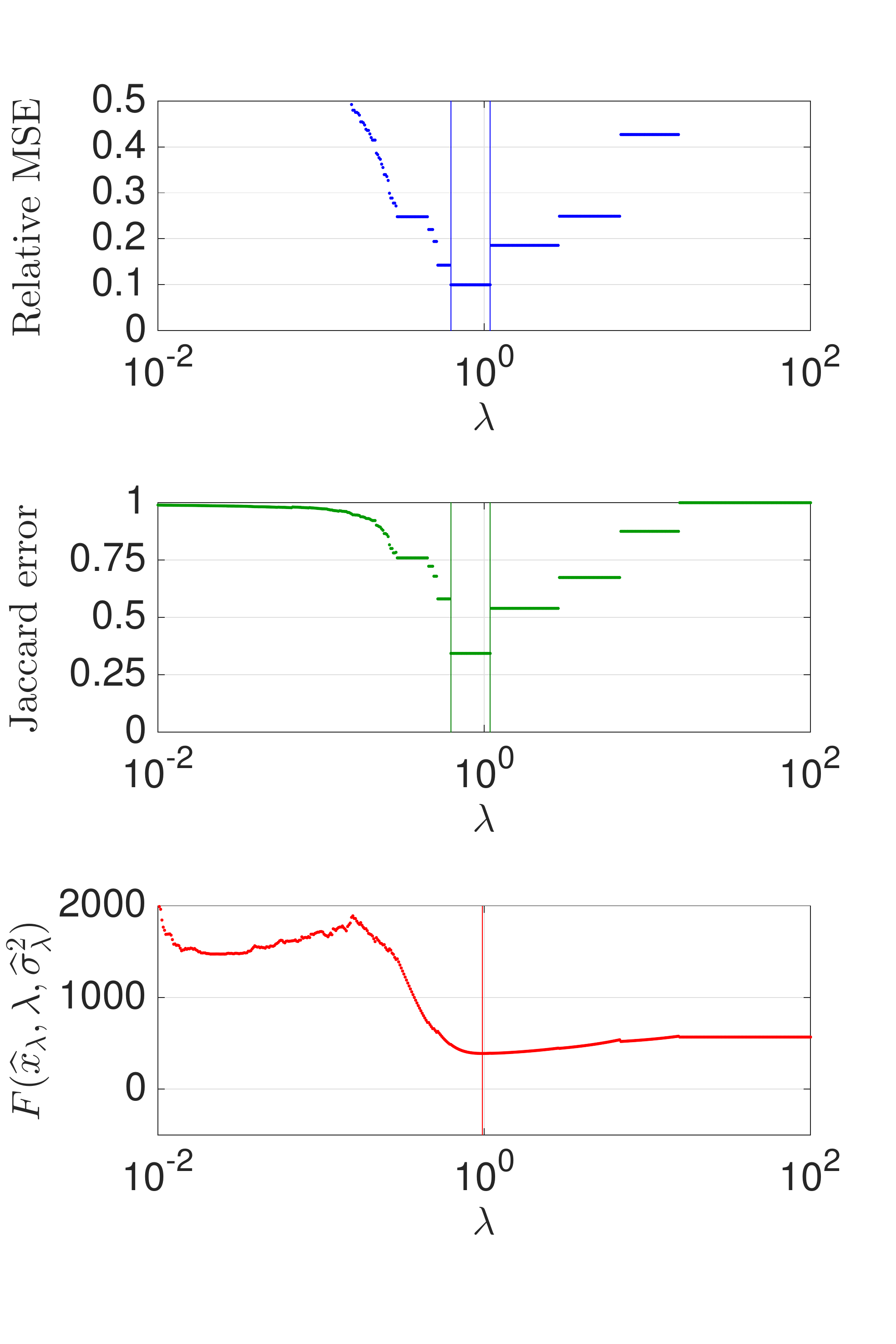}&
\includegraphics[width=.25\linewidth,clip=true,trim=0cm 1cm 0cm 0cm]{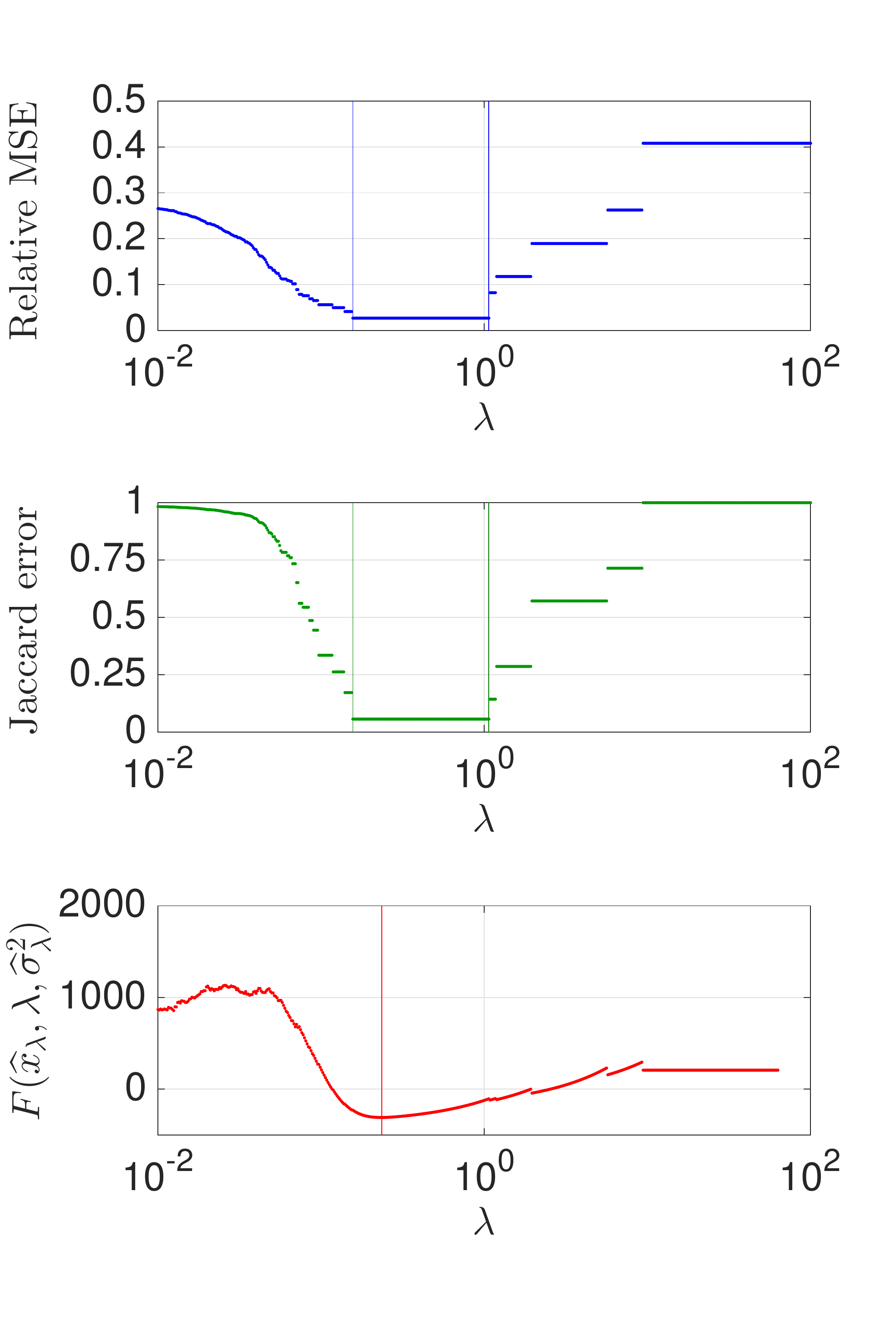}&
\includegraphics[width=.25\linewidth,clip=true,trim=0cm 1cm 0cm 0cm]{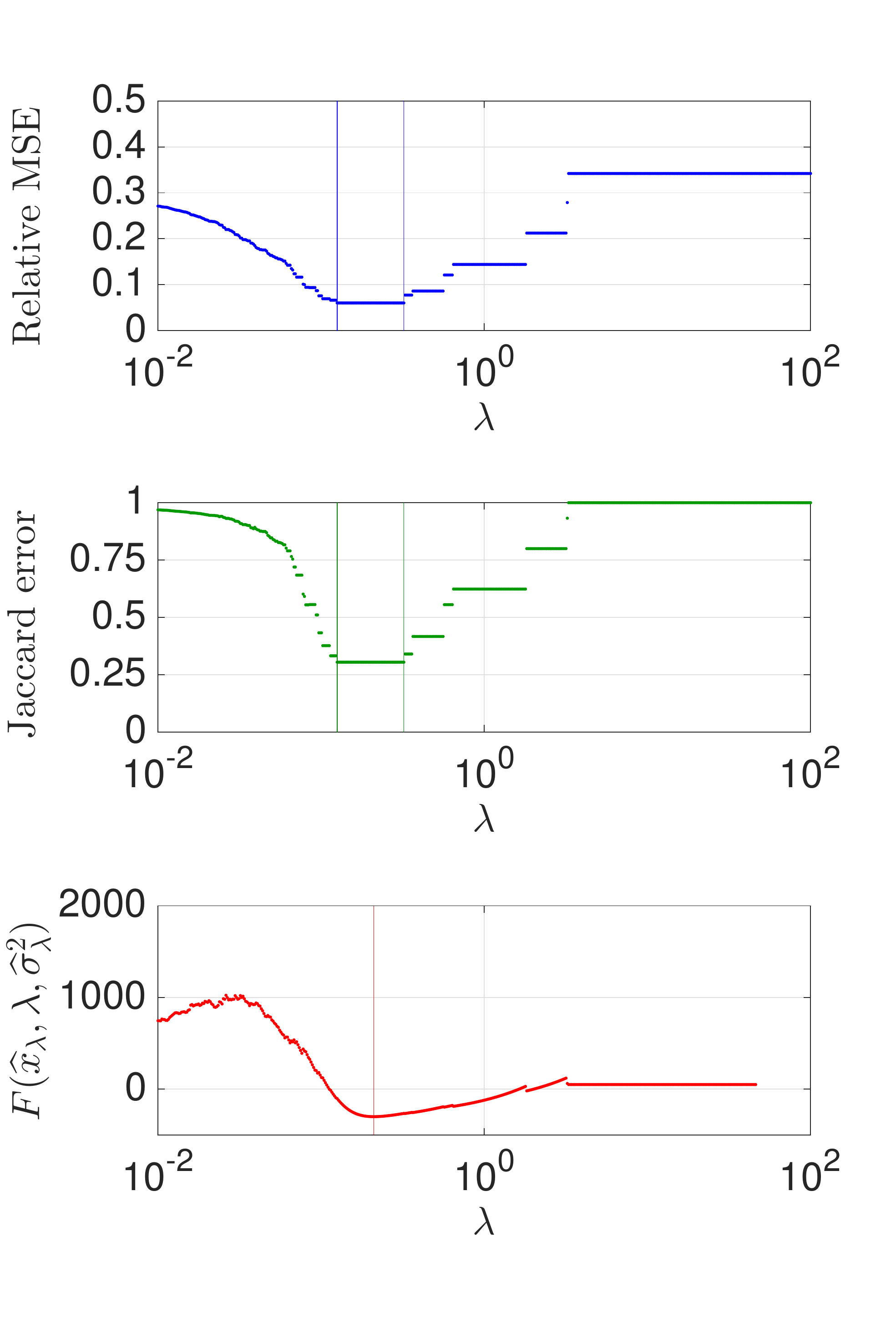}\\
(a)~${\rm ANR}=1$, $p=0.01$. & (b)~${\rm ANR}=2$, $p=0.01$. & (c)~${\rm ANR}=2$, $p=0.015$.
\end{tabular}
\vskip.3\baselineskip
\caption{\textbf{Illustration of the automated tuning of $\lambda$:} Top: Data $\boldsymbol{y}$ to which are superimposed true signal $\overline{\boldsymbol{x}} $ and oracle signals $\widehat{\boldsymbol{\xvar}}_{\lambda_{\rm{MSE}}}$ (blue) and $\widehat{\boldsymbol{\xvar}}_{\lambda_{\rm{Jac}}}$ (green) obtained for ${\lambda_{\rm{MSE}}}$ and $\lambda_{\rm{Jac}}$ minimizing the MSE and the Jaccard error, together with estimated $\widehat{\boldsymbol{\xvar}}_{{\widehat{\lambda}}}$ (red) obtained from automated selection of $\lambda$.
Second and third lines: relative MSE and Jaccard error as functions of $\lambda$.
Vertical lines locate ${\lambda_{\rm{MSE}}}$ and $\lambda_{\rm{Jac}}$.
Bottom line: Criterion $F$ (cf. \eqref{eq:function_F}) as a function of $\lambda$.
Automatically selected $\widehat{\lambda}$ is indicated by vertical red lines and is satisfactorily located in between the vertical lines indicating ${\lambda_{\rm{MSE}}}$ and $\lambda_{\rm{Jac}}$.
(left) $p = 0.01$ and ANR = 1, (middle)~$p = 0.01$ and ANR = 2, (right) $p = 0.15$ and ANR = 2. For all configurations, $\overline{x}_{\max} - \overline{x}_{\min}=1$.
\label{fig:choiceLambda}}
\end{figure*}

Fig.~\ref{fig:choiceLambda} illustrates the principle of the automated selection of $\lambda$, under various scenarios, with different values for $p$ and different \emph{amplitude-to-noise-ratios}\footnote{This measure allows amplitudes between successive segments to be compared w.r.t. to noise power. Since segment amplitudes are drawn uniformly between $\overline{x}_{\min}$ and $\overline{x}_{\max}$ the average difference between successive segments  is $\frac{\overline{x}_{\max} - \overline{x}_{\min}}{3}$.} (ANR) where ANR$=\frac{\overline{x}_{\max} - \overline{x}_{\min}}{3\sigma}$.

For all scenarios, Fig.~\ref{fig:choiceLambda} shows that the automatically selected $\widehat{\lambda}$, obtained as the minimum of the devised criterion  $F$ (cf. \eqref{eq:function_F}, vertical red line in bottom row), satisfactorily falls within the ranges of $\lambda$ achieving the  MSE minimum (denoted $\Lambda_{\rm MSE}$ and marked with vertical lines in the second row) or the minimum of Jaccard error (denoted $\Lambda_{\rm Jac}$ and marked with vertical lines in the third row):
$\widehat{\lambda}\in\Lambda_{\rm MSE}\cap\Lambda_{\rm Jac}$.
In addition, on the first row of Fig.~\ref{fig:choiceLambda}, the corresponding solution $\widehat{\boldsymbol{\xvar}}_{{\widehat{\lambda}}}$ (red) visually appears as a satisfactory estimator of $\overline{\boldsymbol{x}}$ (black), similar to the  ``oracle'' solutions $\widehat{\boldsymbol{\xvar}}_{\lambda_{\rm{MSE}}}$ (blue) and $\widehat{\boldsymbol{\xvar}}_{\lambda_{\rm{Jac}}}$ (green) that rely on a perfect knowledge of the noise-free signal $\overline{\boldsymbol{\xvar}}$.
Solution $\widehat{\boldsymbol{x}}_{\widehat{\lambda}}$ indeed systematically benefits from lower relative MSE and Jaccard error than $\widehat{\boldsymbol{x}}_\lambda$ for any other $\lambda$.
While, by construction, $\widehat{\boldsymbol{\xvar}}_{\lambda_{\rm{MSE}}}$ and $\widehat{\boldsymbol{\xvar}}_{\lambda_{\rm{Jac}}}$  are identical for all $\lambda$ within $\Lambda_{\rm MSE}$ or $\Lambda_{\rm Jac}$, the automated selection procedure for $\lambda$ yields interestingly a single global minimizer.

When $\rm ANR$ decreases, a closer inspection of Fig.~\ref{fig:choiceLambda}~(left column) further shows that the supports of oracle ${\lambda}$, $\Lambda_{\rm MSE}$ and $\Lambda_{\rm Jac}$ are drastically shrinking, yet the automated selection of $\lambda$ remains satisfactory even in these more difficult contexts. The same holds when $p$ increases (see Fig.~\ref{fig:choiceLambda}, right column).

\subsection{Estimation performance quantification}
To assess and quantify estimation performance of $\widehat{\lambda}$ as functions of data parameters $\sigma^2$,  $\overline{x}_{\max} - \overline{x}_{\min}$, and $p$, we have performed Monte Carlo simulations under various settings.

\begin{figure*}
\begin{tabular}{@{}c@{}c@{}cc@{}c@{}c@{}}
\includegraphics[width=.15\linewidth,clip=true,trim=0.2cm 0cm .5cm .5cm]{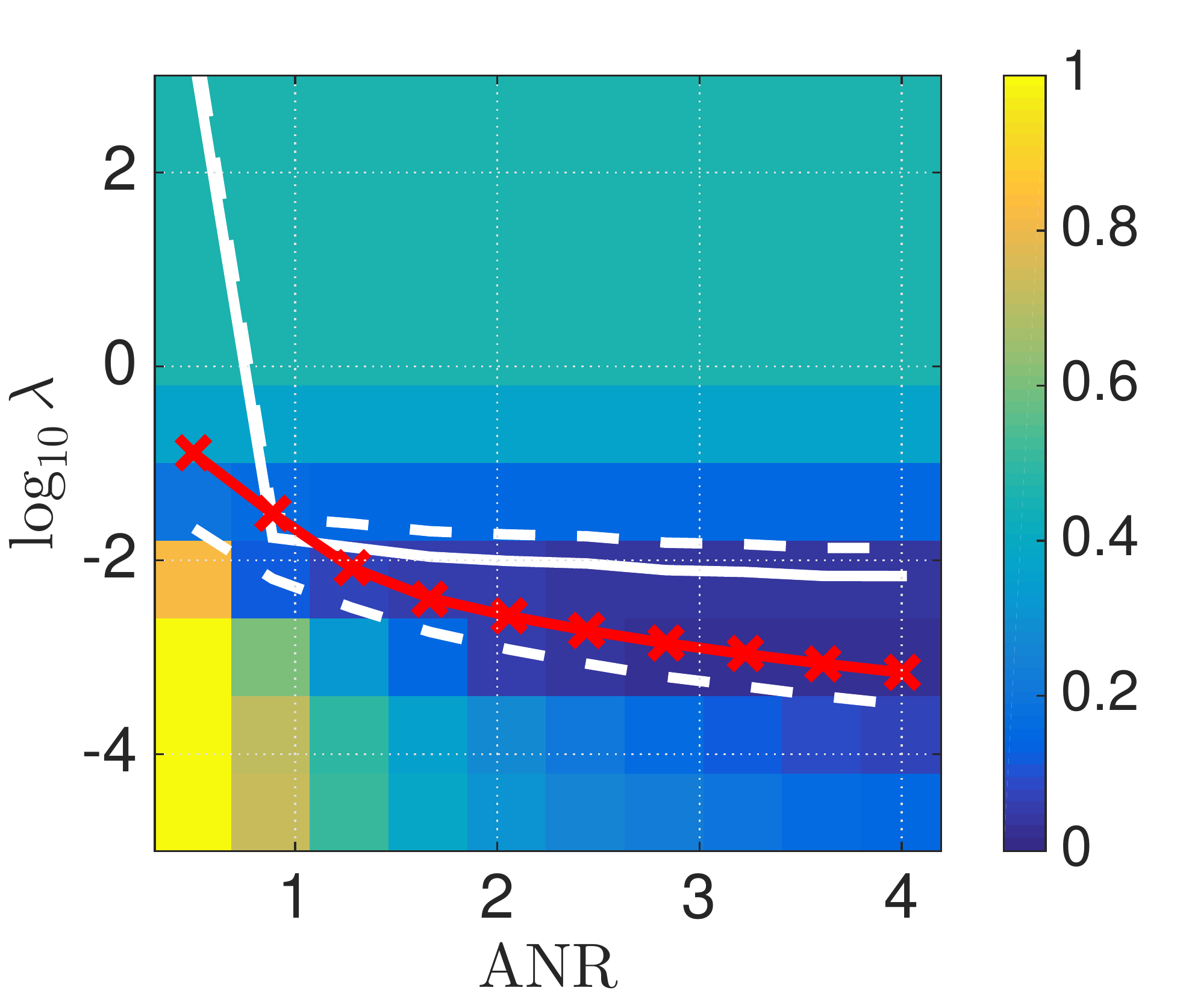}&
\includegraphics[width=.15\linewidth,clip=true,trim=0.2cm 0cm .5cm .5cm]{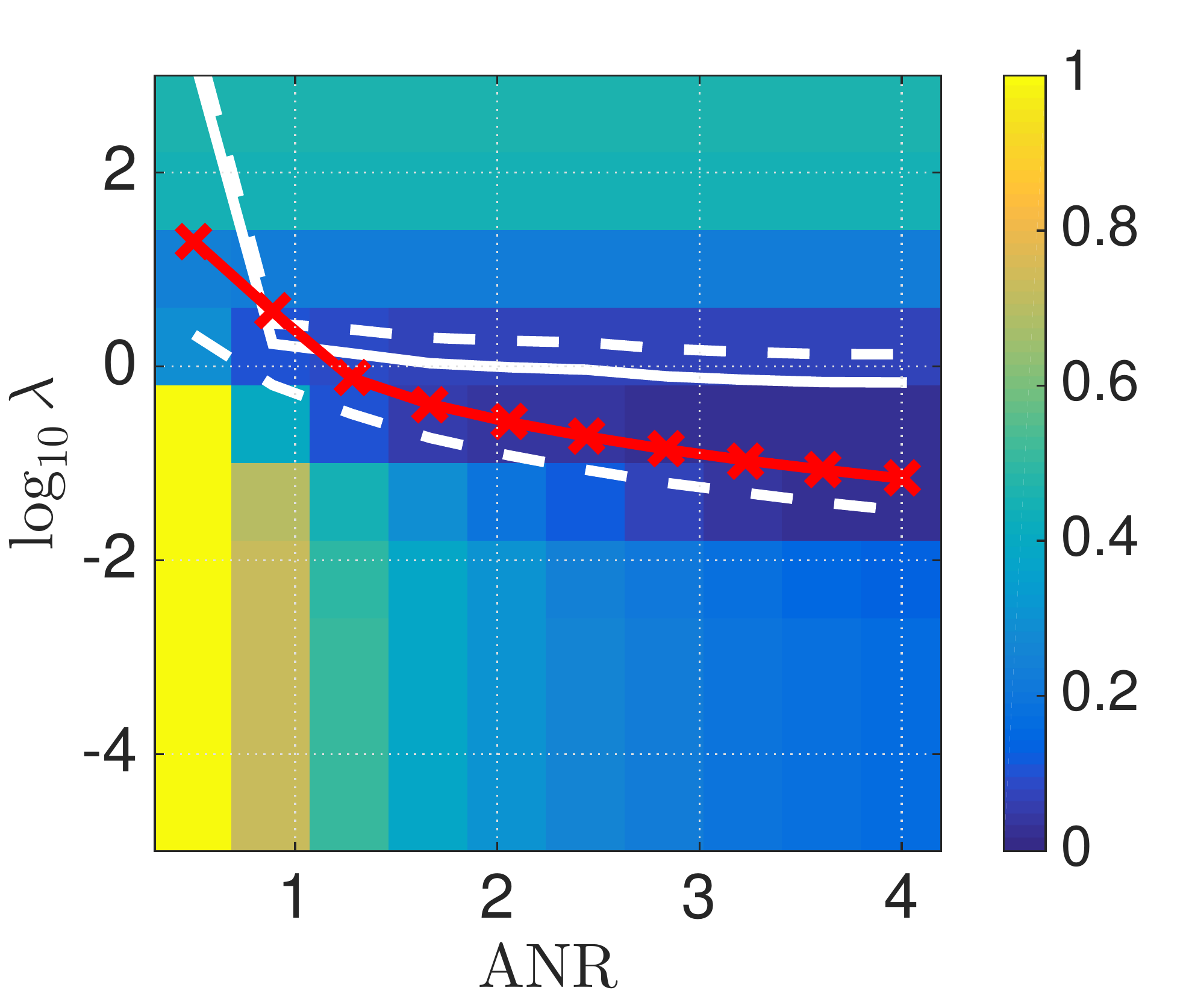}&
\includegraphics[width=.15\linewidth,clip=true,trim=0.2cm 0cm .5cm .5cm]{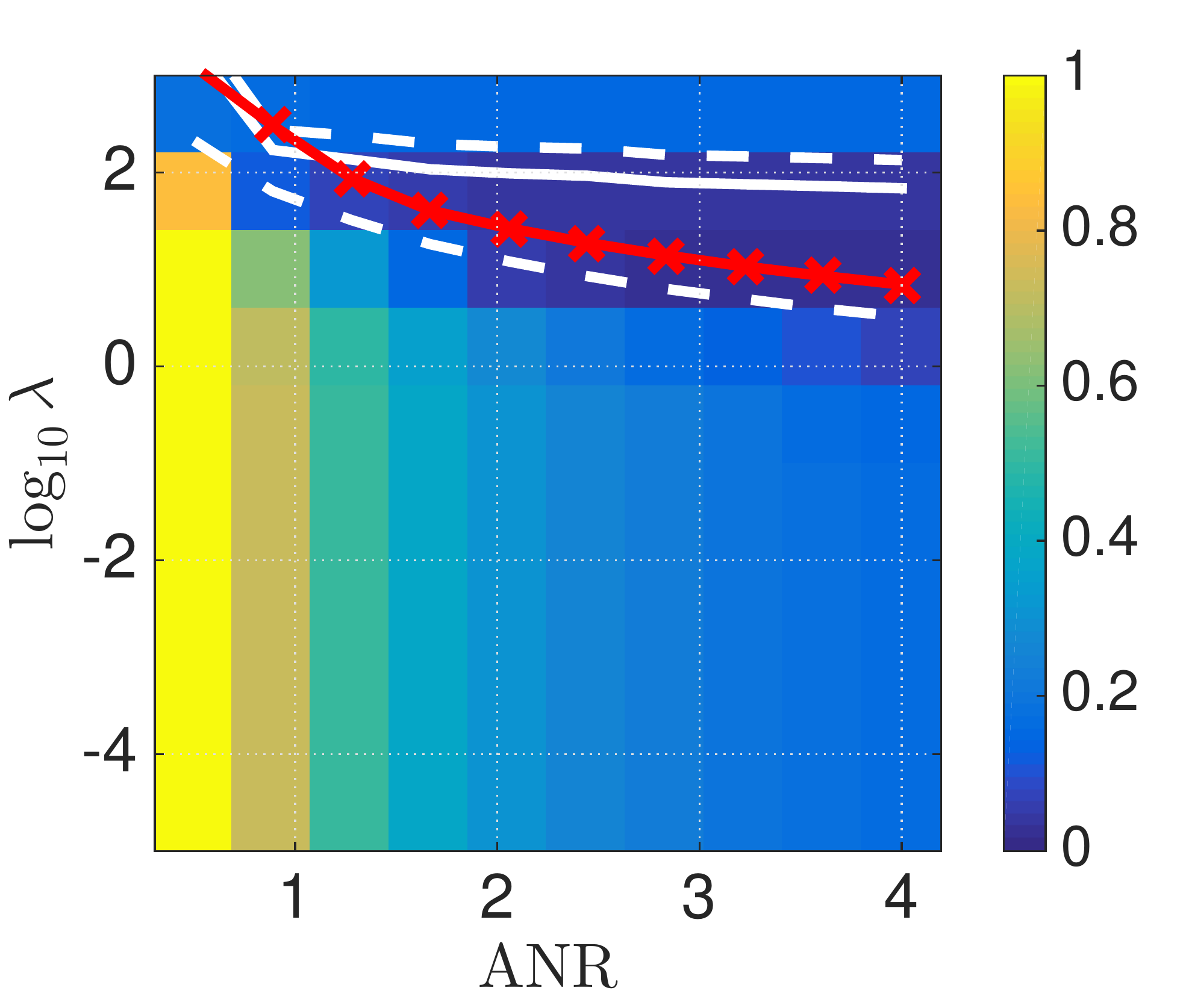}&
\includegraphics[width=.15\linewidth,clip=true,trim=0.2cm 0cm .5cm .5cm]{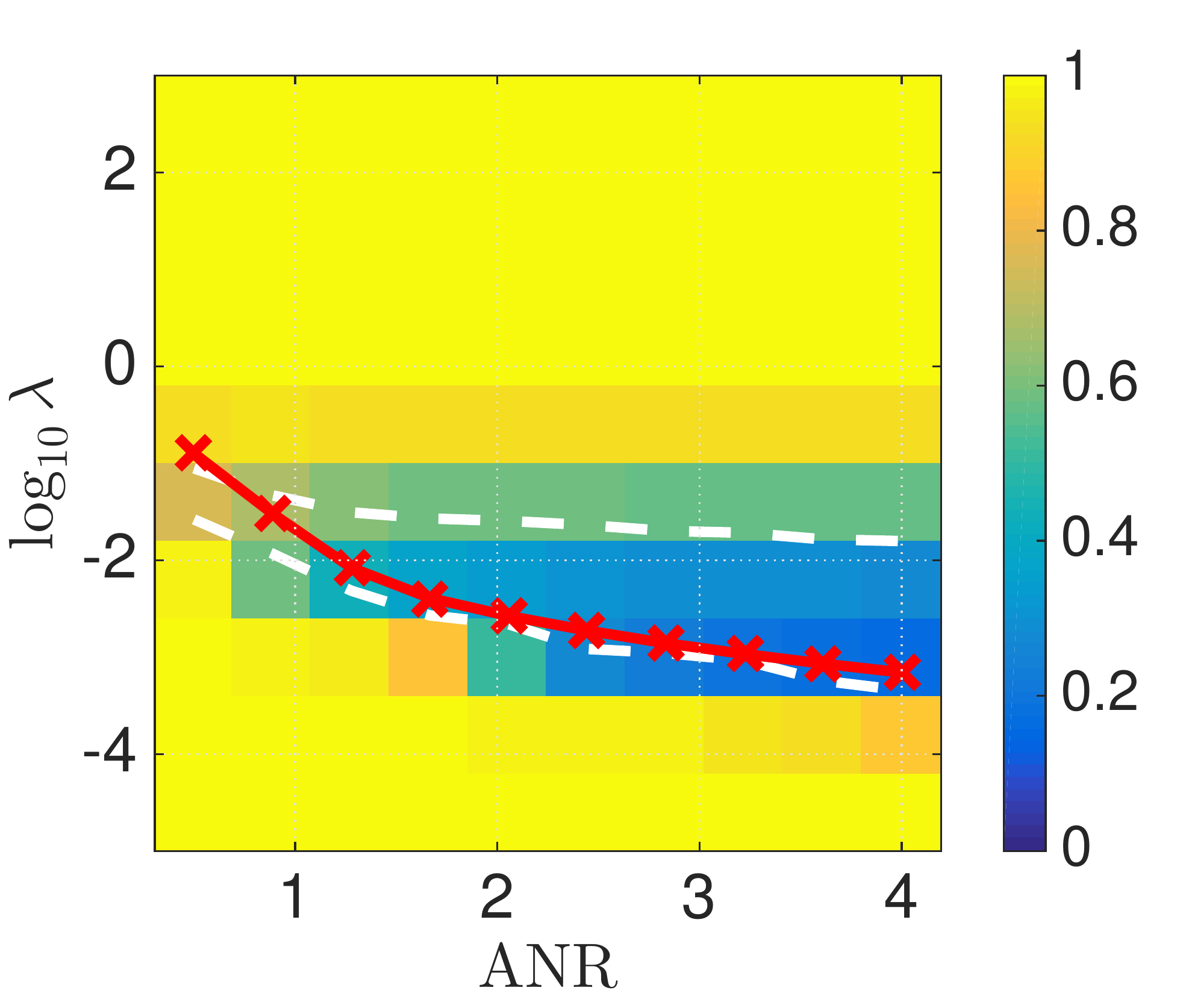}&
\includegraphics[width=.15\linewidth,clip=true,trim=0.2cm 0cm .5cm .5cm]{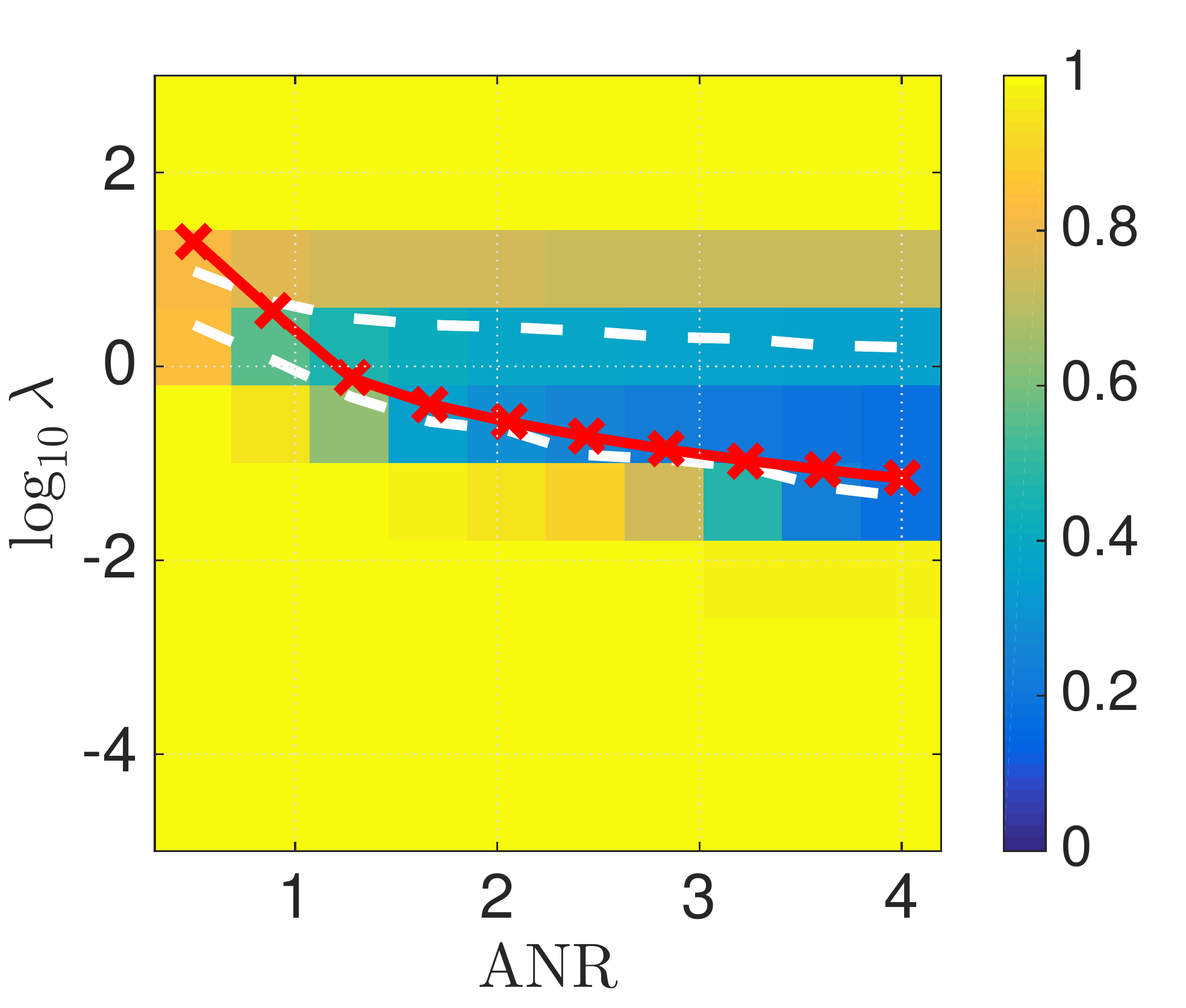}&
\includegraphics[width=.15\linewidth,clip=true,trim=0.2cm 0cm .5cm .5cm]{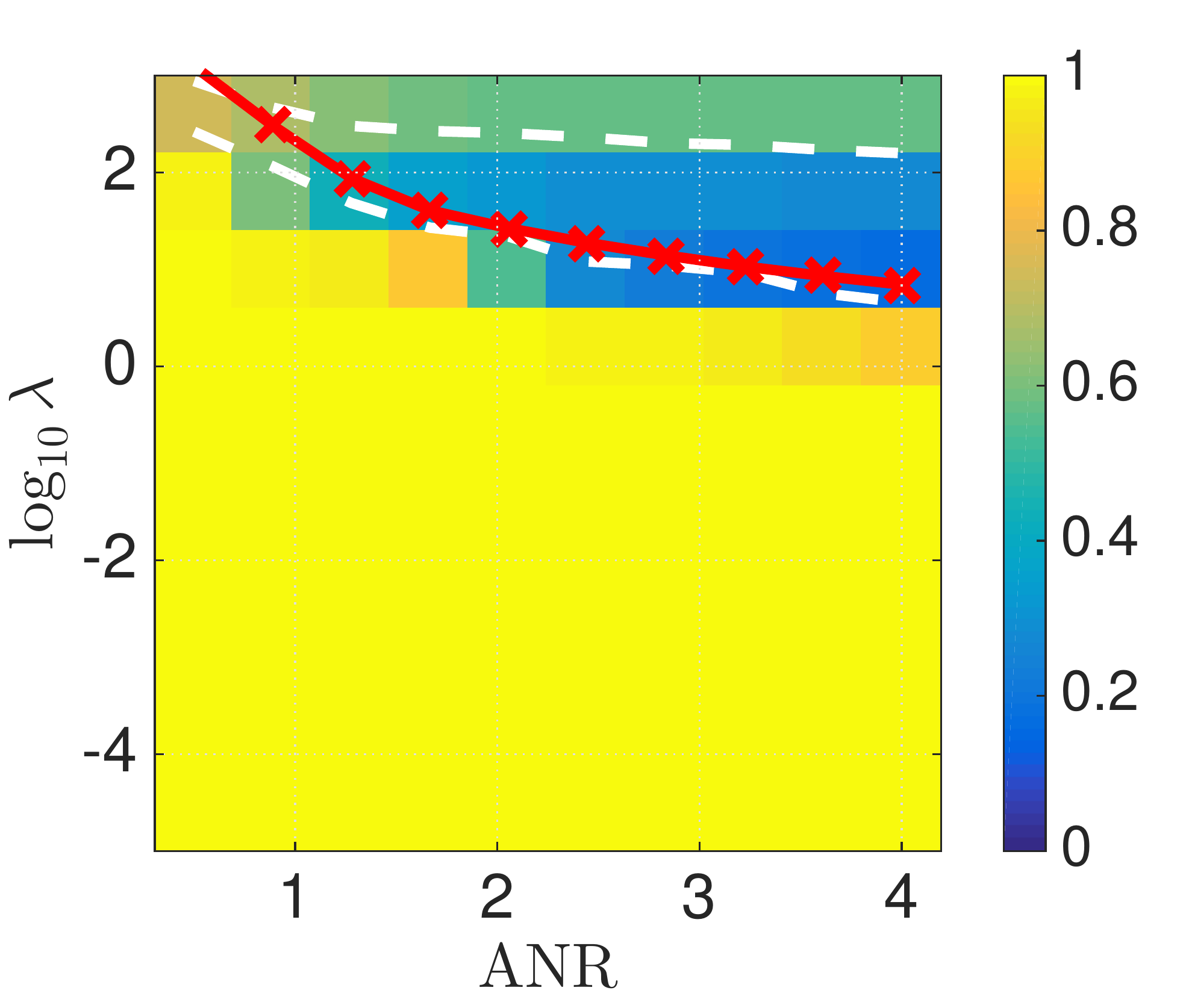}\\
\includegraphics[width=.15\linewidth,clip=true,trim=0.2cm 0cm .5cm .5cm]{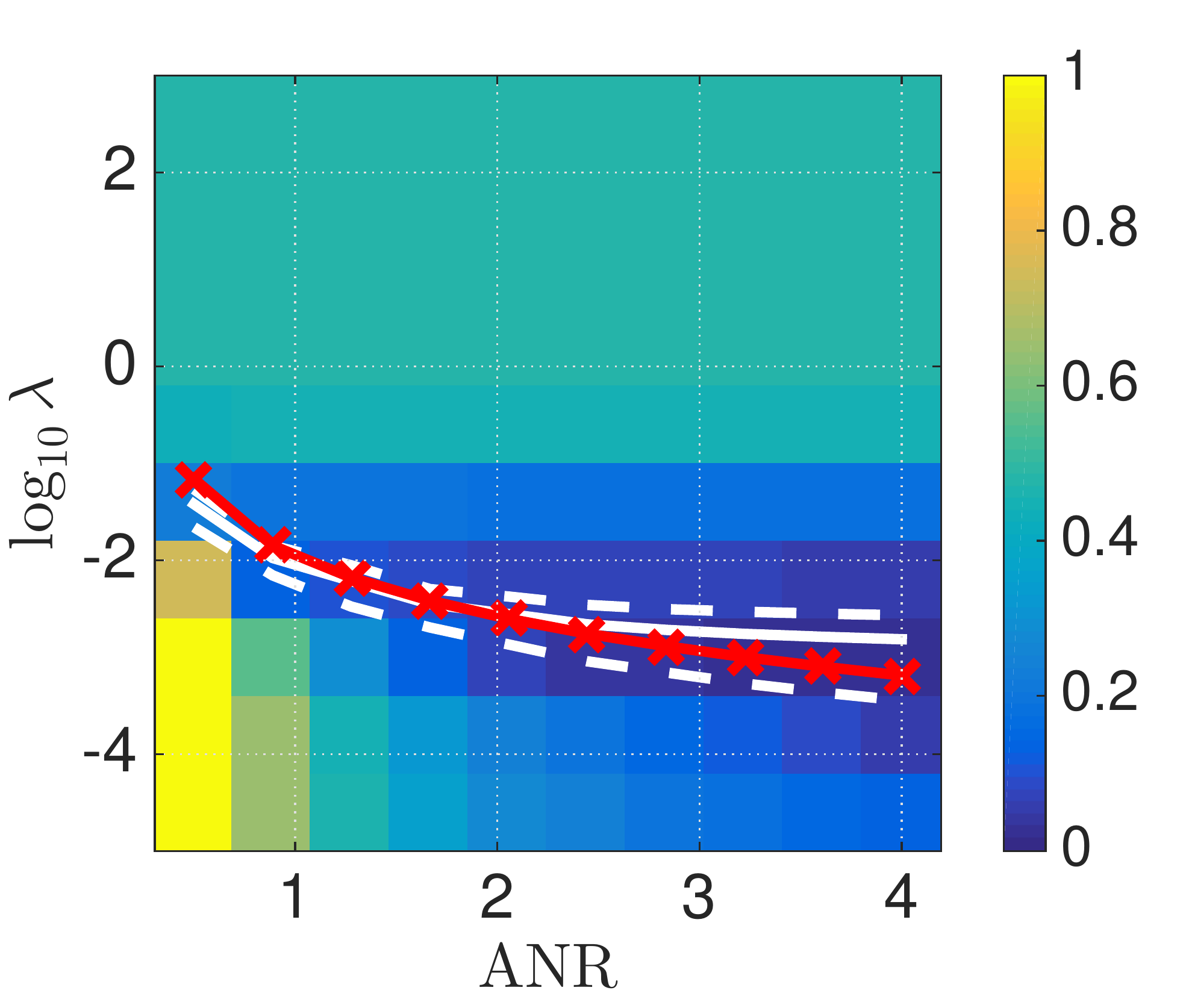}&
\includegraphics[width=.15\linewidth,clip=true,trim=0.2cm 0cm .5cm .5cm]{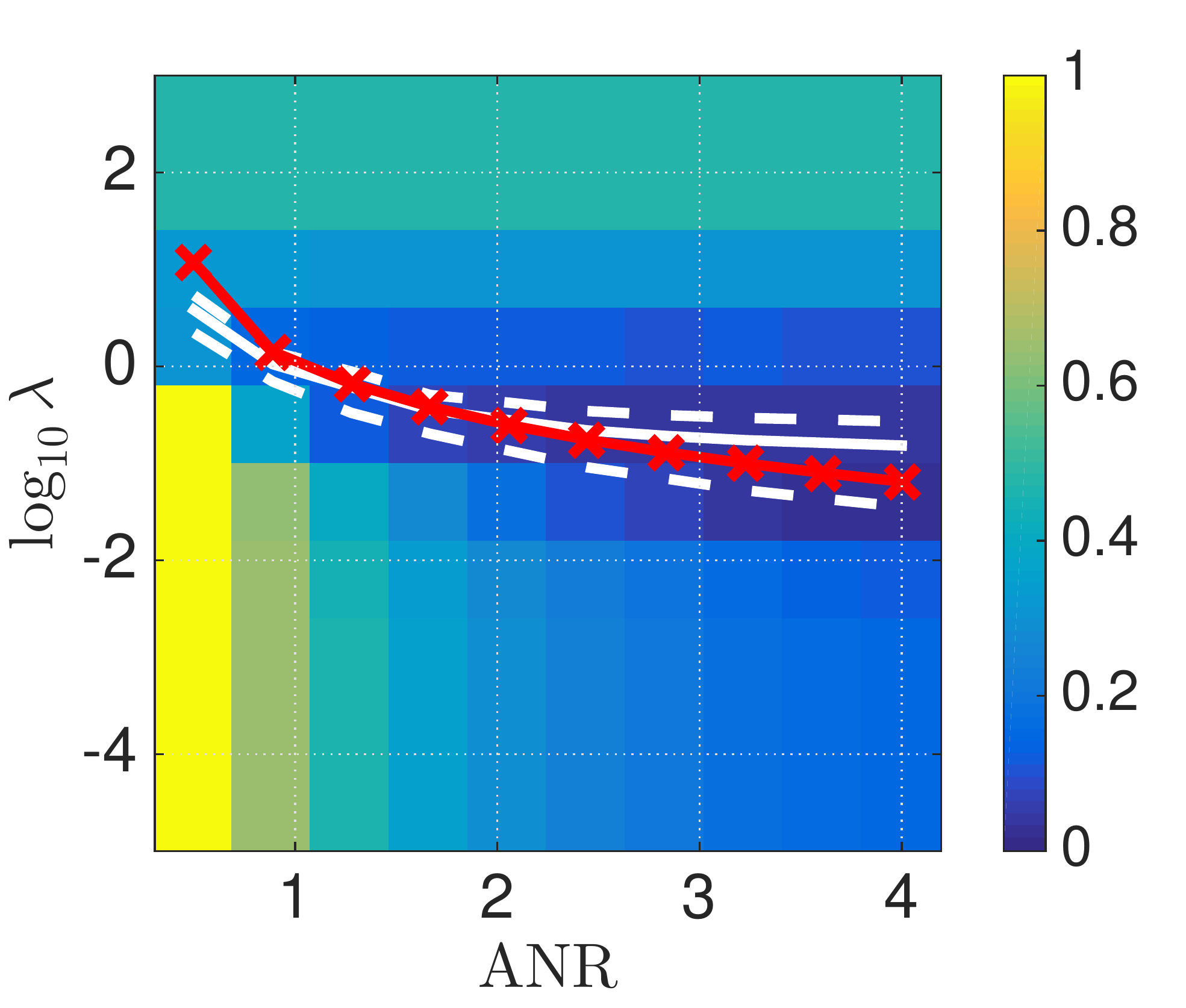}&
\includegraphics[width=.15\linewidth,clip=true,trim=0.2cm 0cm .5cm .5cm]{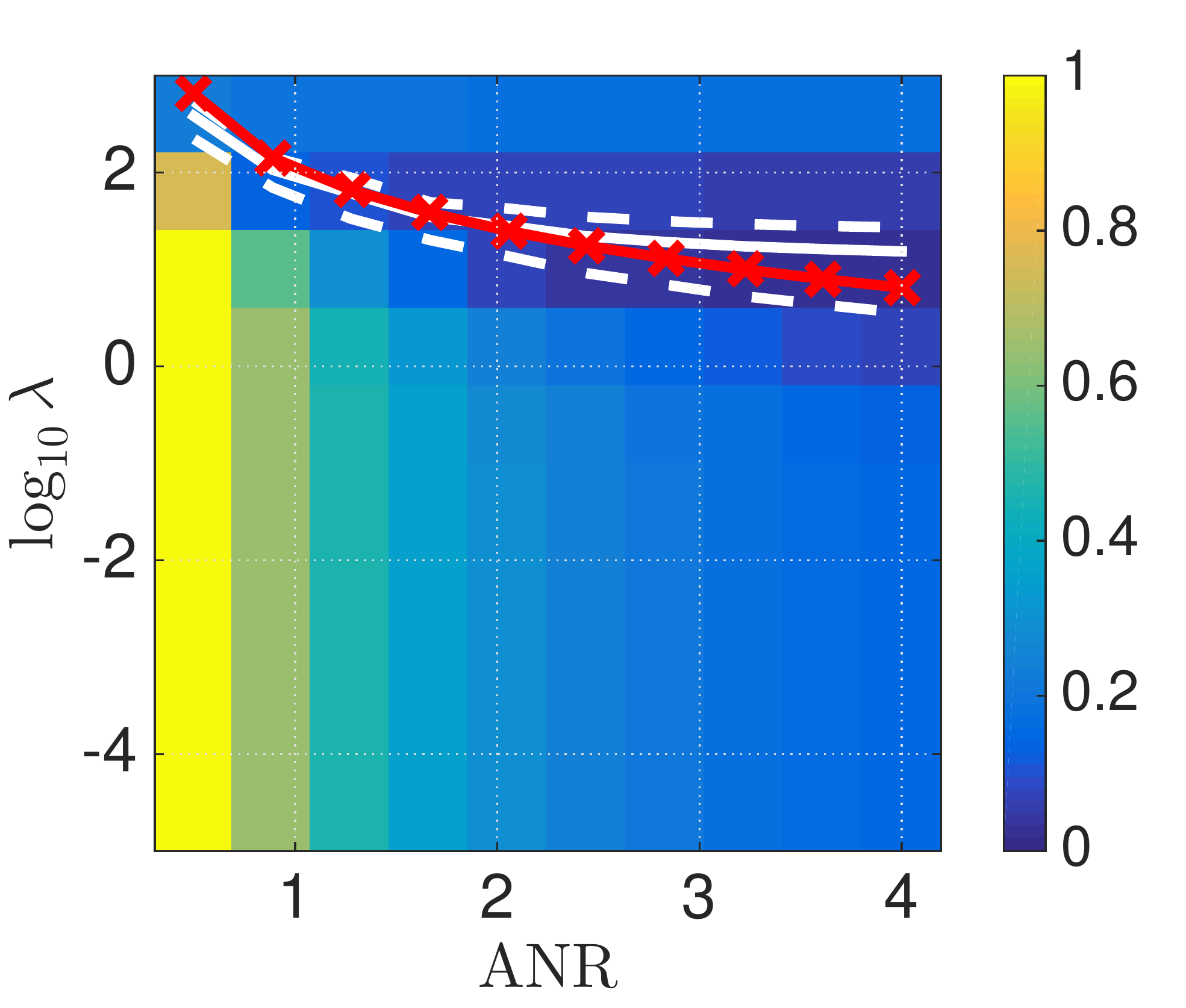}&
\includegraphics[width=.15\linewidth,clip=true,trim=0.2cm 0cm .5cm .5cm]{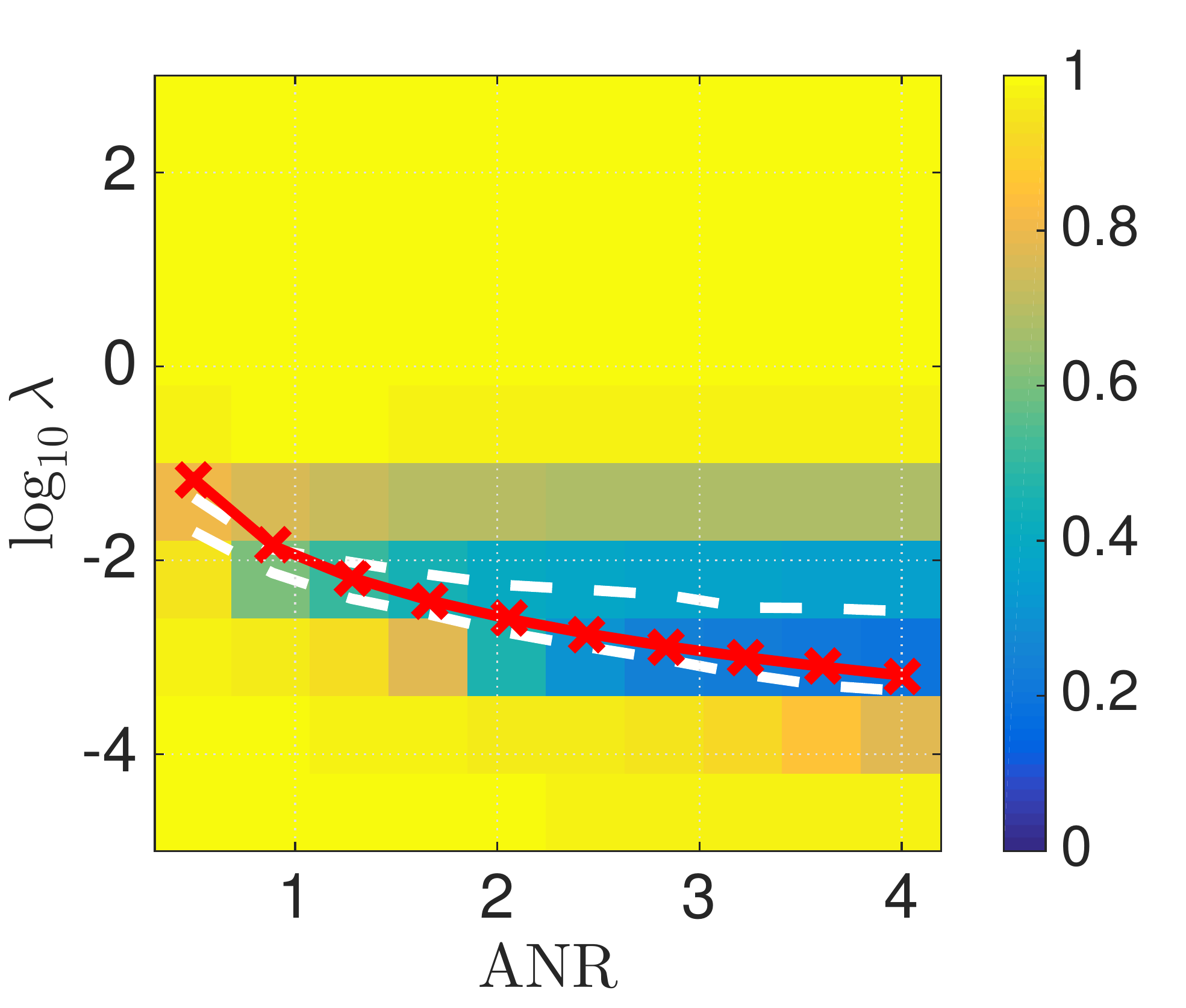}&
\includegraphics[width=.15\linewidth,clip=true,trim=0.2cm 0cm .5cm .5cm]{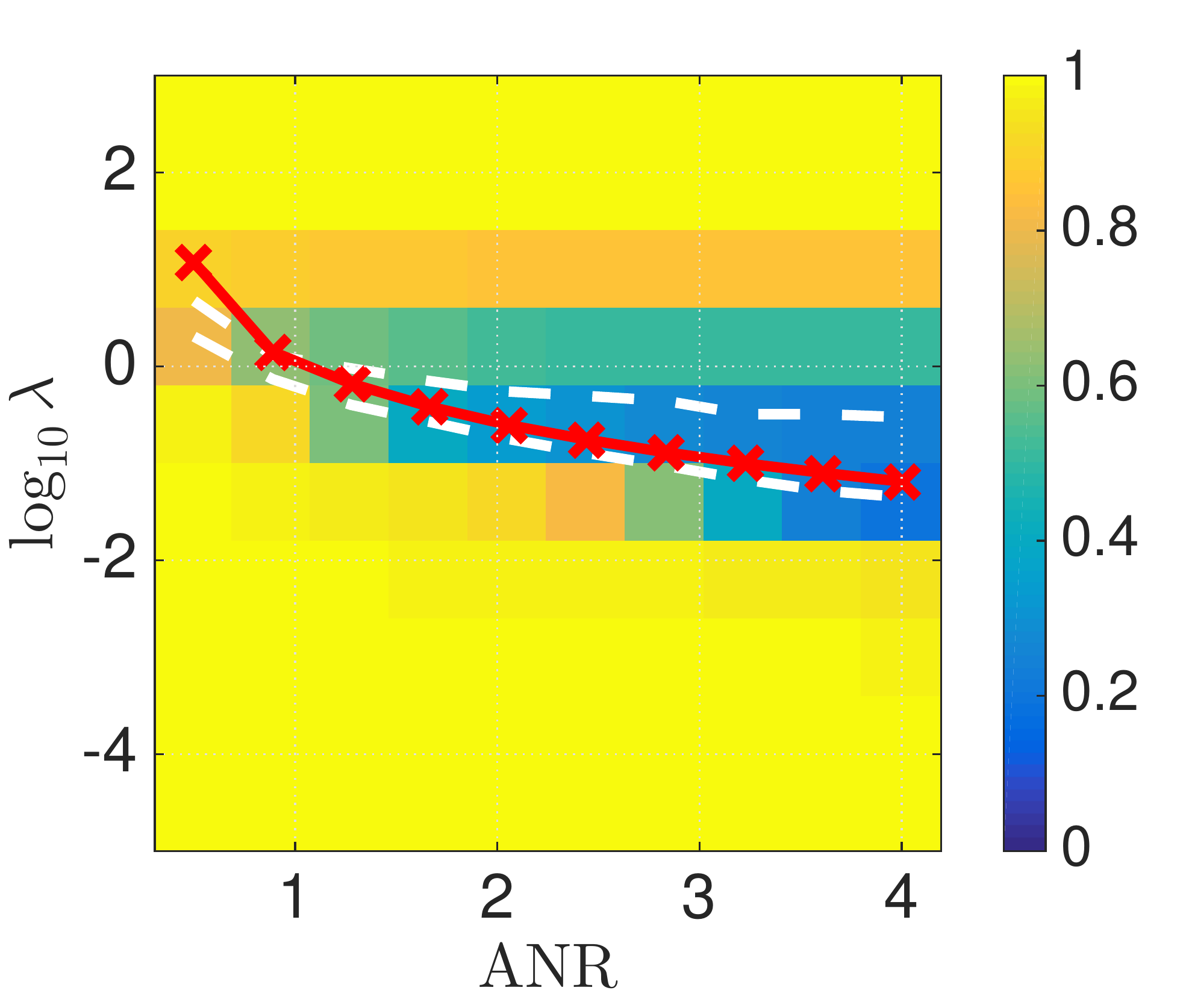}&
\includegraphics[width=.15\linewidth,clip=true,trim=0.2cm 0cm .5cm .5cm]{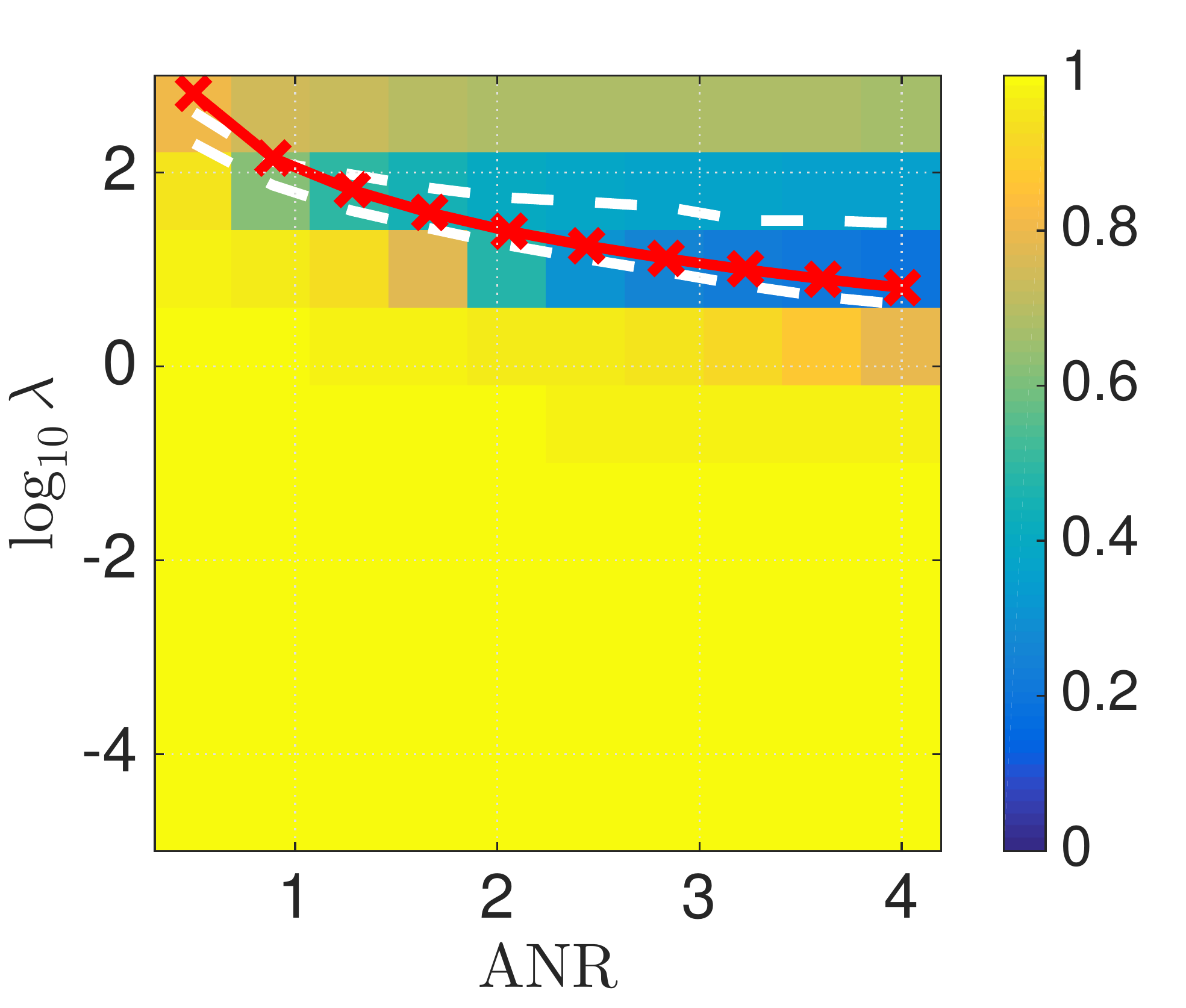}\\
\includegraphics[width=.15\linewidth,clip=true,trim=0.2cm 0cm .5cm .5cm]{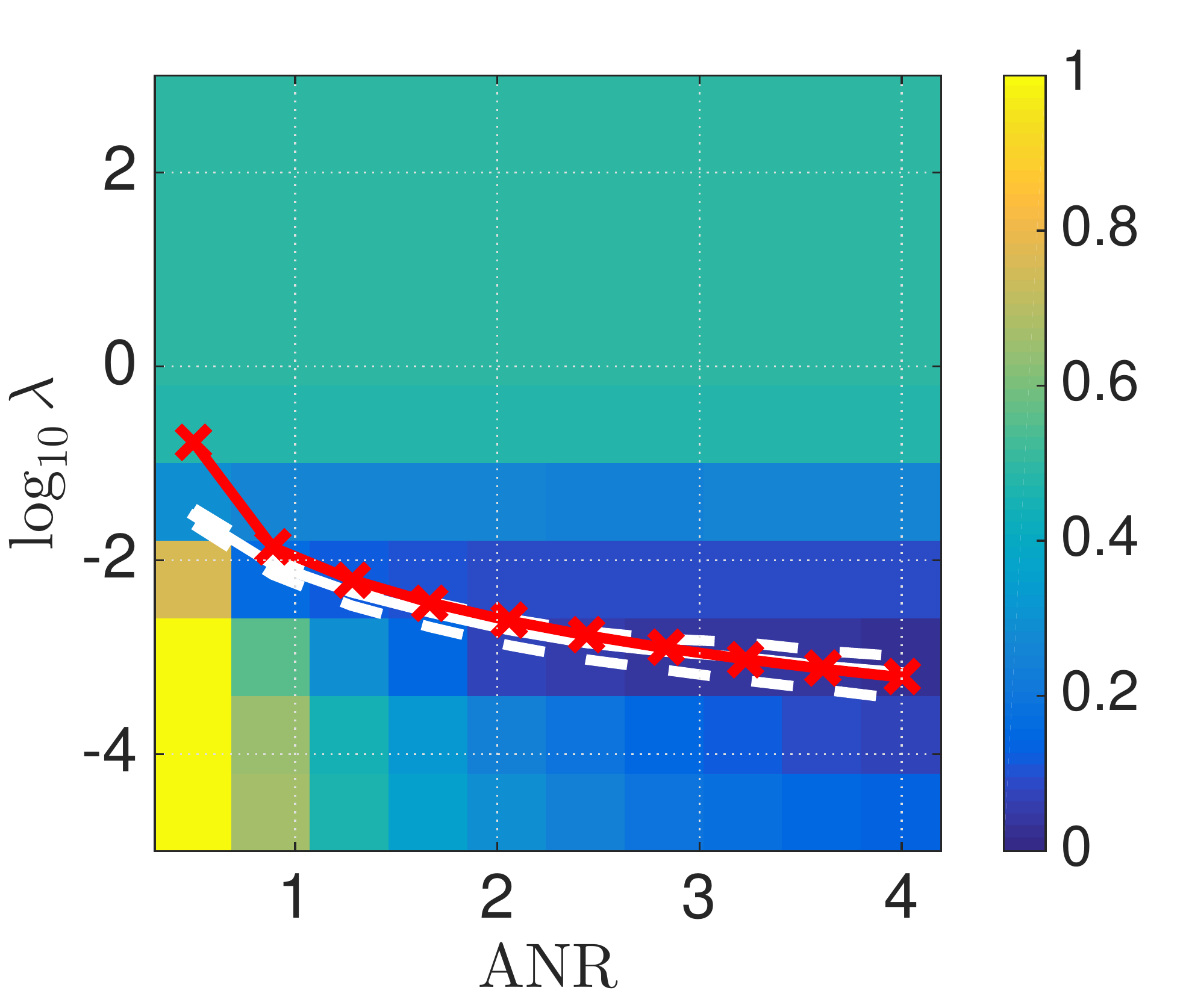}&
\includegraphics[width=.15\linewidth,clip=true,trim=0.2cm 0cm .5cm .5cm]{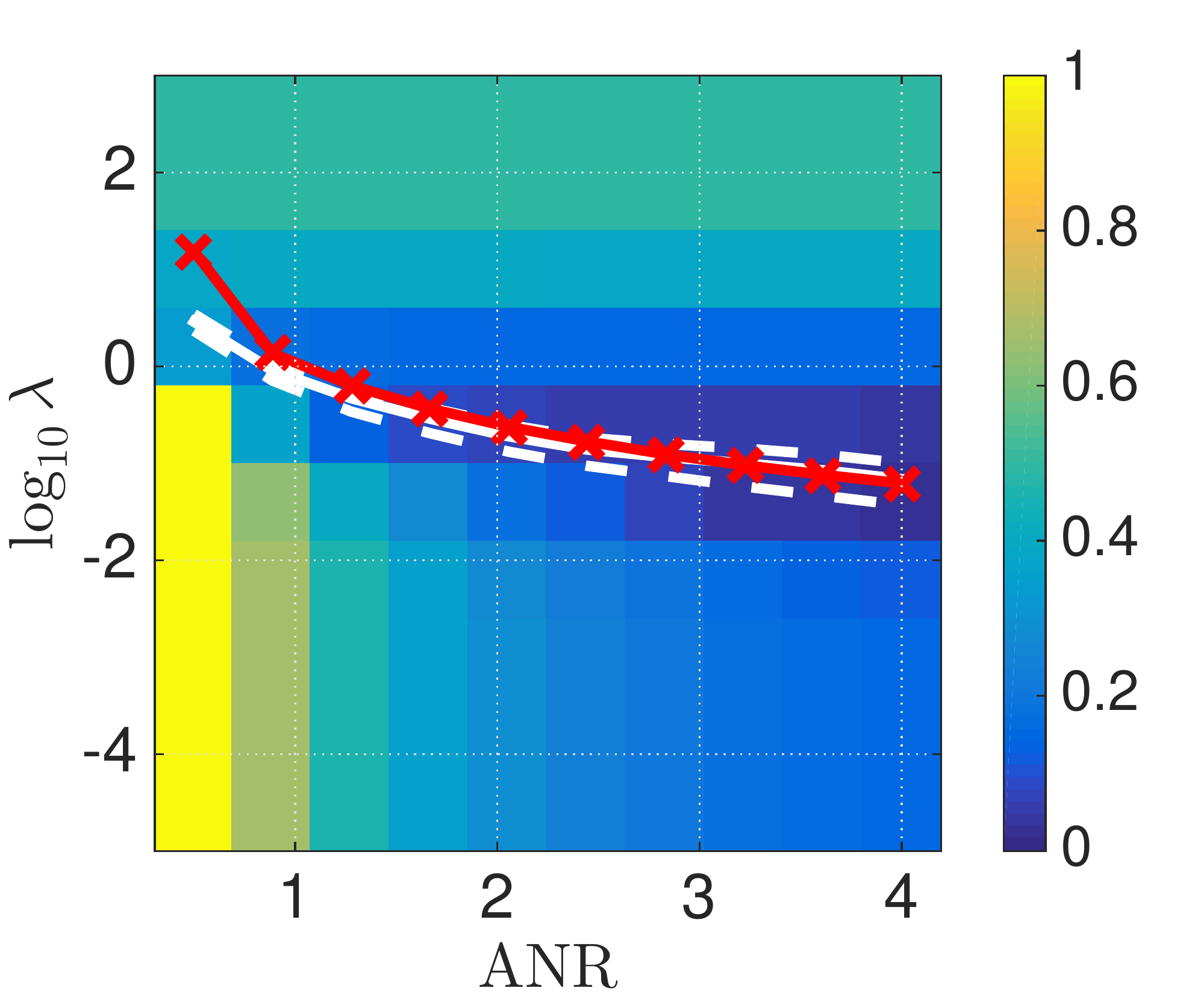}&
\includegraphics[width=.15\linewidth,clip=true,trim=0.2cm 0cm .5cm .5cm]{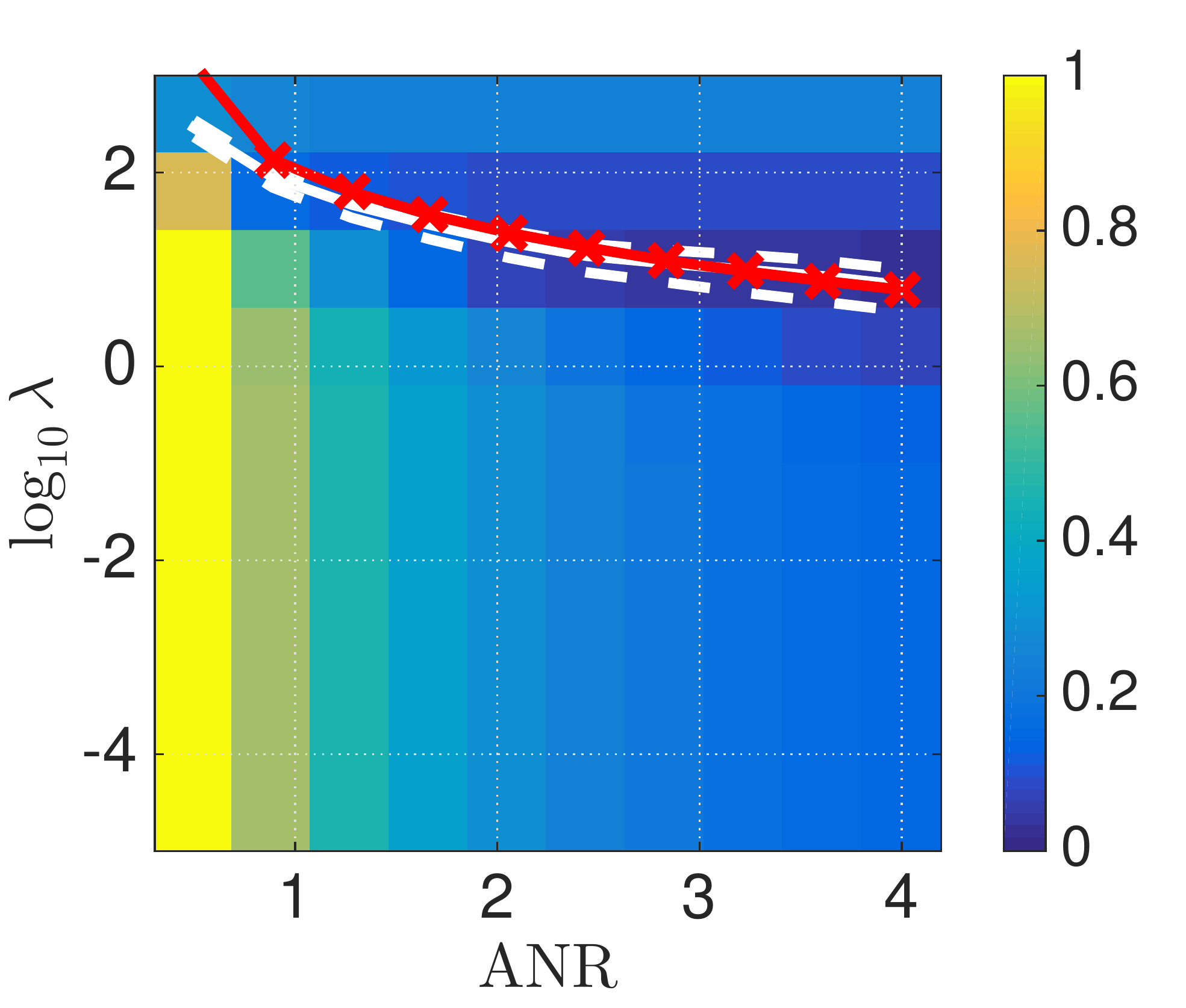}&
\includegraphics[width=.15\linewidth,clip=true,trim=0.2cm 0cm .5cm .5cm]{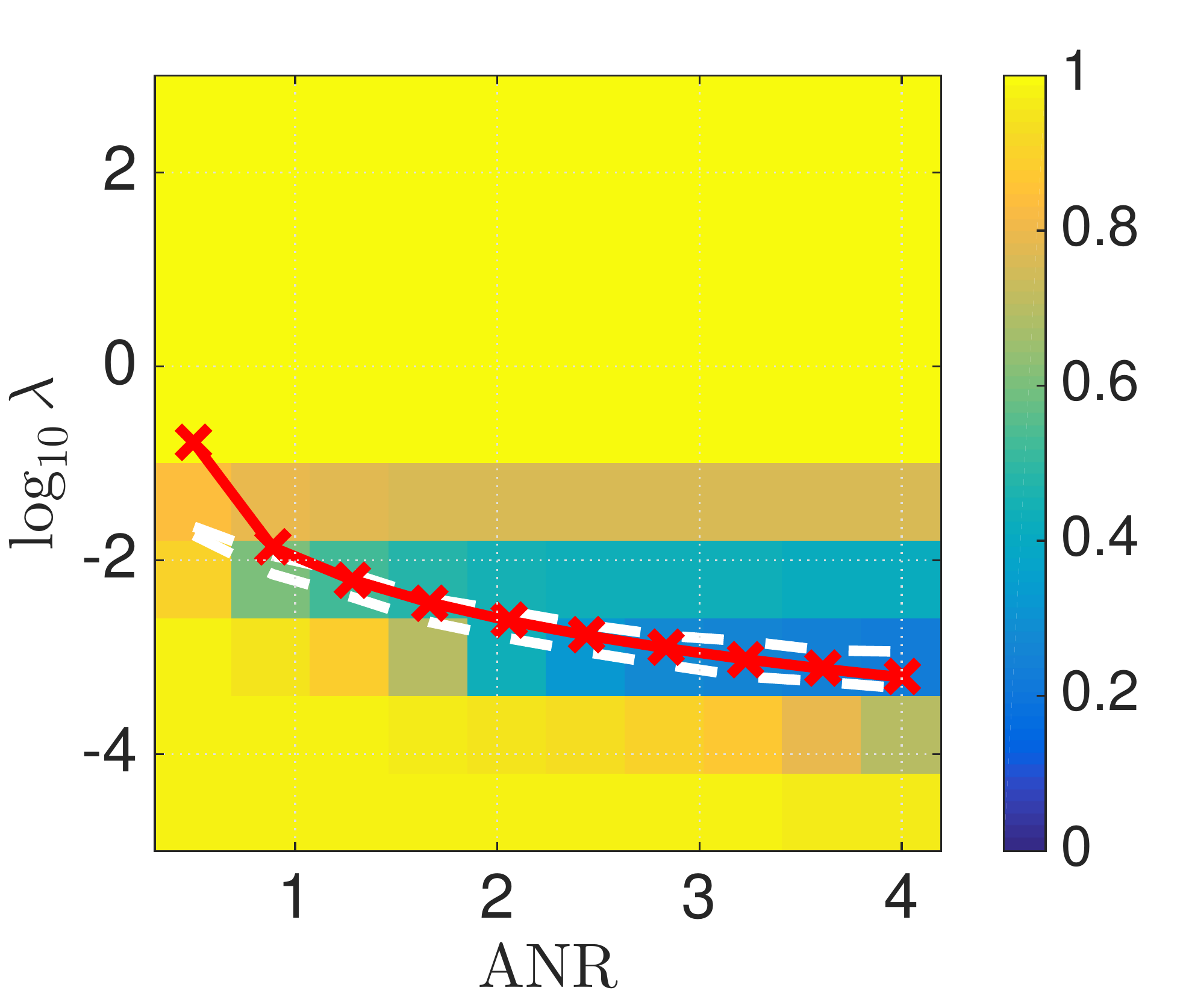}&
\includegraphics[width=.15\linewidth,clip=true,trim=0.2cm 0cm .5cm .5cm]{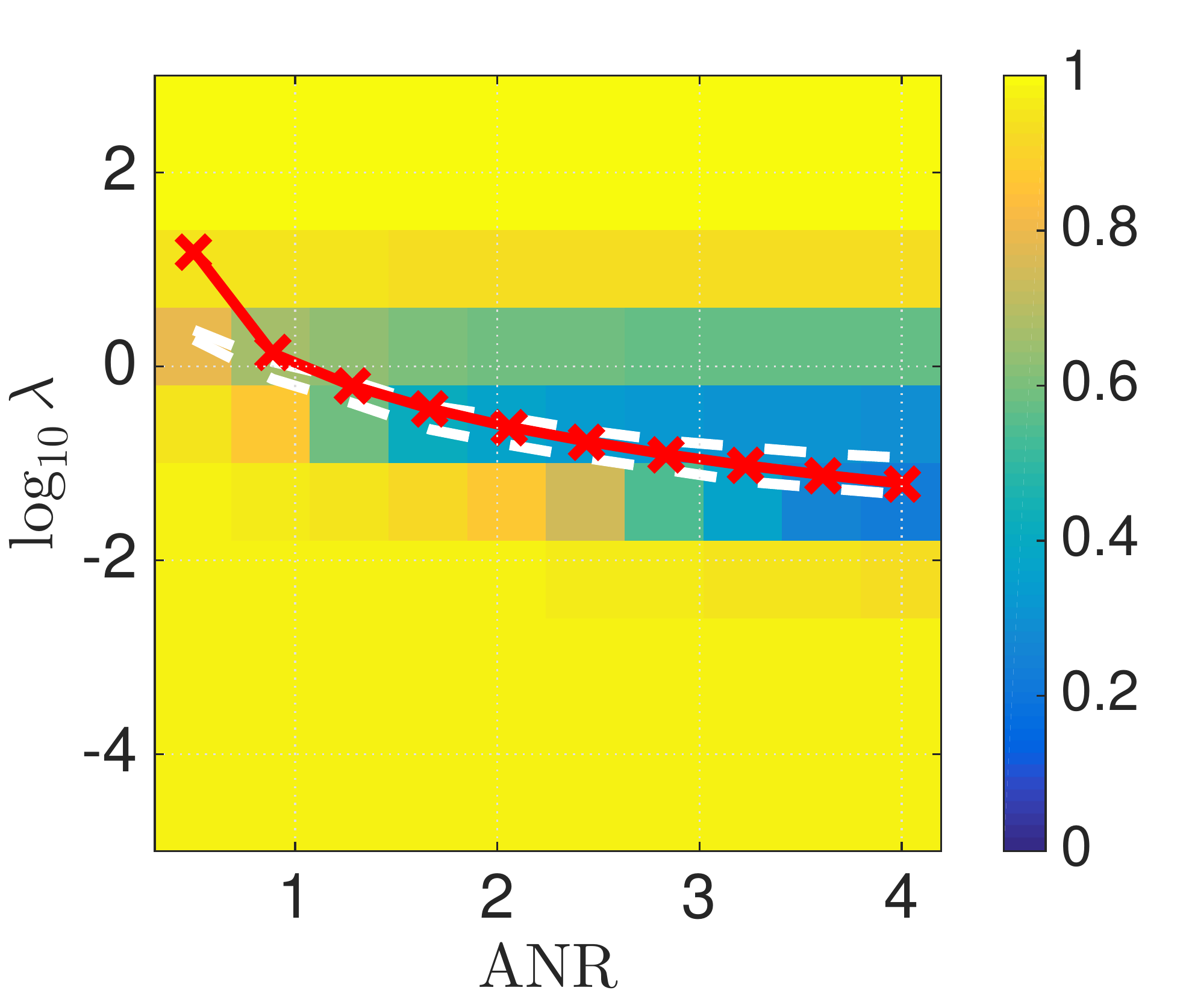}&
\includegraphics[width=.15\linewidth,clip=true,trim=0.2cm 0cm .5cm .5cm]{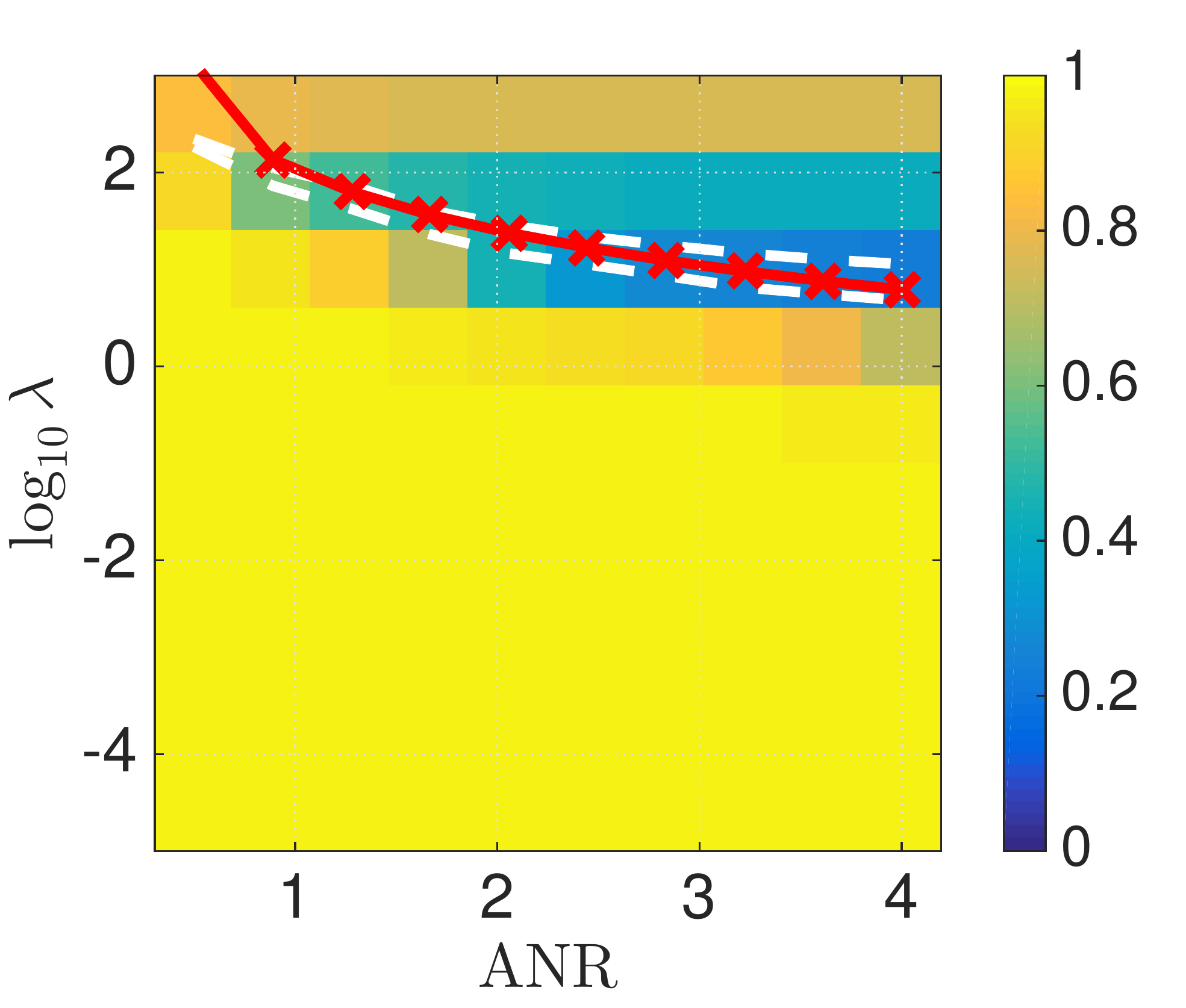}\\
\multicolumn{3}{c}{(a) relative MSE} &\multicolumn{3}{c}{(b) Jaccard error}
\end{tabular}
 \caption{\textbf{Estimation performance: RMSE and Jaccard error as functions of ${\lambda}$ and ${\rm ANR}$.
 \label{fig:mes_rmse2}}
The background displays the relative MSE (left) or Jaccard error (right) w.r.t. $\lambda$ and ANR. We superimpose in red the estimate $\widehat{\lambda}$ (average over $50$ realizations), as a function of the $\rm ANR$, which is shown to satisfactorily remain within the range
of oracles $ \lambda$, delimited by dashed white lines and to closely follow oracle Monte Carlo average indicated by solid white lines (left: relative MSE, right: Jaccard error).
 From top to bottom: $p=0.005$, $0.010$ and $0.015$.
 From left to right: $\overline{x}_{\max} - \overline{x}_{\min}=0.1$, $1$, and $10$.
}
\end{figure*}

First, Fig.~\ref{fig:mes_rmse2} reports estimation performance for $\widehat{\lambda}$ as a function of the ANR.
It shows that the estimated value $\widehat{\lambda}$ (red), averaged over Monte Carlo simulations, satisfactorily remains within the range of MSE/Jaccard error oracle values for $\lambda$ (dashed white lines) and tightly follows the average oracle values (solid white line).
This holds for different $\overline{x}_{\max} - \overline{x}_{\min}$.
As $p$ grows (cf. Fig.~\ref{fig:mes_rmse2} from top to bottom), the oracle regions in dashed white shrink, thus indicating that the selection of $\lambda$ becomes more intricate when more segments are to be detected.
The proposed automated selection for $\lambda$ still performs well in these more difficult situations.
In addition, it can also be observed that $\widehat{\lambda}$ depends, as expected, on $\sigma$ (or equivalently on $\overline{x}_{\max} - \overline{x}_{\min}$) cf. Fig.~\ref{fig:mes_rmse2}, from left to right.

\begin{figure*}
\begin{tabular}{@{}c@{}c@{}cc@{}c@{}c@{}}
\includegraphics[width=.162\linewidth,clip=true,trim=.3cm 0cm .5cm .5cm]{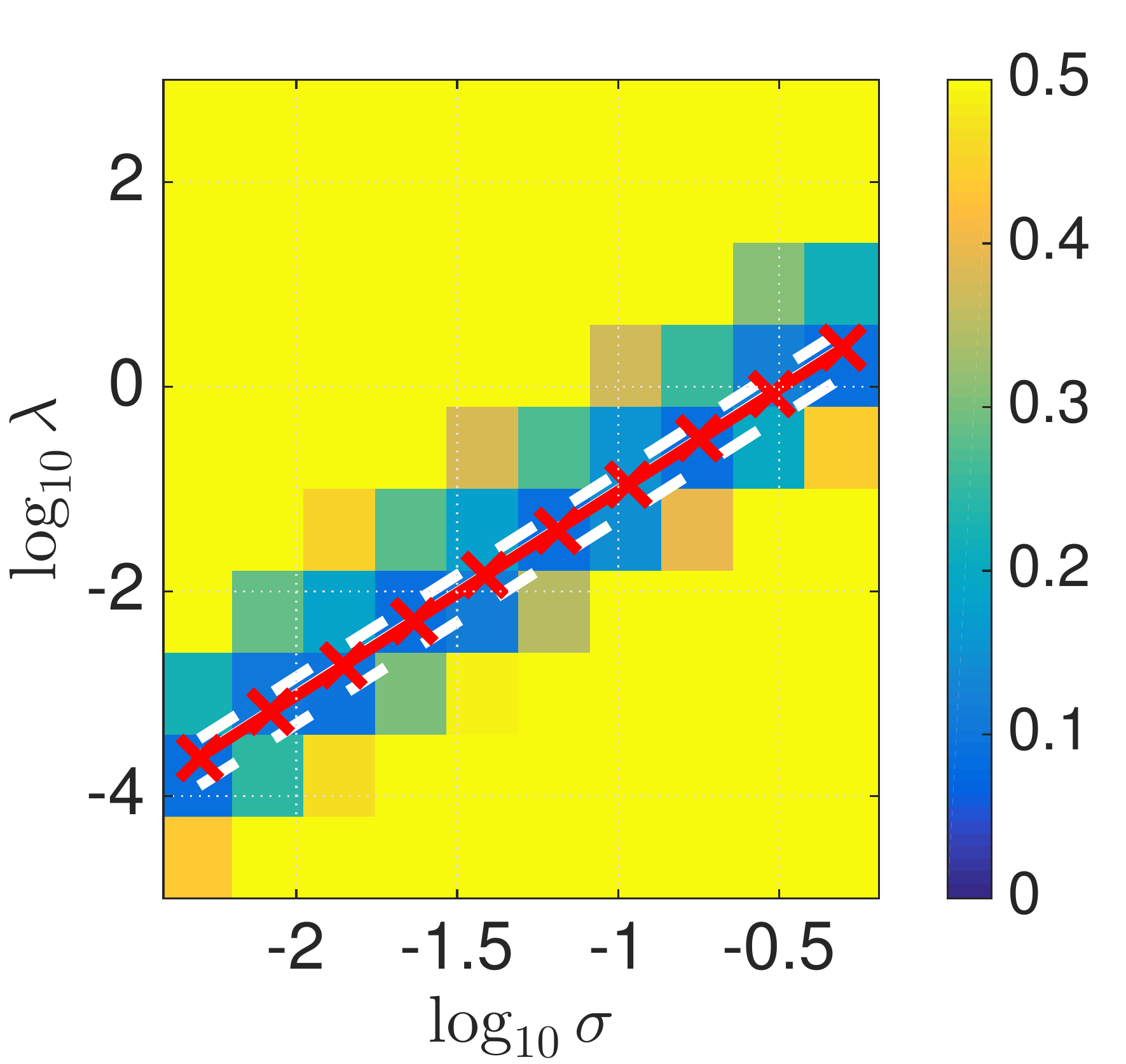}&
\includegraphics[width=.162\linewidth,clip=true,trim=.3cm 0cm .5cm .5cm]{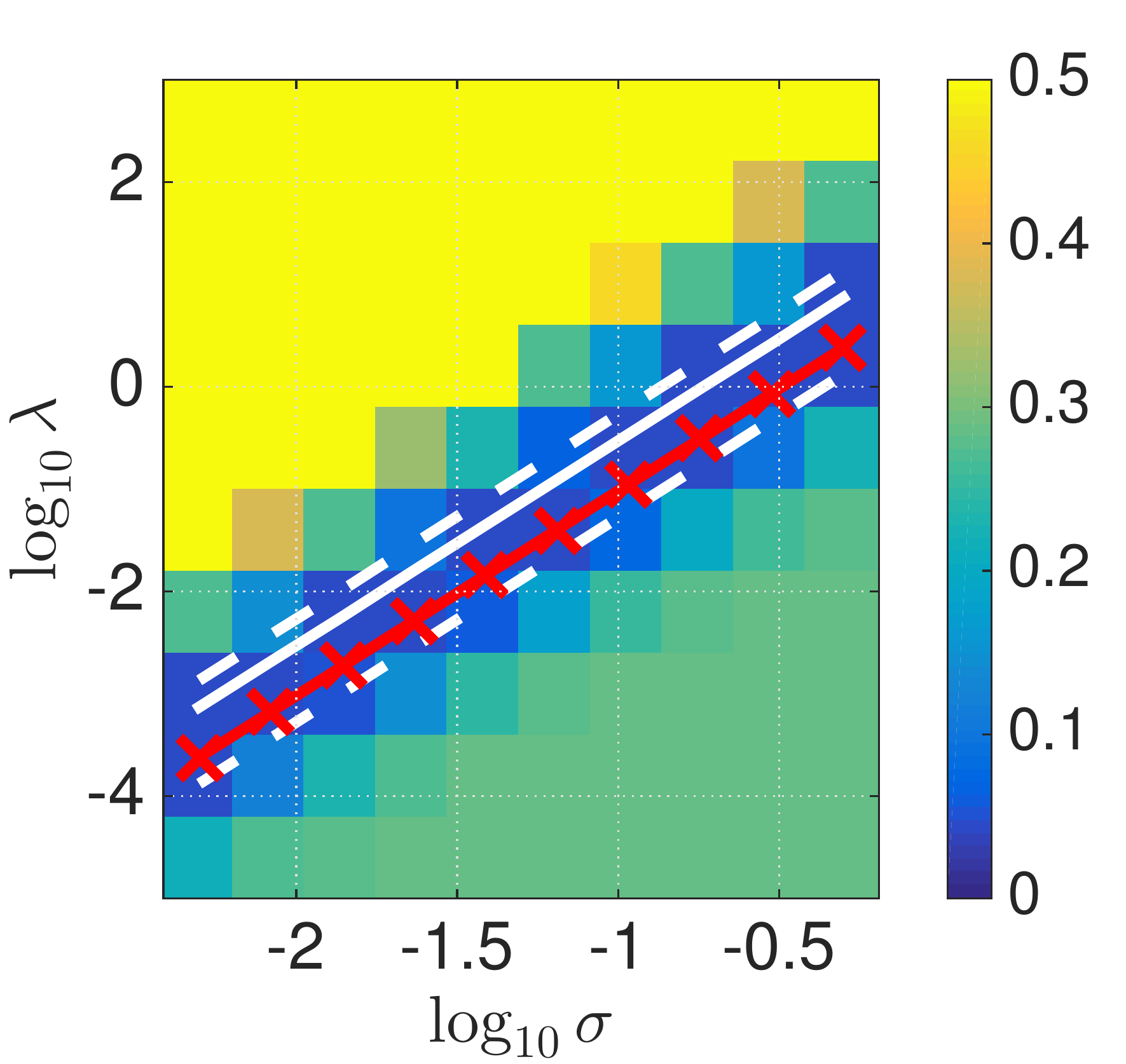}&
\includegraphics[width=.162\linewidth,clip=true,trim=.3cm 0cm .5cm .5cm]{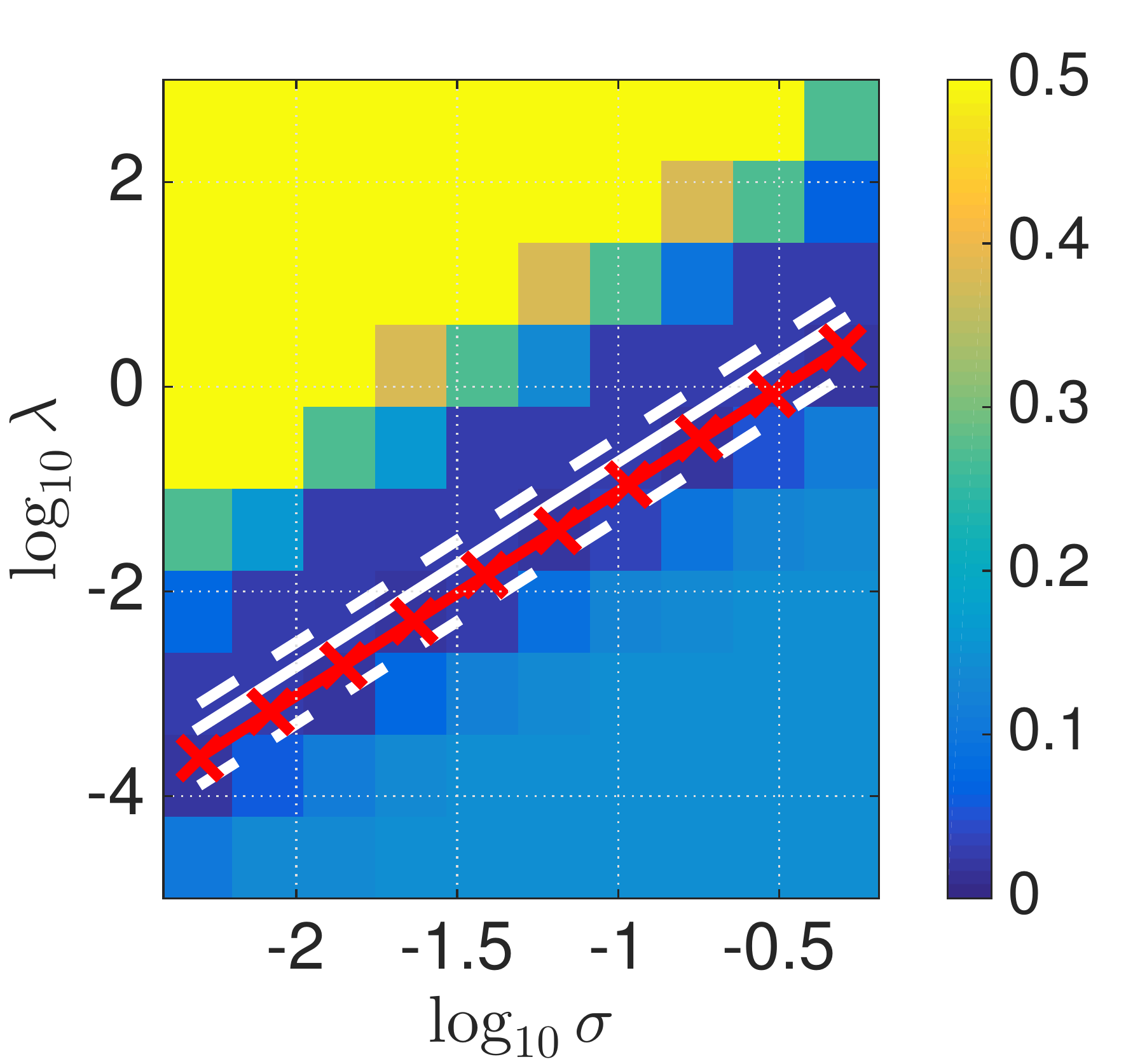}&
\includegraphics[width=.162\linewidth,clip=true,trim=.3cm 0cm .5cm .5cm]{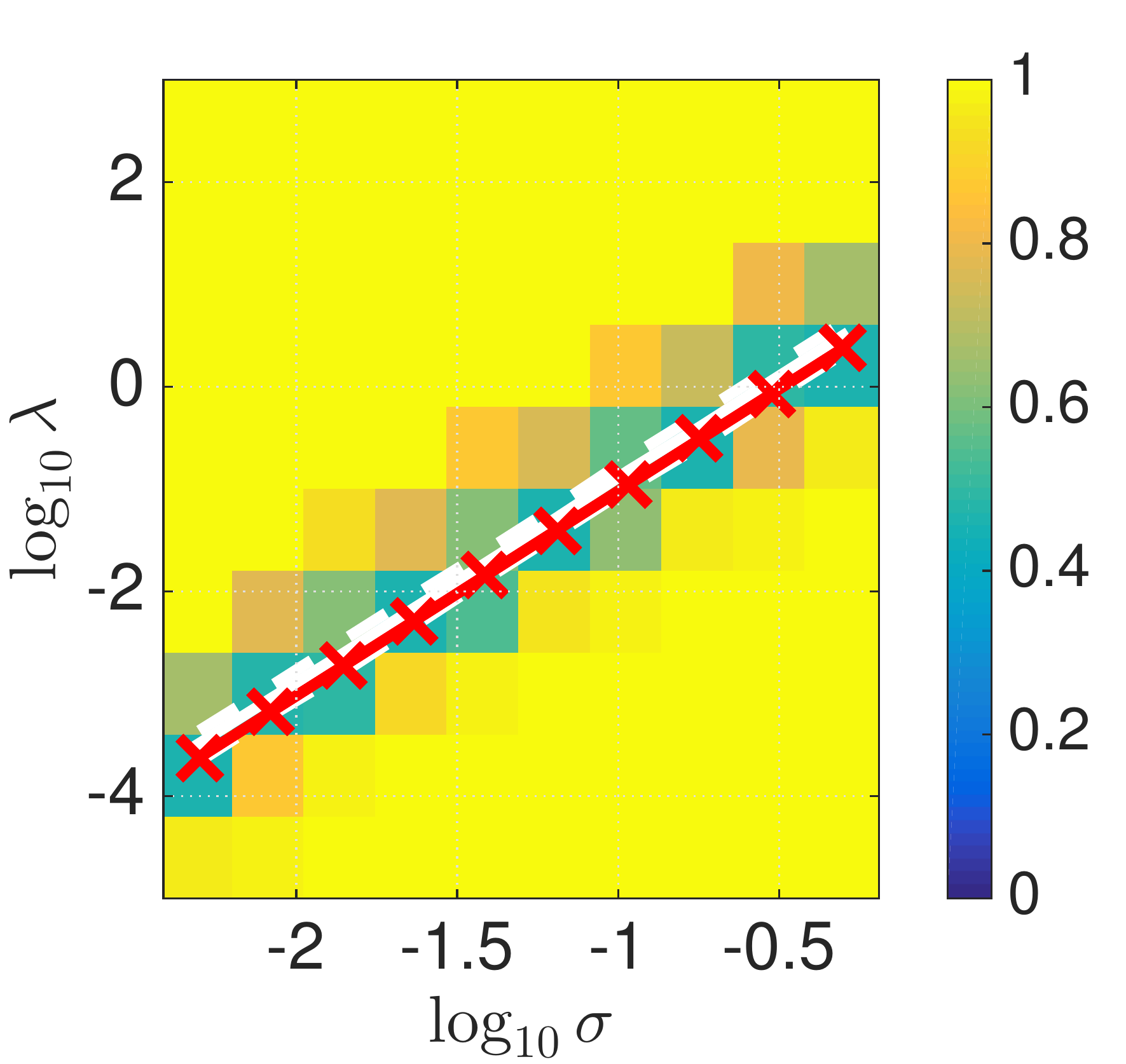}&
\includegraphics[width=.162\linewidth,clip=true,trim=.3cm 0cm .5cm .5cm]{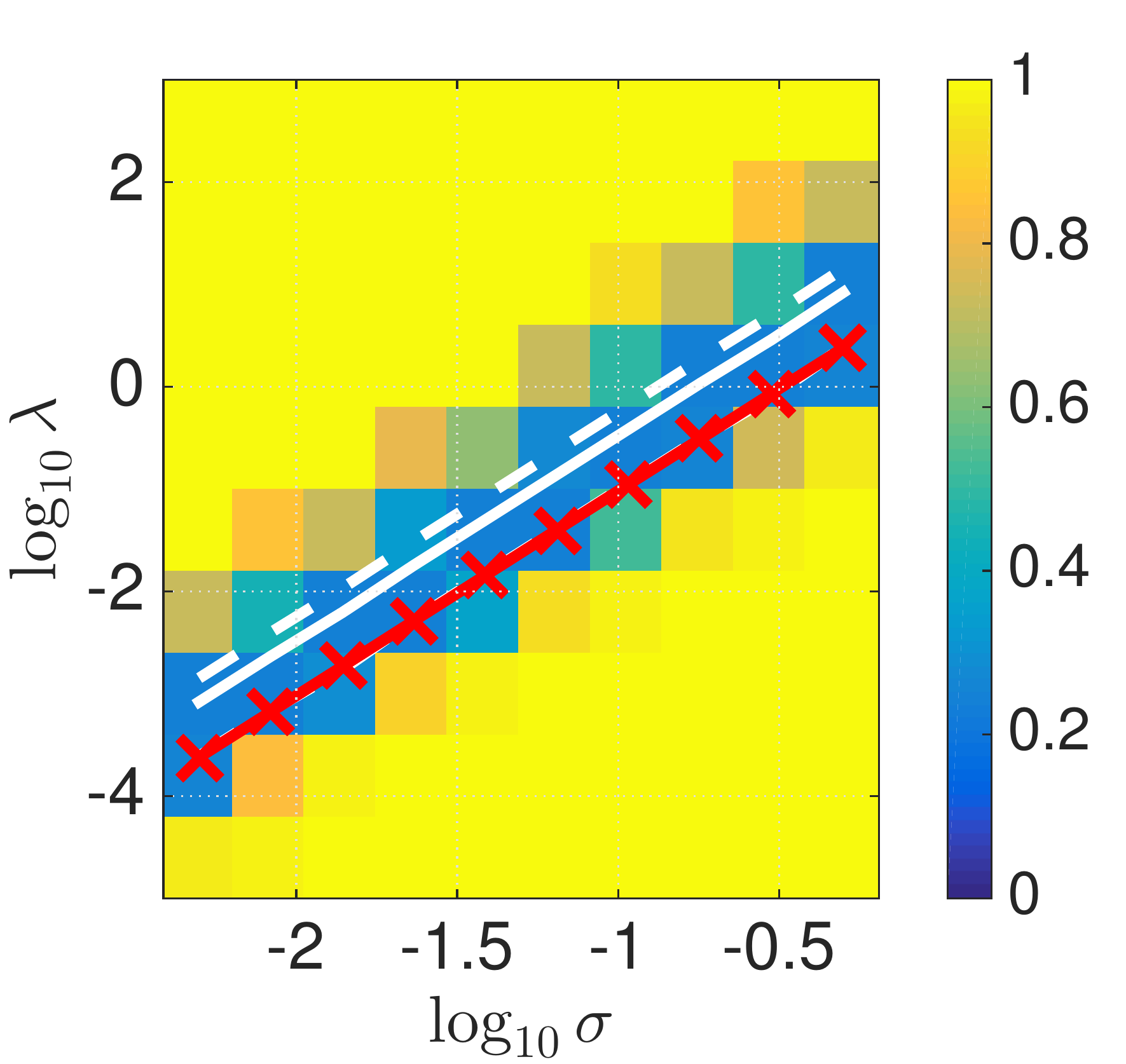}&
\includegraphics[width=.162\linewidth,clip=true,trim=.3cm 0cm .5cm .5cm]{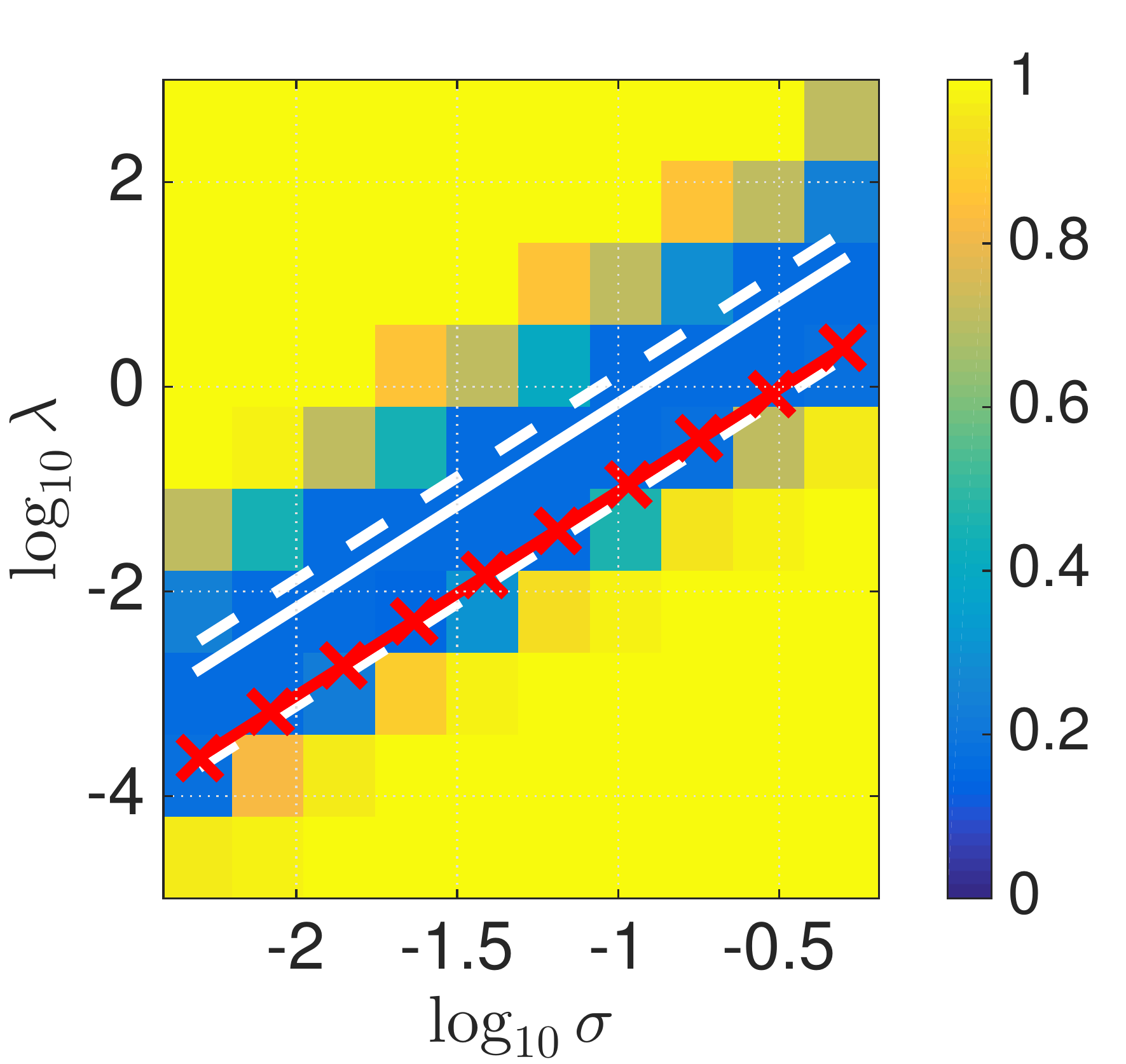}\\
\multicolumn{3}{c}{(a) relative MSE} &\multicolumn{3}{c}{(b) Jaccard error}
\end{tabular}
 \caption{\textbf{Estimation performance: RMSE and Jaccard error as functions of ${\lambda}$ and $\sigma$. 
 \label{fig:mes_rmse}}
The background displays the relative MSE (left) or Jaccard error (right) w.r.t. $\lambda$ and $\sigma$. We superimpose in red the estimate $\widehat{\lambda}$ (average over $50$ realizations), as a function of $ \log_{10} \sigma$, which is shown to satisfactorily remain within the range
of oracles $ \lambda$, delimited by dashed white lines and to closely follow oracle Monte Carlo average indicated by solid white lines (left: relative MSE, right: Jaccard error).
For each configuration $p=0.01$ and from left to right: ${\rm ANR}=1$, $2$ and $4$.
 This illustrates that $\widehat{\lambda}$ leads to solutions with same performance as oracle $\lambda$ and highlights that $\widehat{\lambda}$ varies linearly with $\sigma^2$, as expected.
 }
\end{figure*}

Second, Fig.~\ref{fig:mes_rmse} focuses on the behavior of  the estimated $\widehat{\lambda}$ as a function of $\sigma$, for different values of ANR.
Again, it shows satisfactory performance of $\widehat{\lambda}$ compared to the oracle $\lambda$.
Incidently, it also very satisfactorily reproduces the linear dependence of $\lambda$ with $\sigma^2$, which can be predicted from a mere dimensional analysis
of the $\ell_2$-Potts model yielding:
\begin{equation}
\lambda \sim \frac{\sigma^2}{2p}.
\end{equation}

\subsection{Comparison with state-of-the-art estimators}
\label{ss:bayes}

The proposed method has been compared to classical Bayesian estimators associated with the hierarchical Bayesian model derived in Section~\ref{ss:hbm} for which an MCMC procedure has been derived  (cf.  Appendix~\ref{app:bayesianEstimator}). The number of Monte Carlo iterations is fixed to $T_{\rm MC}=10^3$  and the amplitude hyperprior parameters are chosen as the mean of  $\boldsymbol{y}$ for $\mu_0$ and $\sigma_0^2=\widehat{\mathrm{var}}(\boldsymbol{y})$, where $\widehat{\mathrm{var}}(\cdot)$ stands for the empirical variance.

\begin{figure*}
\includegraphics[width=.24\linewidth]{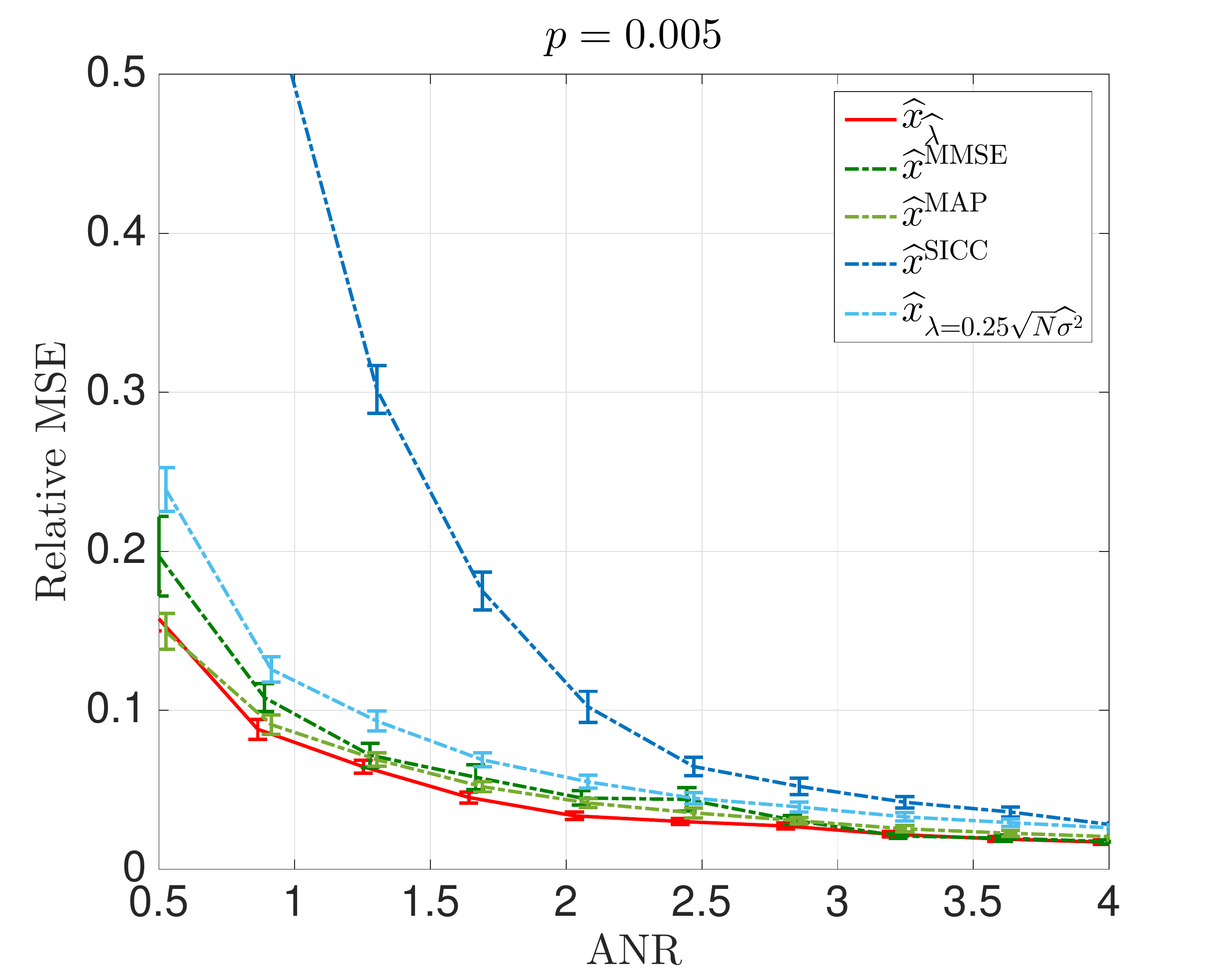}
\includegraphics[width=.24\linewidth]{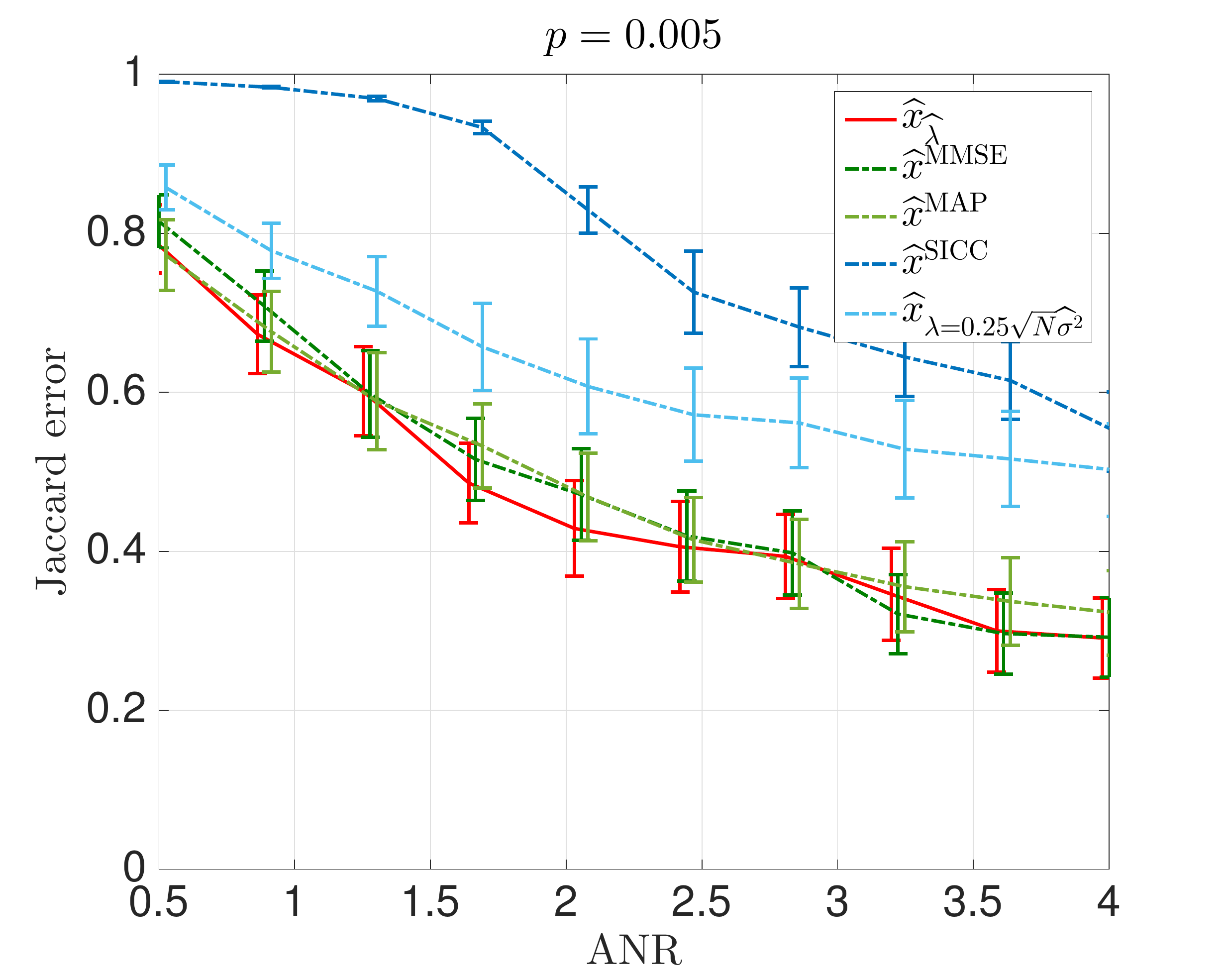}
\includegraphics[width=.24\linewidth]{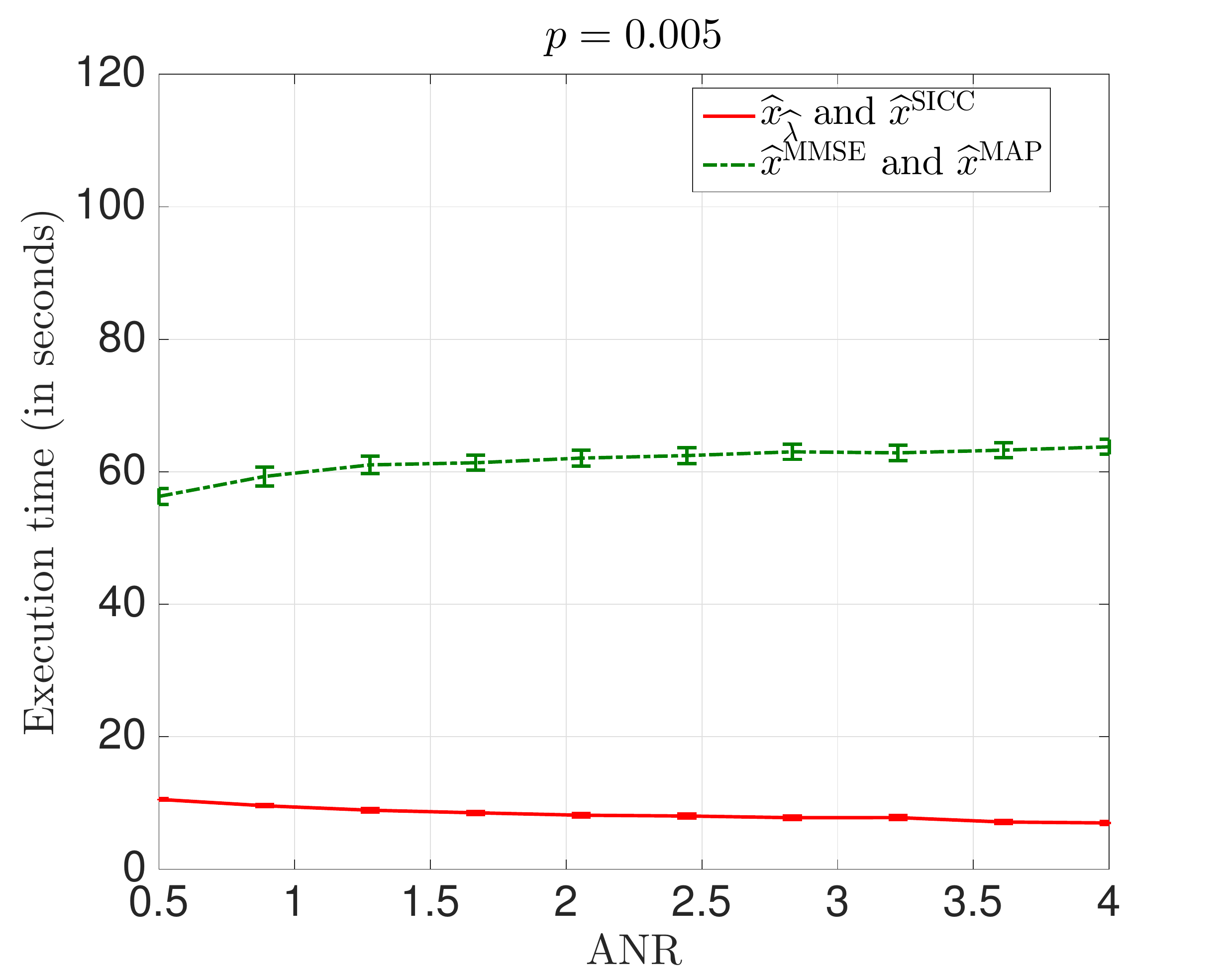}
\includegraphics[width=.24\linewidth]{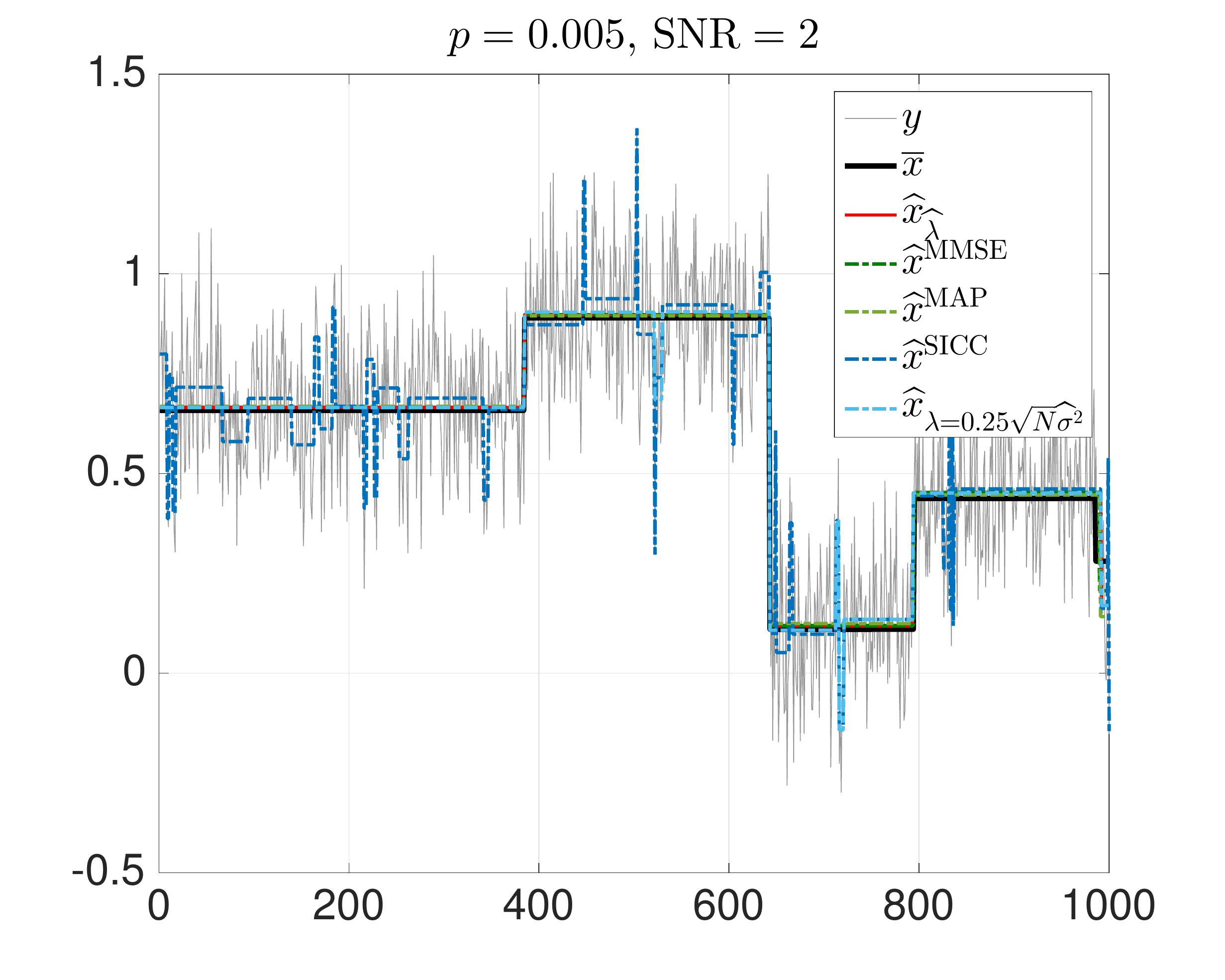}\\
\includegraphics[width=.24\linewidth]{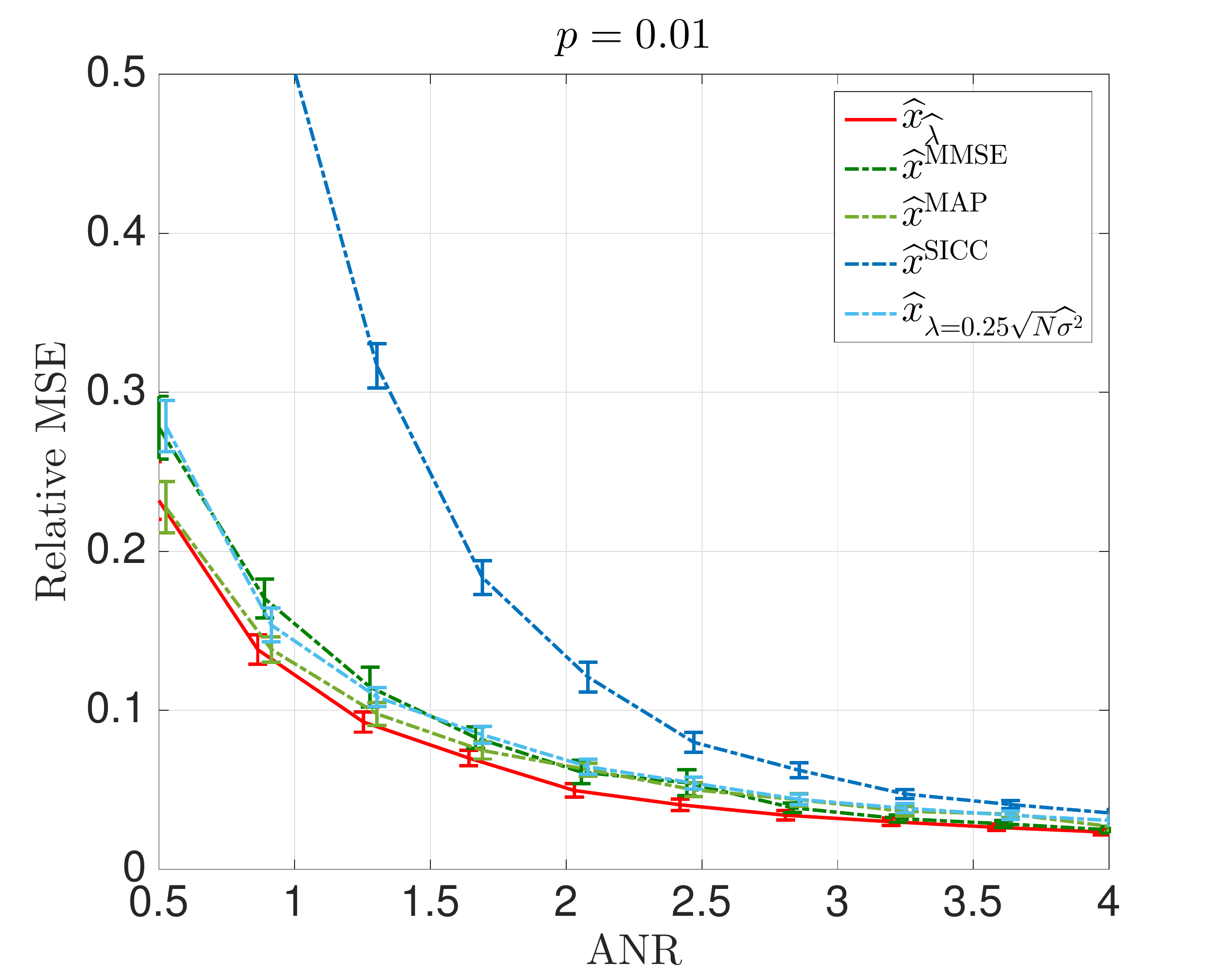}
\includegraphics[width=.24\linewidth]{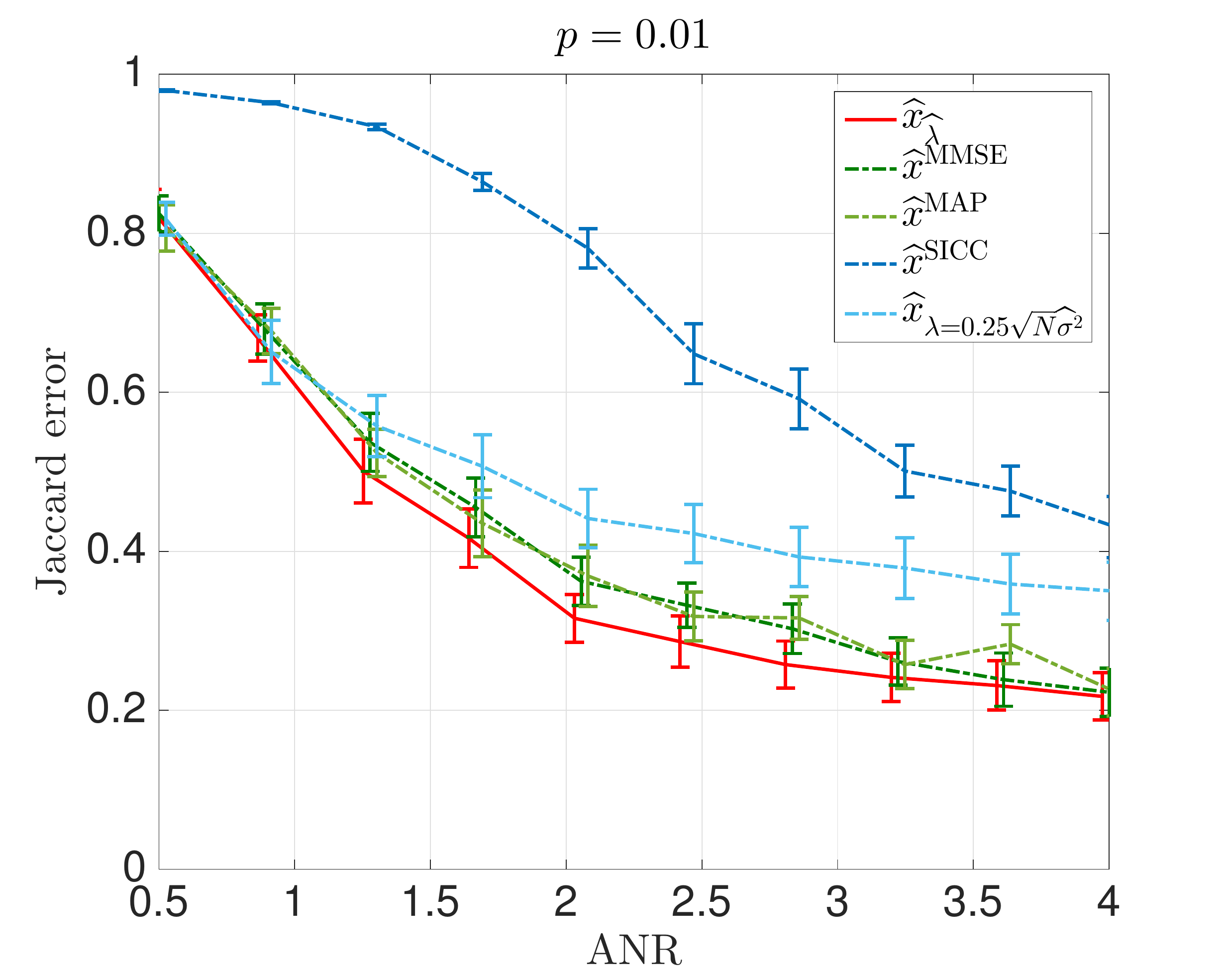}
\includegraphics[width=.24\linewidth]{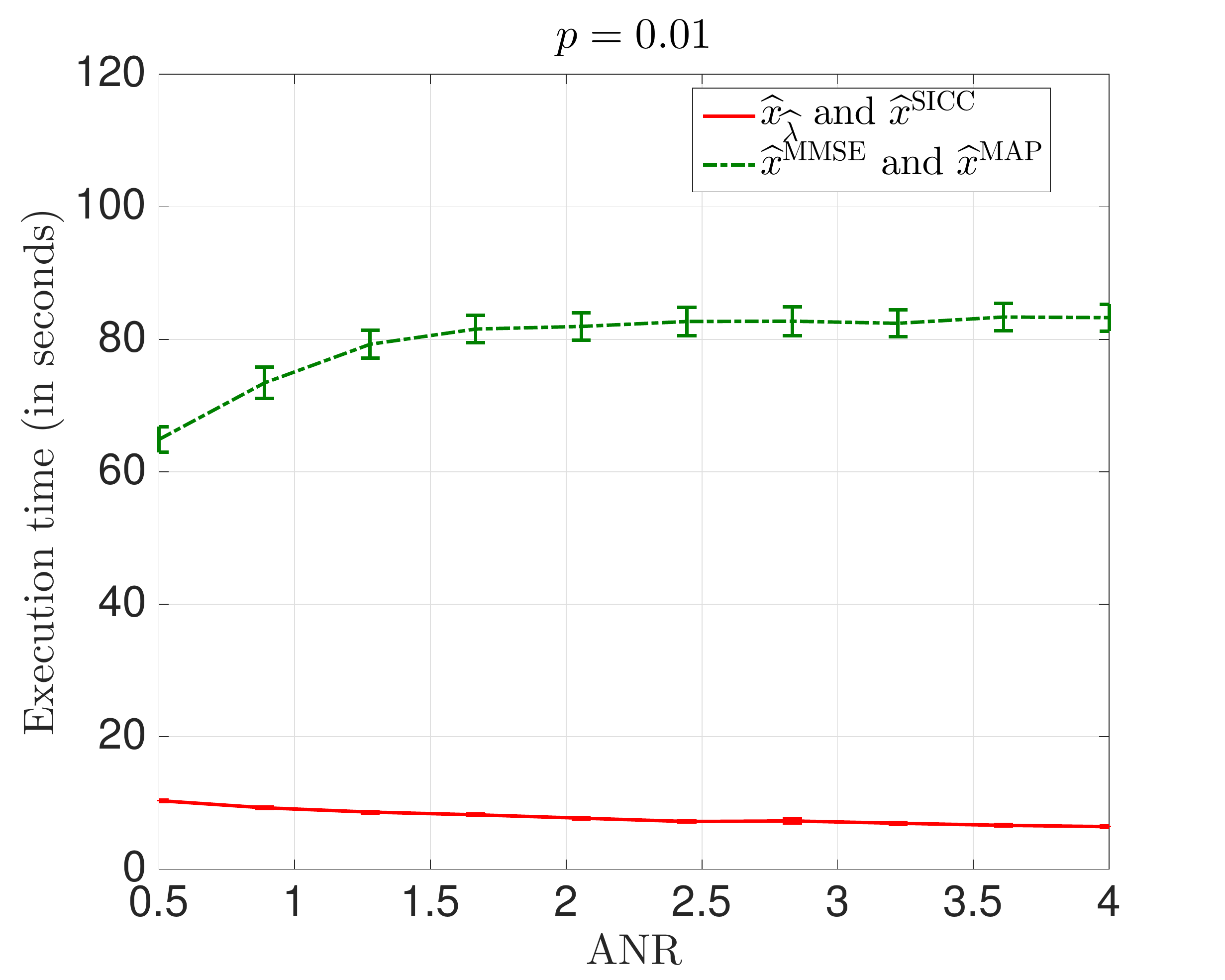}
\includegraphics[width=.24\linewidth]{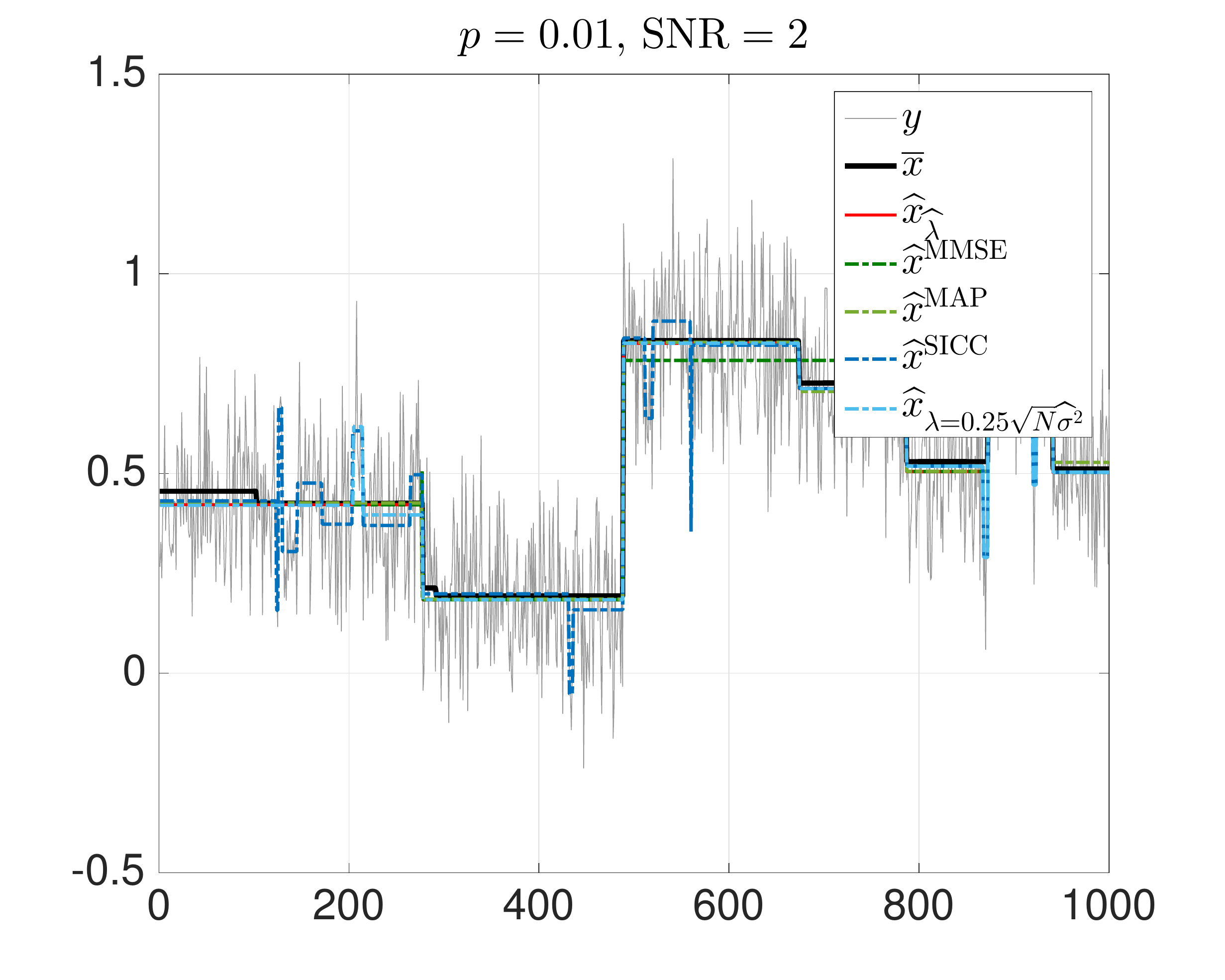}\\
\includegraphics[width=.24\linewidth]{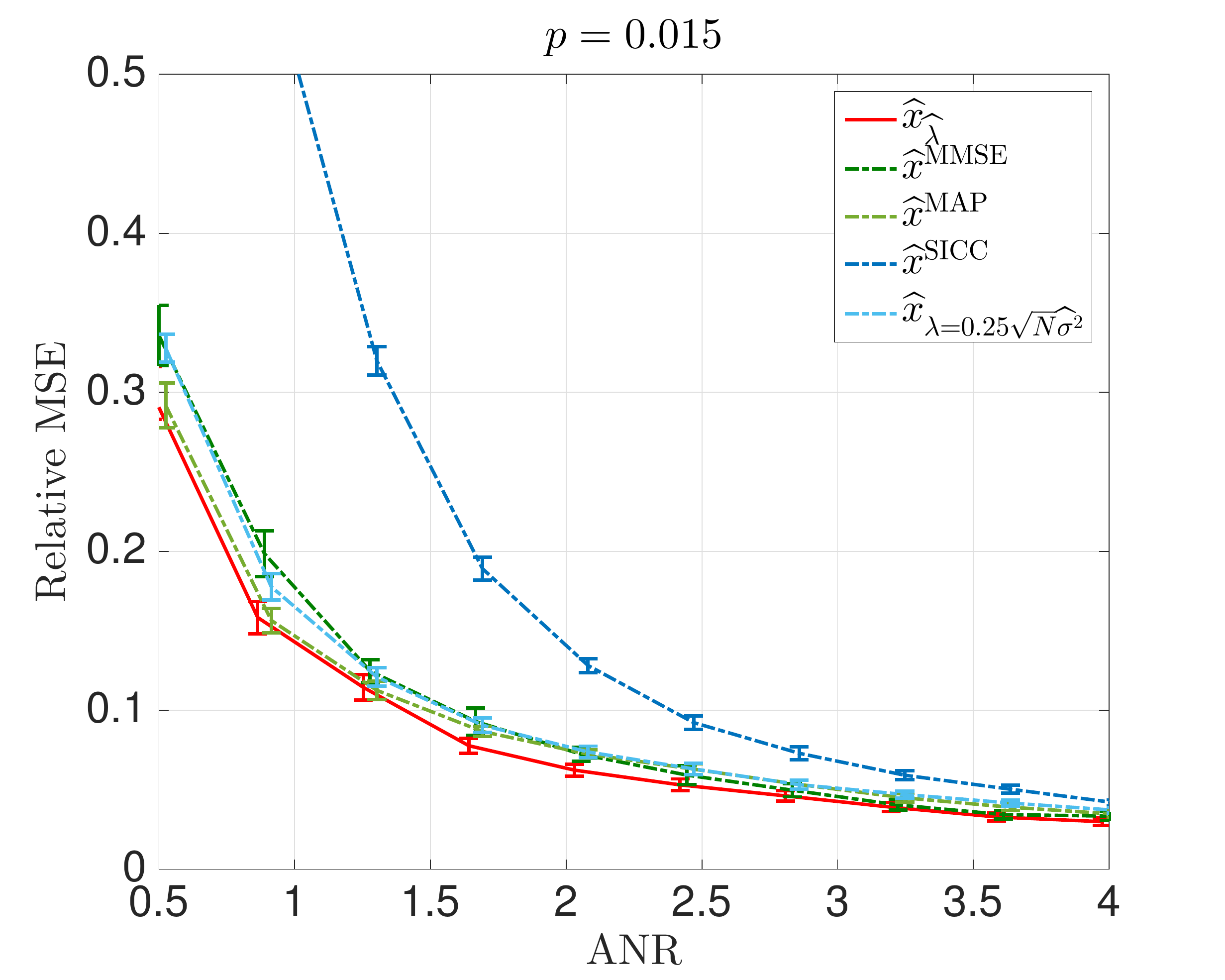}
\includegraphics[width=.24\linewidth]{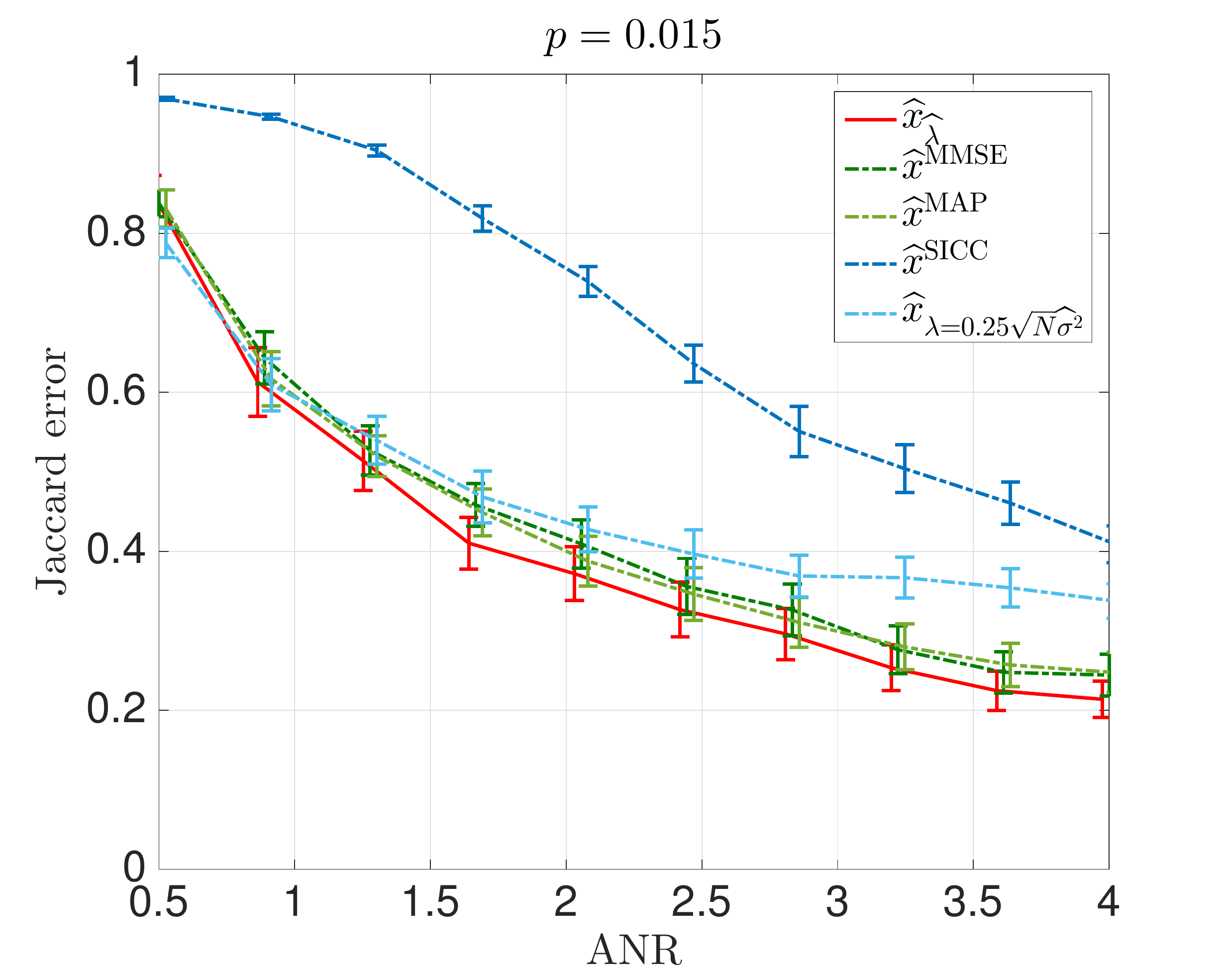}
\includegraphics[width=.24\linewidth]{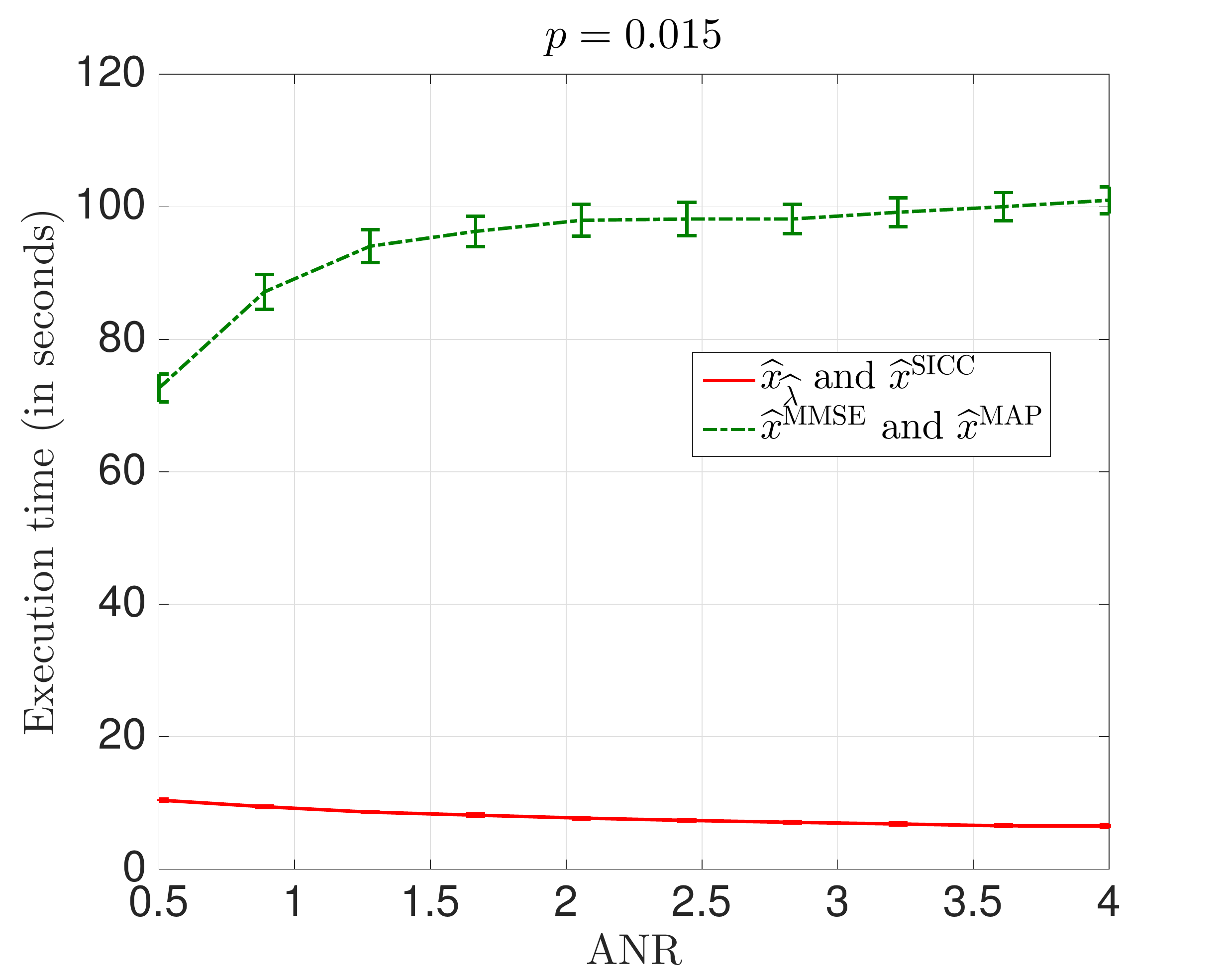}
\includegraphics[width=.24\linewidth]{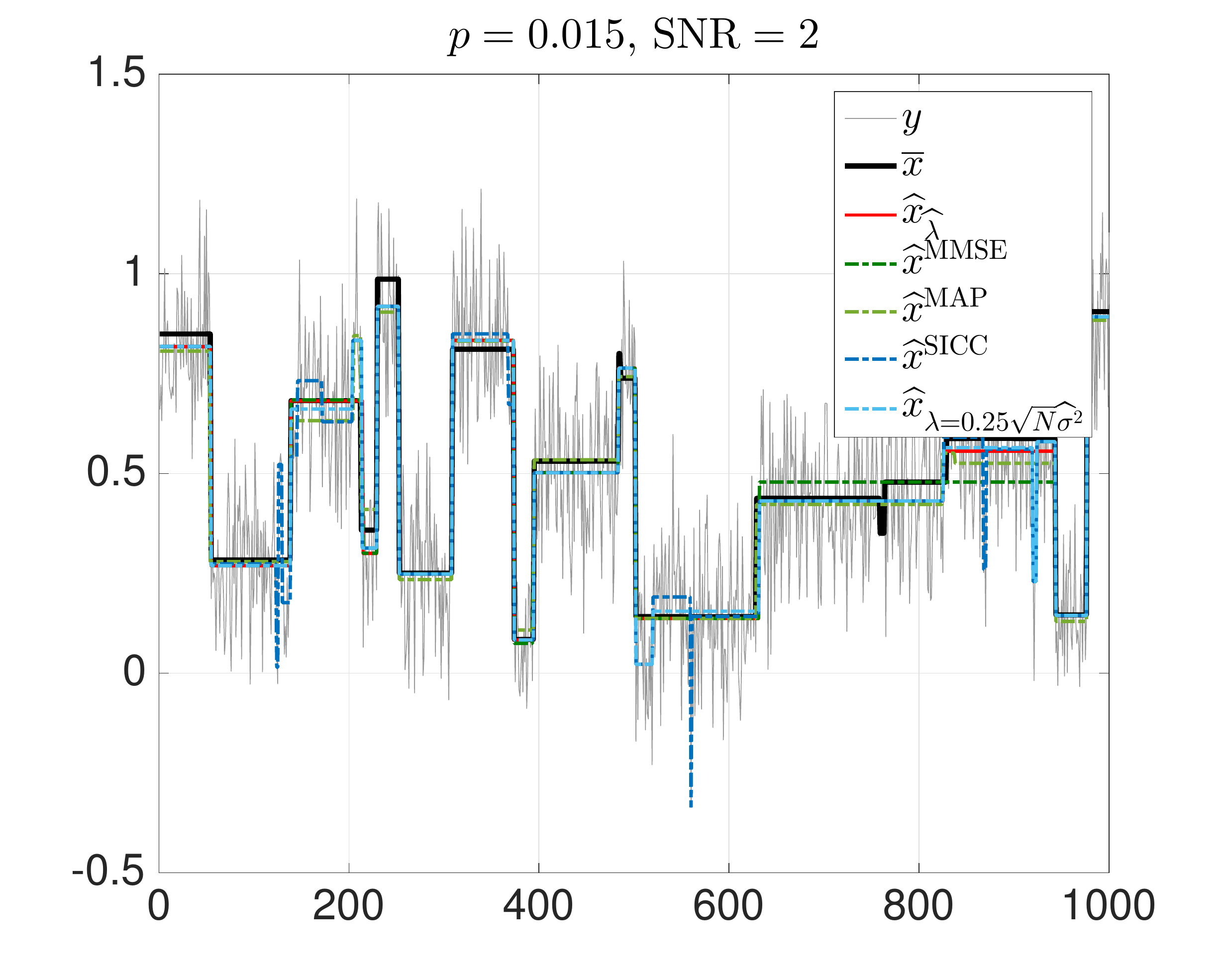}
 \caption{\textbf{Comparison with state-of-the-art methods.} For each configuration $\rm ANR=2$, $\overline{x}_{\max}-\overline{x}_{\min}=1$ and from top to bottom: $p=0.005$, $0.010$ and $0.015$. From left to right: relative MSE, Jaccard error, execution time and example of solutions. The proposed estimator (red) yields estimation performance comparable to Bayesian estimators (green) while benefiting from significantly lower computational costs. Moreover, it improves significantly the performance compared to SICC estimator (blue) for a similar computation cost. The results obtained with $\lambda = 0.25 \sqrt{N} \widehat{\sigma}^2$ is displayed in light blue.} \label{fig:compBayes}
\end{figure*}

In Fig.~\ref{fig:compBayes}, estimation performance for the proposed procedure (solid red)  are compared against MAP and MMSE hierarchical Bayesian estimators, as functions of ANR. Overall, $\boldsymbol{\widehat{{\xvar}}}_{\widehat{\lambda}}$ is equivalent to $\widehat{\boldsymbol{\xvar}}_{\rm MAP}$ and $\widehat{\boldsymbol{\xvar}}_{\rm MMSE}$  in terms of MSE (first column)  and Jaccard error (second column), while benefiting of significantly lower computational costs.
Interestingly, when $p$ increases (large number of change-points), the larger the gain in using the proposed procedure.
This is also the case when the sample size $N$ increases: For $N=10^4$, the MCMC approach takes more than an hour while the method we propose here provides a relevant solution in a few minutes.\\
We further compared the performance of the proposed strategy with several classical parameter choice based on information criteria such as AIC, SIC, AICC and SICC (see \cite{Winkler_G_2005_j-dmv_dont_stb} for details about these criteria). Fig.~\ref{fig:compBayes} (solid blue) only reports the SICC-based solutions, which performed best among those four criteria. The proposed solution always perform better. Further criteria can be encountered in \cite{Zhang_N_2007_biometrics_mod_bic,Frick_K_2014_j-rsssb_mul_cpi,Xia_Z_2015_biometrika_jum_ics}.
Finally, we follow the regularization parameter choice provided in \cite{Selesnick_I_2015_j-ieee-spl_con_1tv} consisting in the heuristic rule  $\tau=0.25\sqrt{N}\sigma$ derived in \cite{Dumbgen_L_2009_j-ejs_ext_sts} for the $\ell_1$-penalized formulation. Invoking dimensionality arguments, this choice as been adapted to $\lambda = 0.25 \sqrt{N} \sigma^2$ for the $\ell_0$-penalized formulation addressed in this work. One should note that this parameter selection method relies on the perfect knowledge of the noise variance $\sigma^2$, which is not the case in the considered study framework. To provide fair comparisons, the performance of this heuristic rule has been evaluated using an estimate of this variance, derived from the classical median estimator \cite{Donoho_D_1995_j-asa_ada_us} which is particularly suitable for piecewise constant signals. Again, the proposed method always lead to better solutions.

\subsection{Selection of hyperparameter $\sigma_0^2$} \label{ss:sigma0}

\begin{figure*}
\begin{tabular}{@{}c@{}c@{}cc@{}c@{}c@{}}
\includegraphics[width=.162\linewidth,clip=true,trim=.3cm 0cm .5cm .5cm]{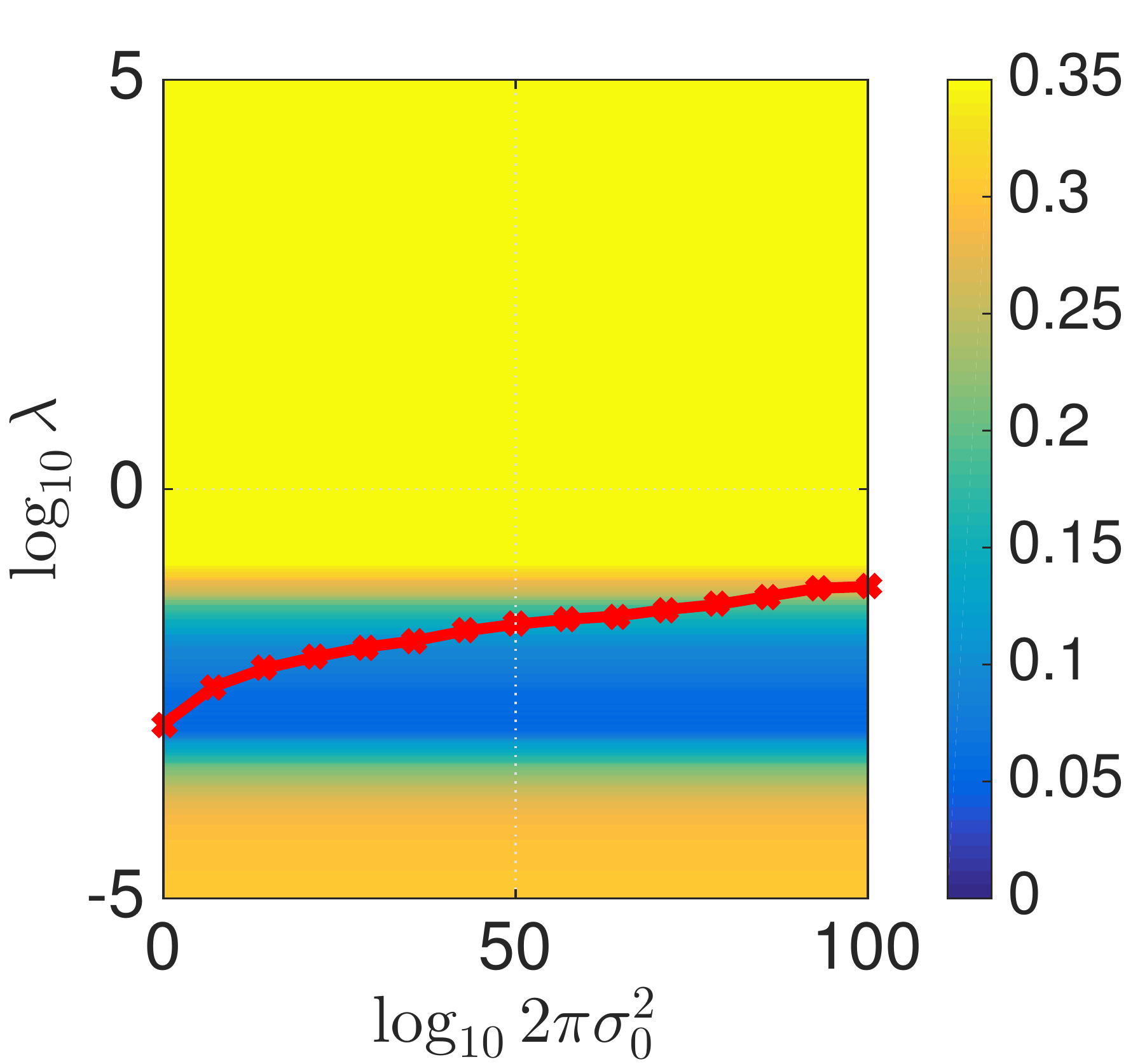}&
\includegraphics[width=.162\linewidth,clip=true,trim=.3cm 0cm .5cm .5cm]{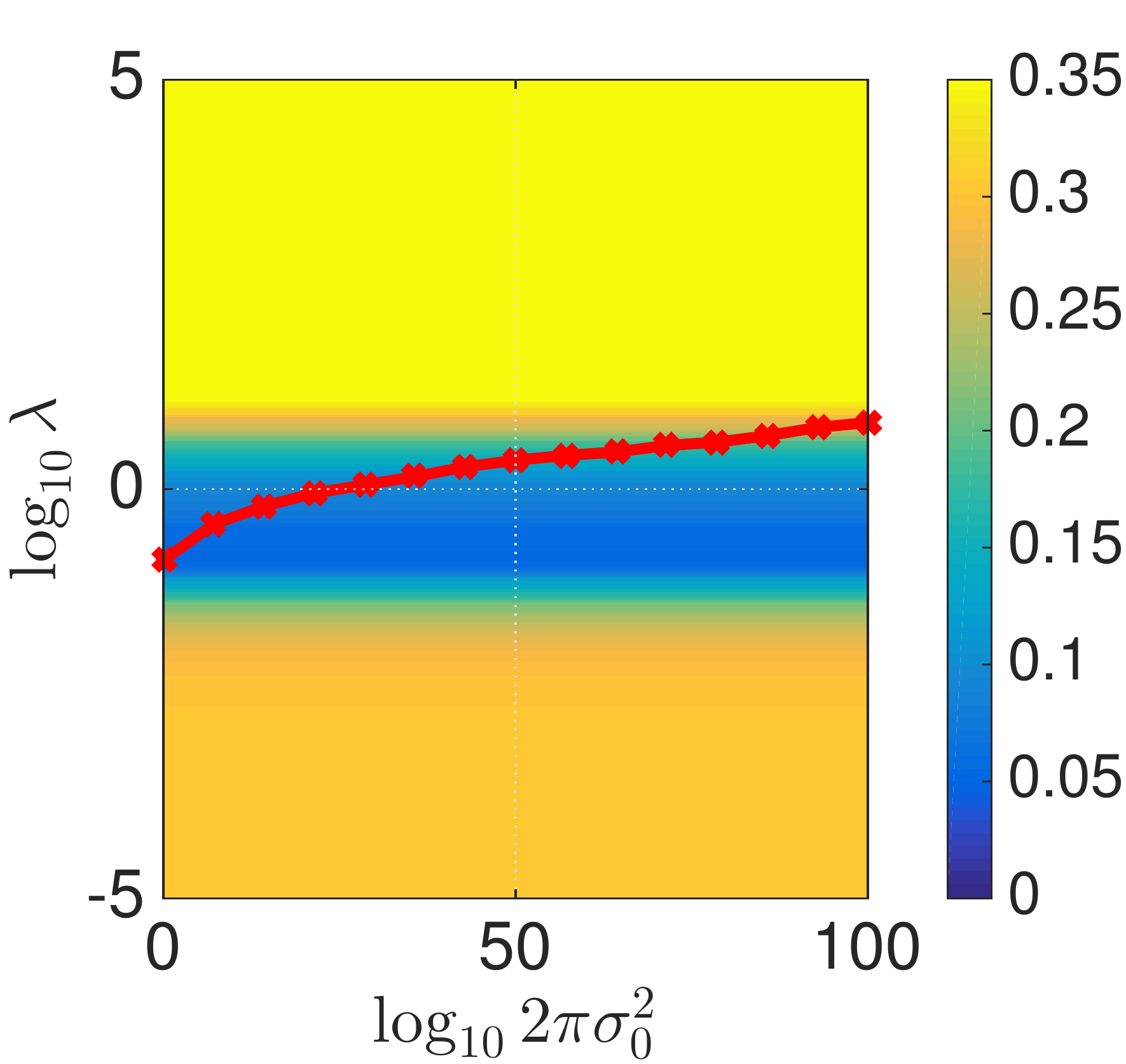}&
\includegraphics[width=.162\linewidth,clip=true,trim=.3cm 0cm .5cm .5cm]{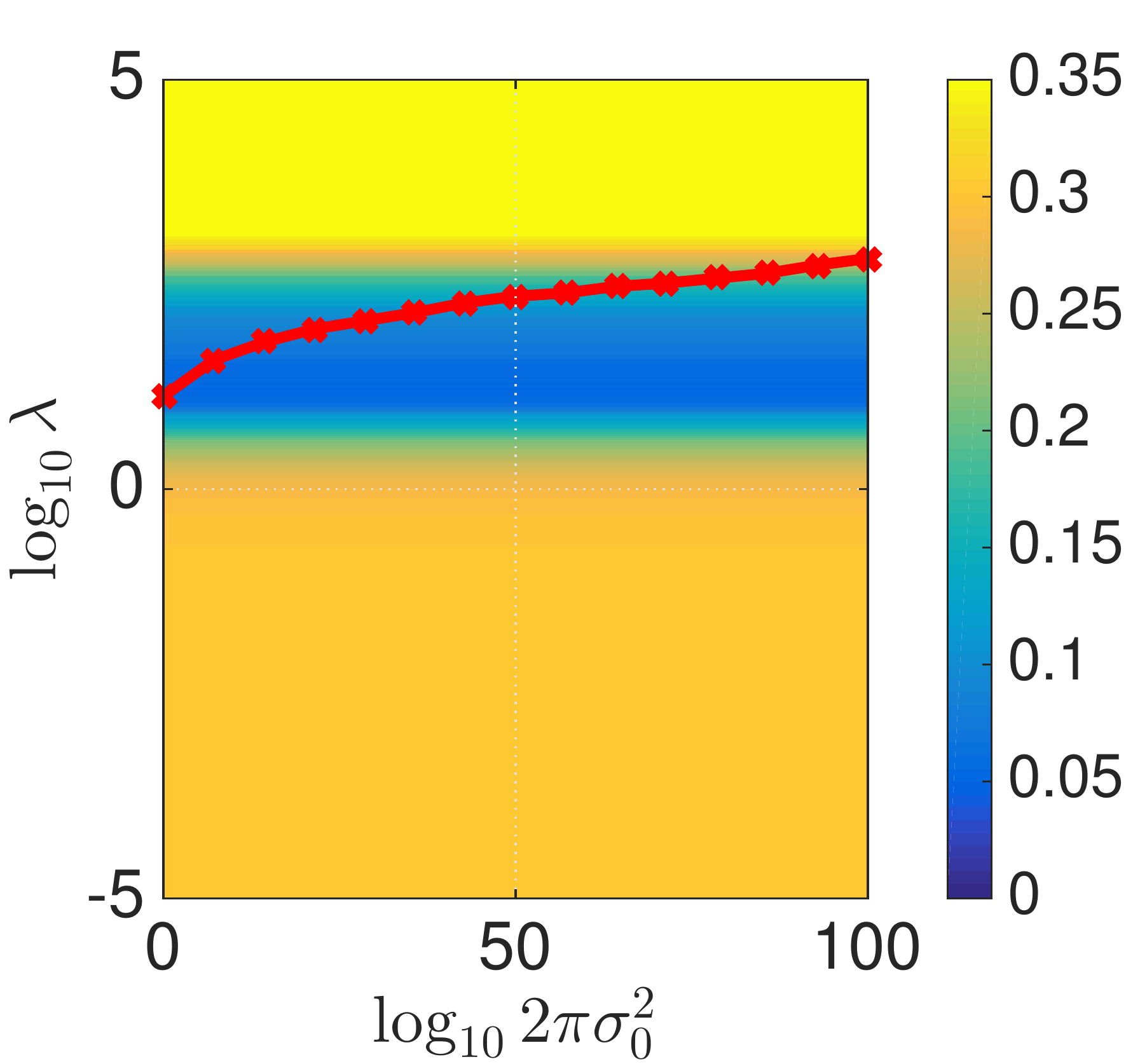}&
\includegraphics[width=.162\linewidth,clip=true,trim=.3cm 0cm .5cm .5cm]{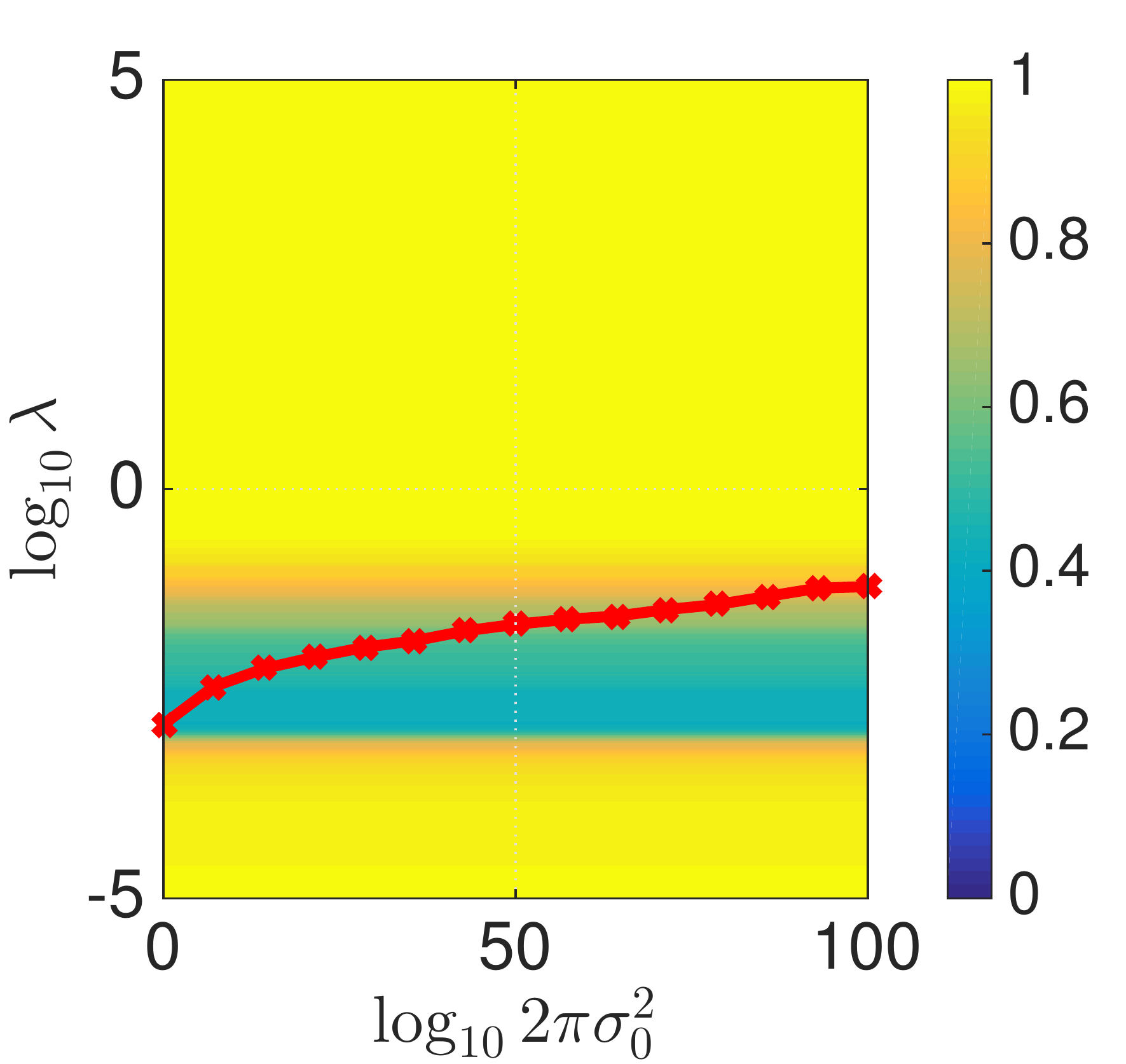}&
\includegraphics[width=.162\linewidth,clip=true,trim=.3cm 0cm .5cm .5cm]{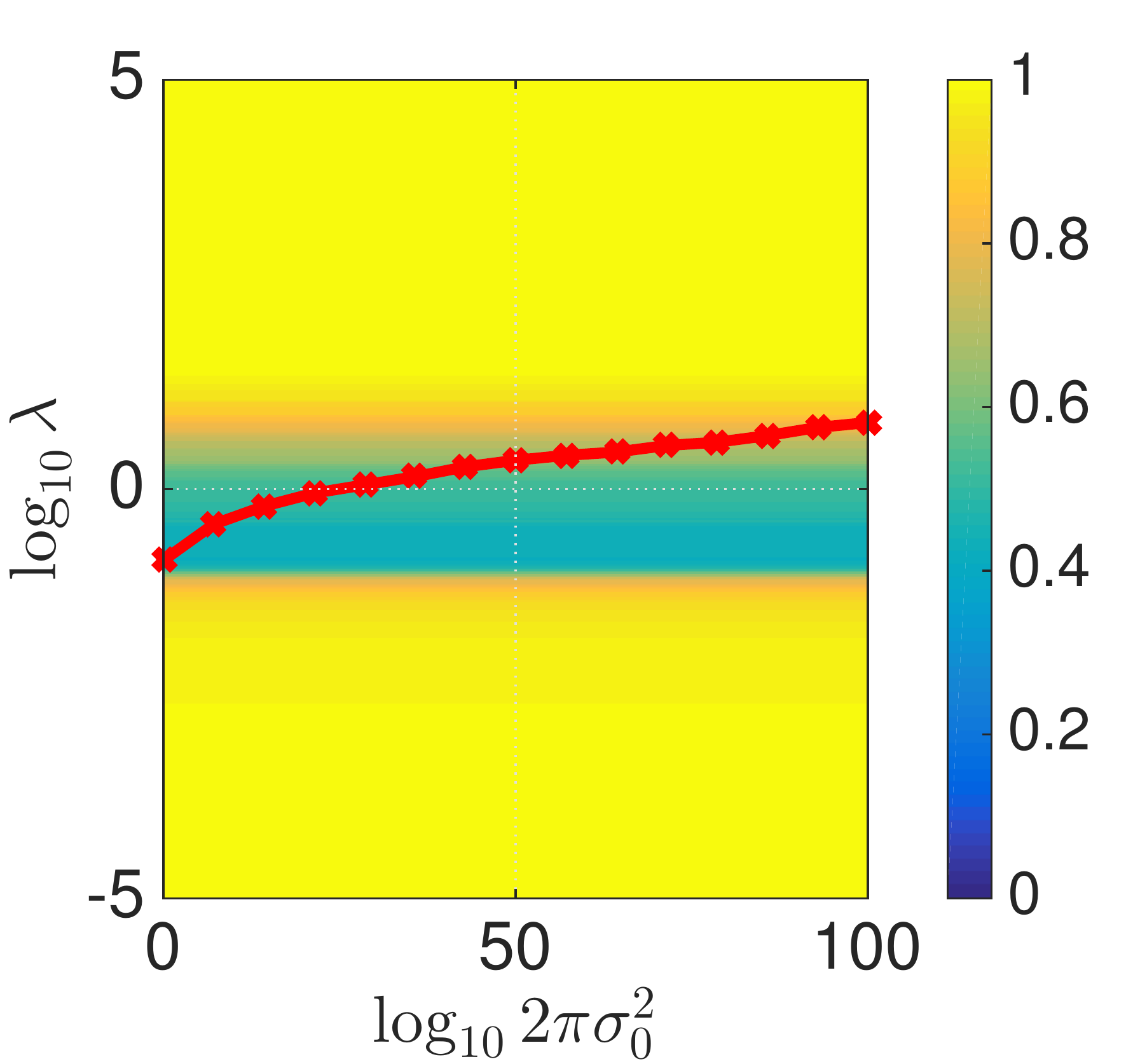}&
\includegraphics[width=.162\linewidth,clip=true,trim=.3cm 0cm .5cm .5cm]{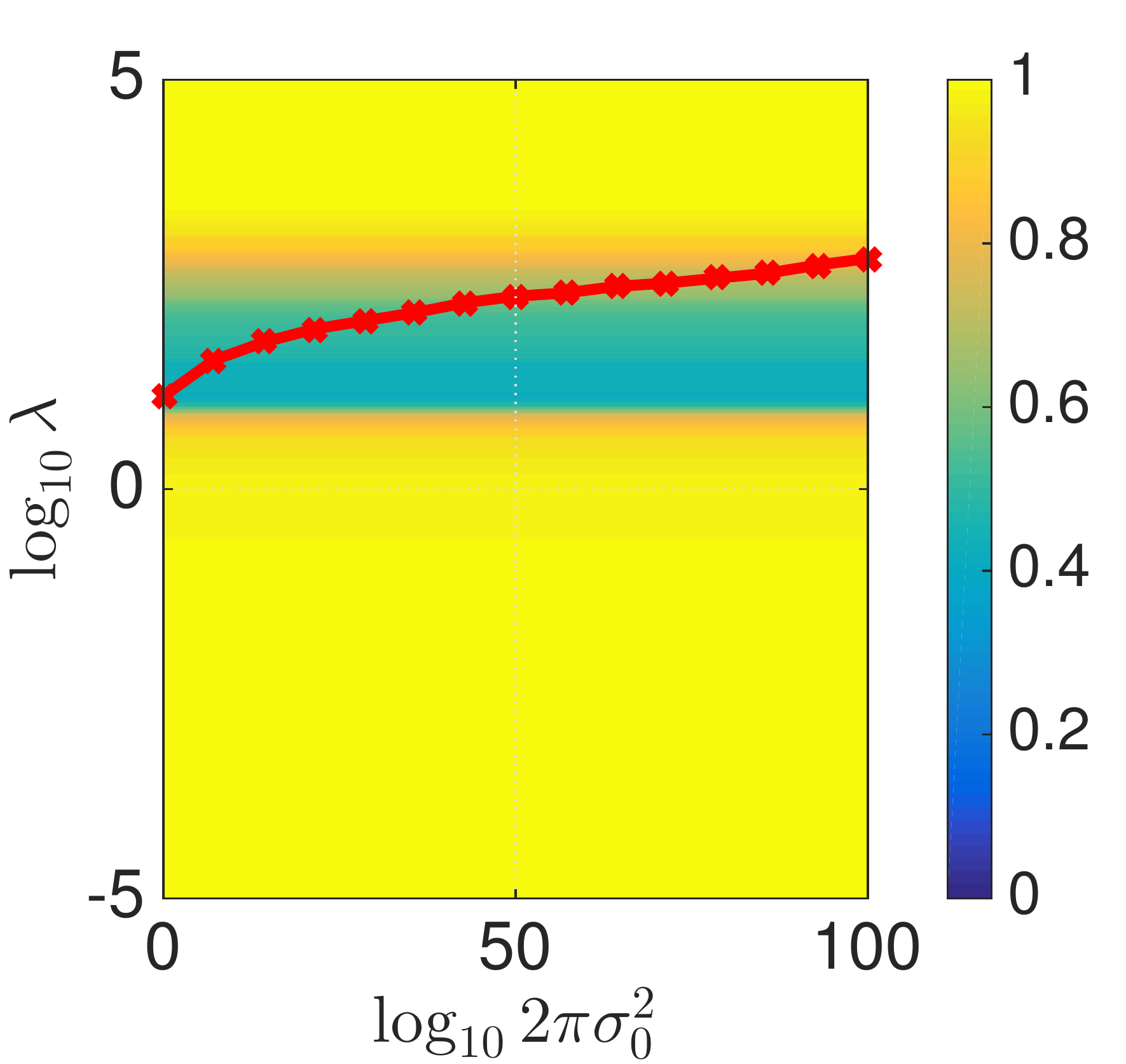}\\
\multicolumn{3}{c}{(a) relative MSE} &\multicolumn{3}{c}{(b) Jaccard error}
\end{tabular}
 \caption{\textbf{Estimation performance: RMSE and Jaccard error as functions of ${\lambda}$ and $\sigma_0^2$.} 
The background displays the relative MSE (left) or Jaccard error (right) w.r.t. $\lambda$ and $\sigma_0^2$. We superimpose in red the estimate  $\widehat{\lambda}$, which a priori explicitly depends on the choice of the hyperprior $\sigma_0^2$, is averaged over $50$ realizations and displayed in red as a function of $2\pi\sigma_0^2$.
From left to right: $\overline{x}_{\max}-\overline{x}_{\min}=0.1$, $1$ and $10$.
For each configuration $p=0.01$, $\rm ANR=1$.
Choosing $\log 2 \pi\sigma_0^2 \in [0,5]$ systematically leads to satisfactory estimates minimizing the relative MSE (left) or Jaccard error (right).
\label{fig:sigma0dyn}}
\end{figure*}

\begin{figure*}
\begin{tabular}{@{}c@{}c@{}cc@{}c@{}c@{}}
\includegraphics[width=.162\linewidth,clip=true,trim=.3cm 0cm .5cm .5cm]{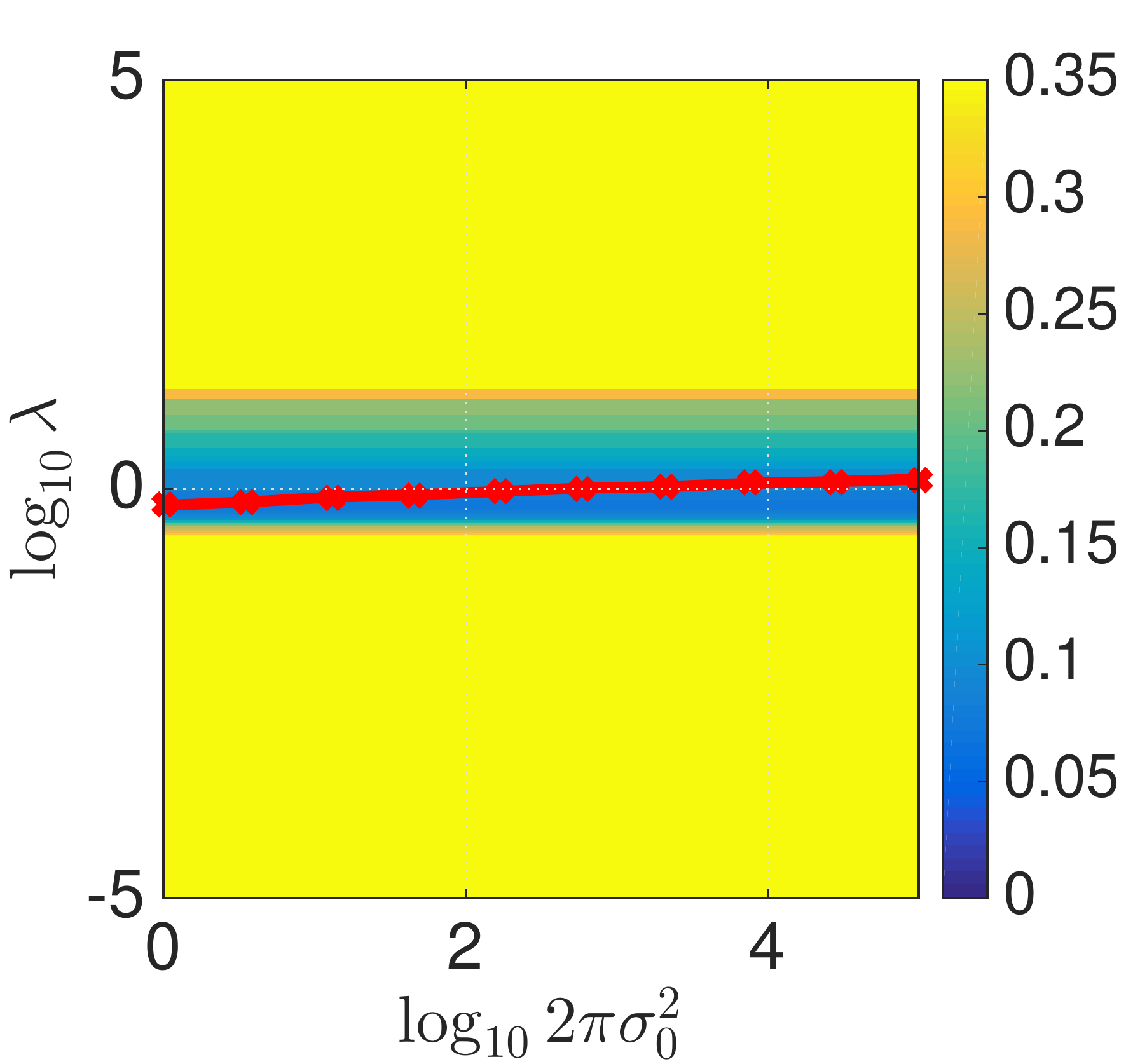}&
\includegraphics[width=.162\linewidth,clip=true,trim=.3cm 0cm .5cm .5cm]{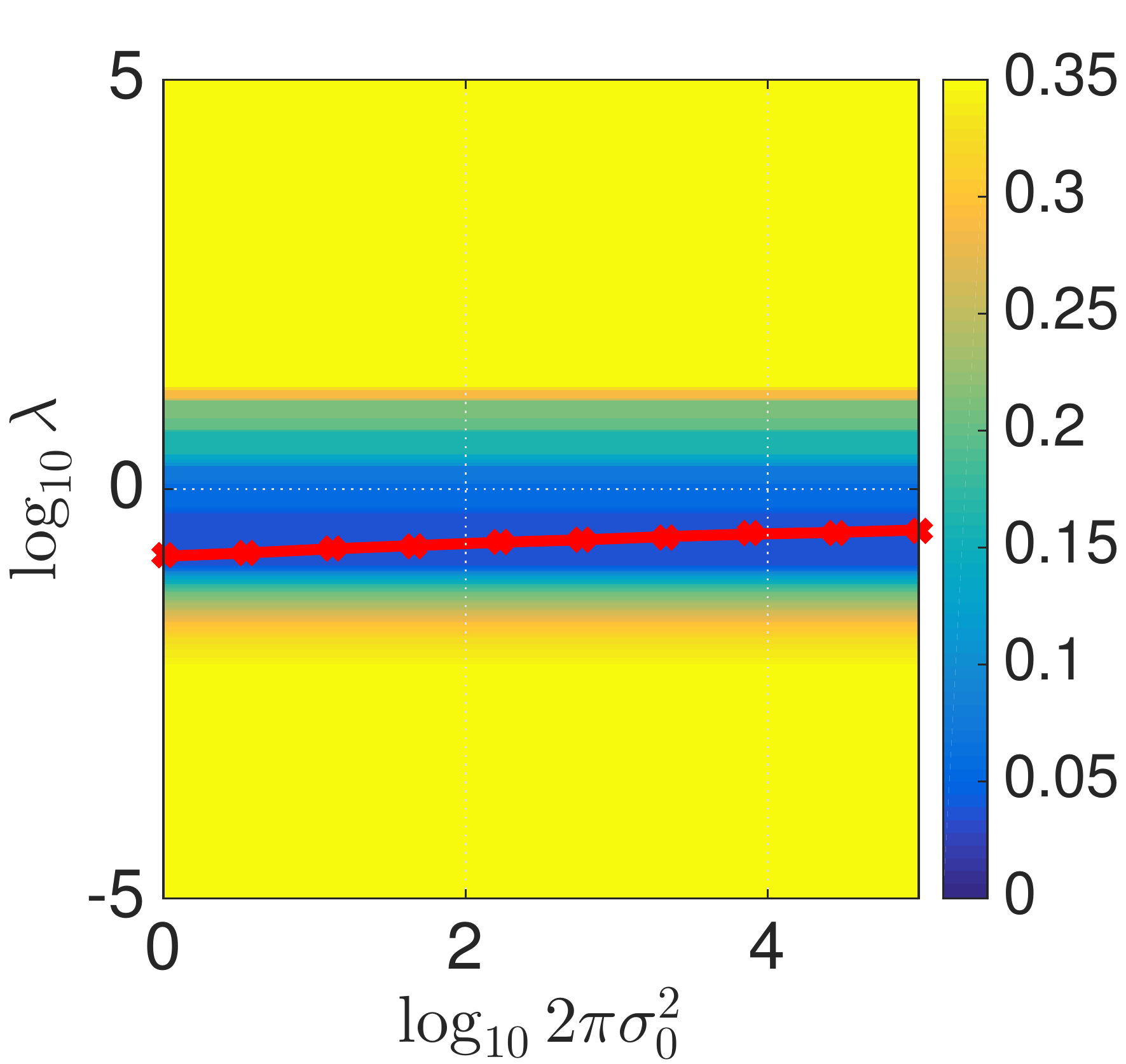}&
\includegraphics[width=.162\linewidth,clip=true,trim=.3cm 0cm .5cm .5cm]{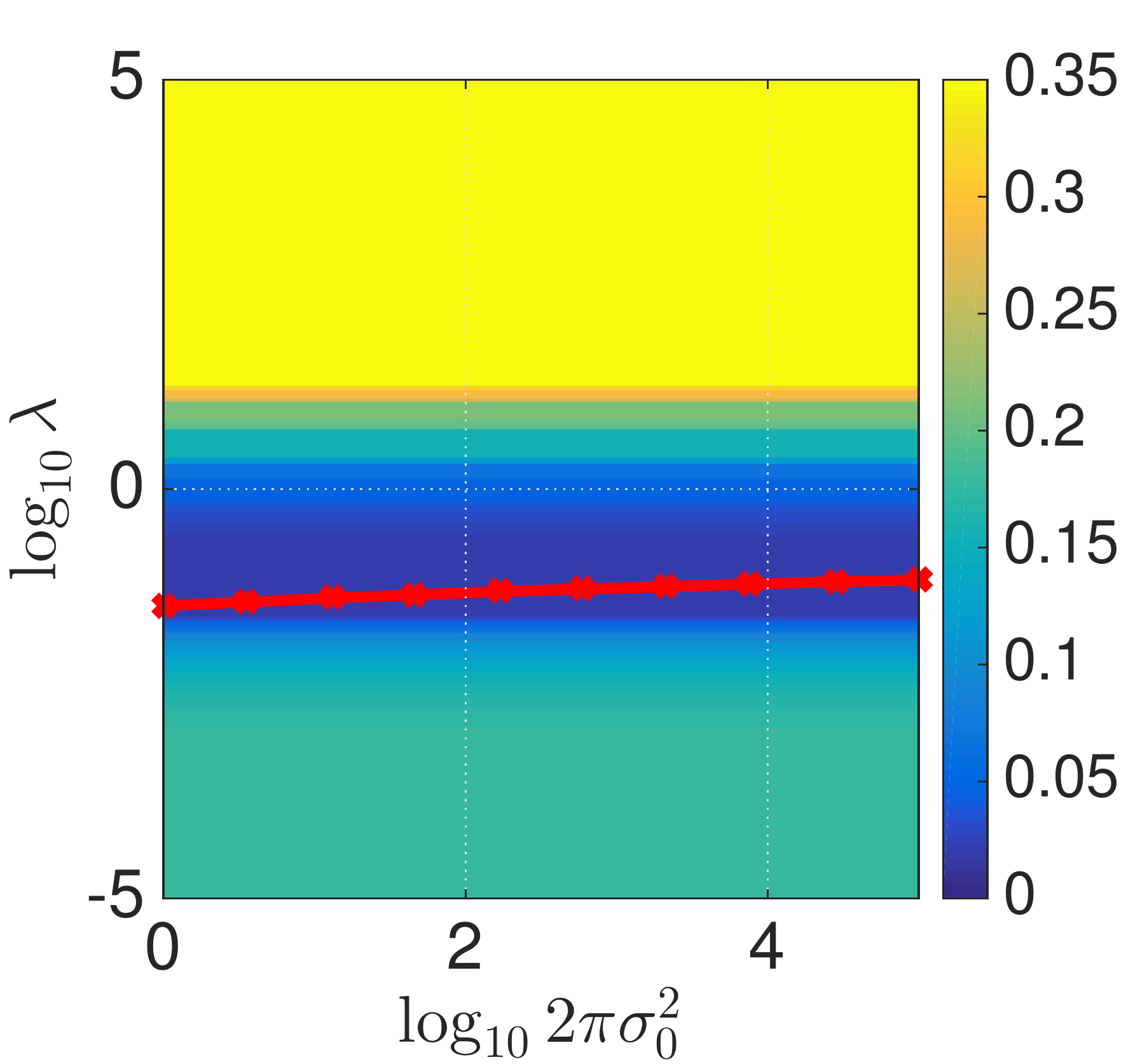}&
\includegraphics[width=.162\linewidth,clip=true,trim=.3cm 0cm .5cm .5cm]{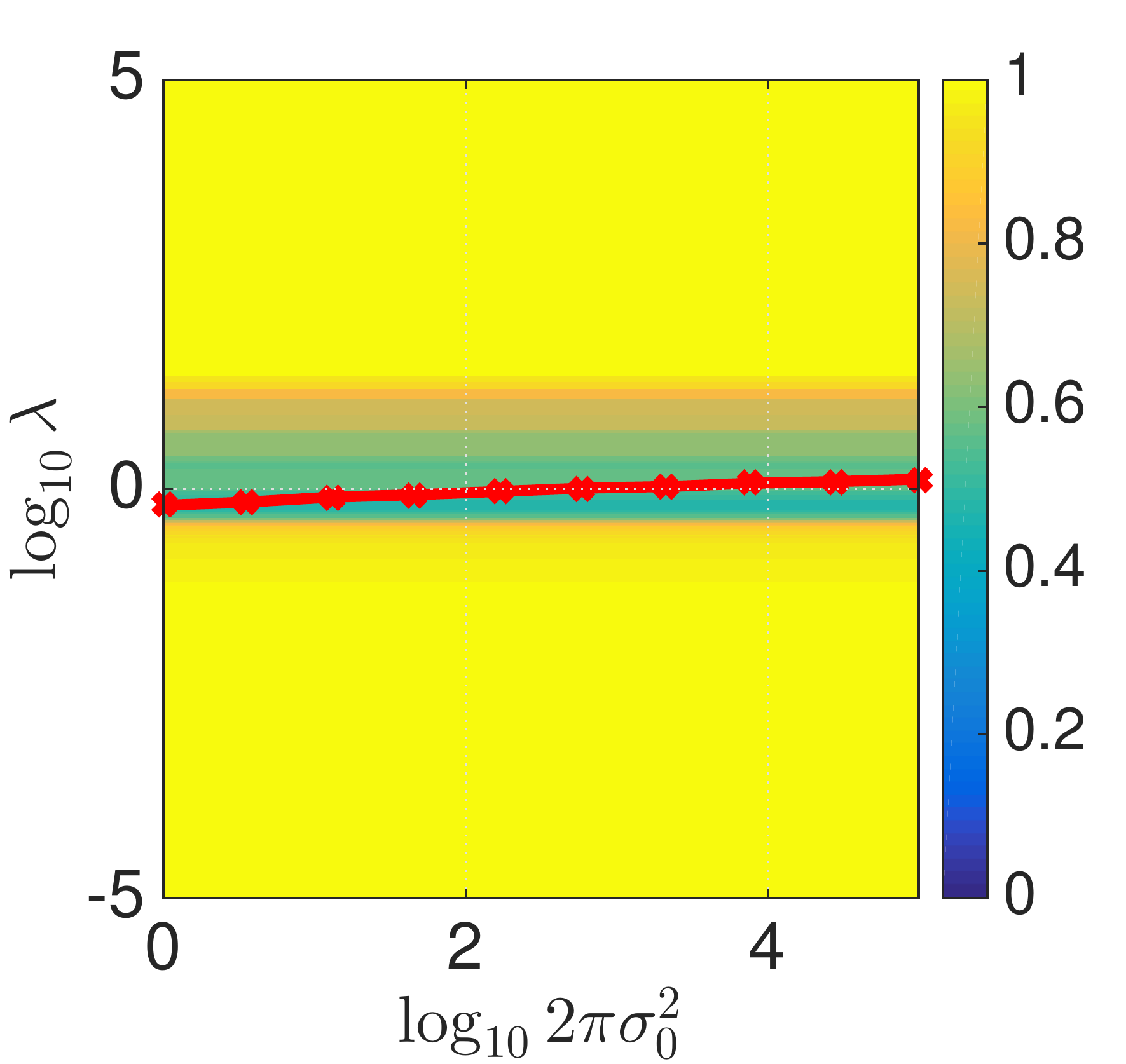}&
\includegraphics[width=.162\linewidth,clip=true,trim=.3cm 0cm .5cm .5cm]{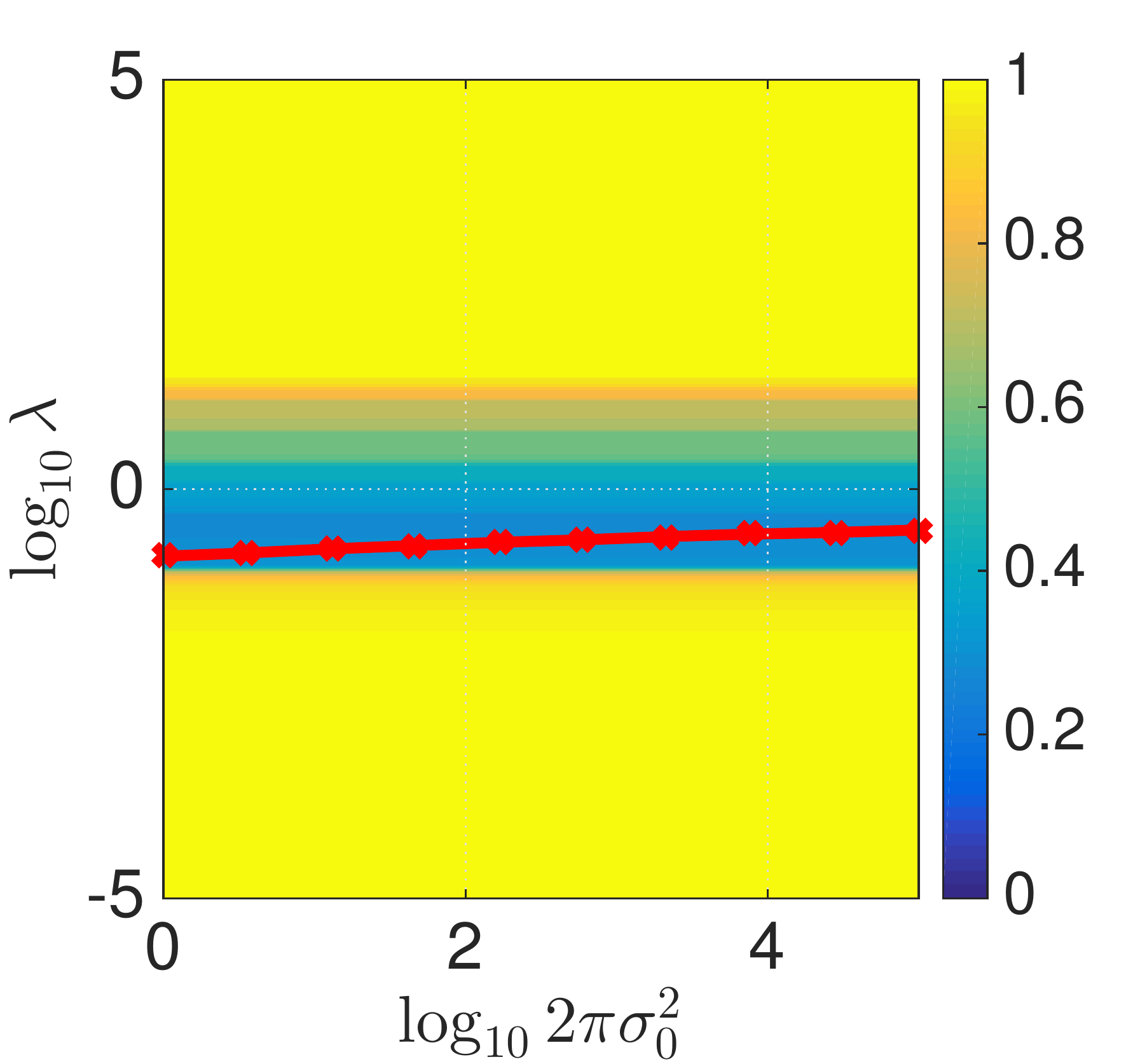}&
\includegraphics[width=.162\linewidth,clip=true,trim=.3cm 0cm .5cm .5cm]{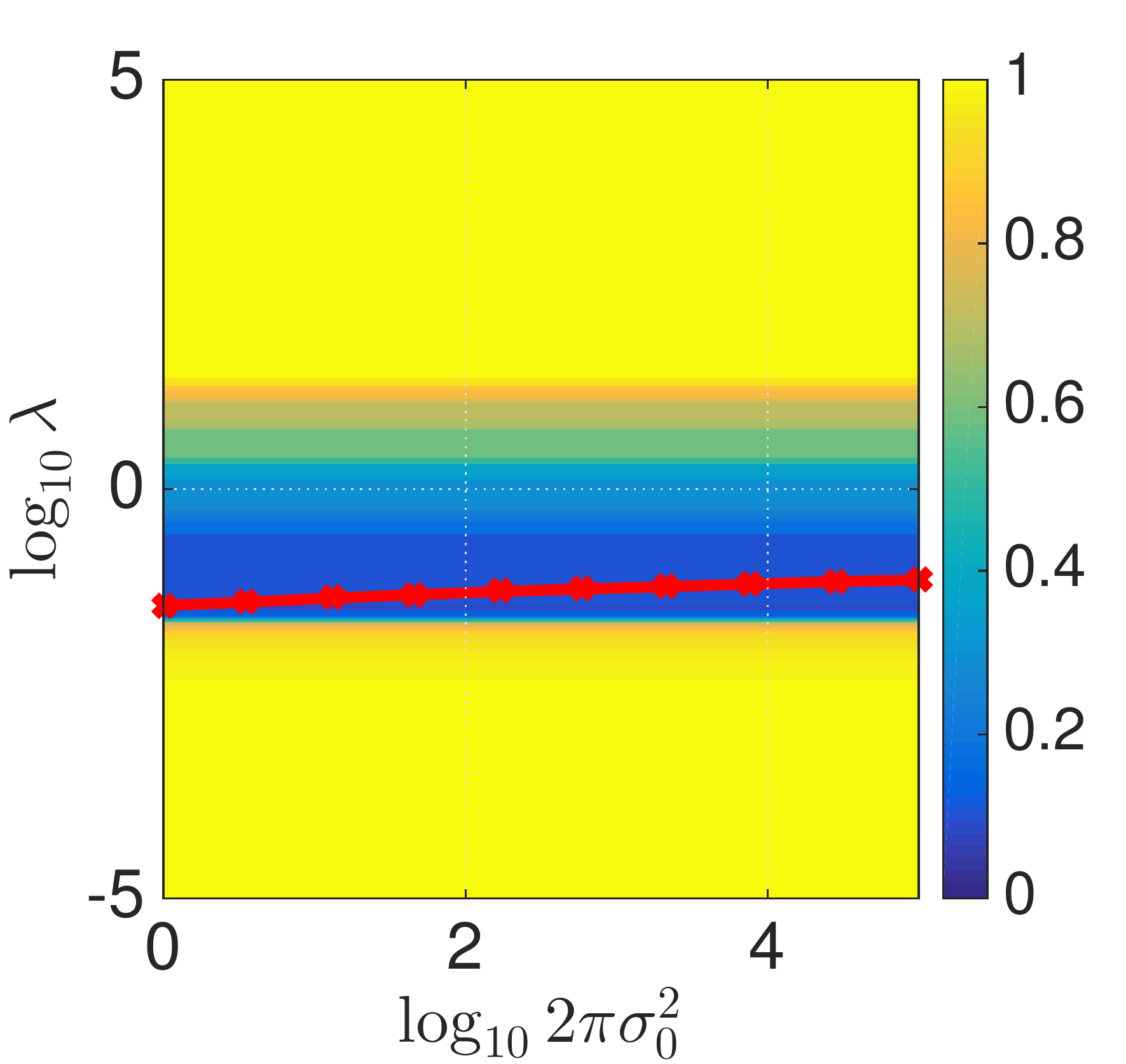}\\
\includegraphics[width=.162\linewidth,clip=true,trim=.3cm 0cm .5cm .5cm]{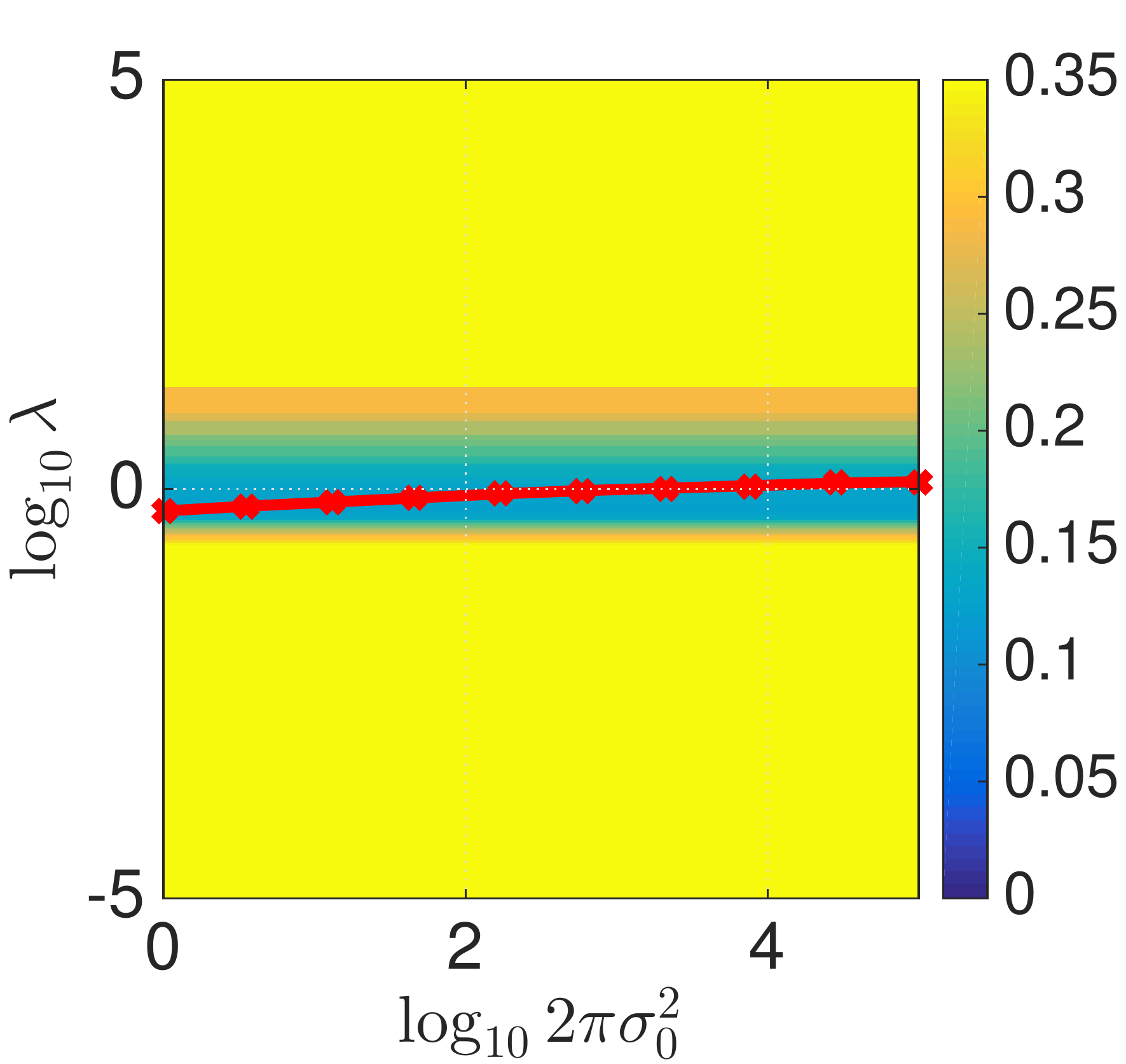}&
\includegraphics[width=.162\linewidth,clip=true,trim=.3cm 0cm .5cm .5cm]{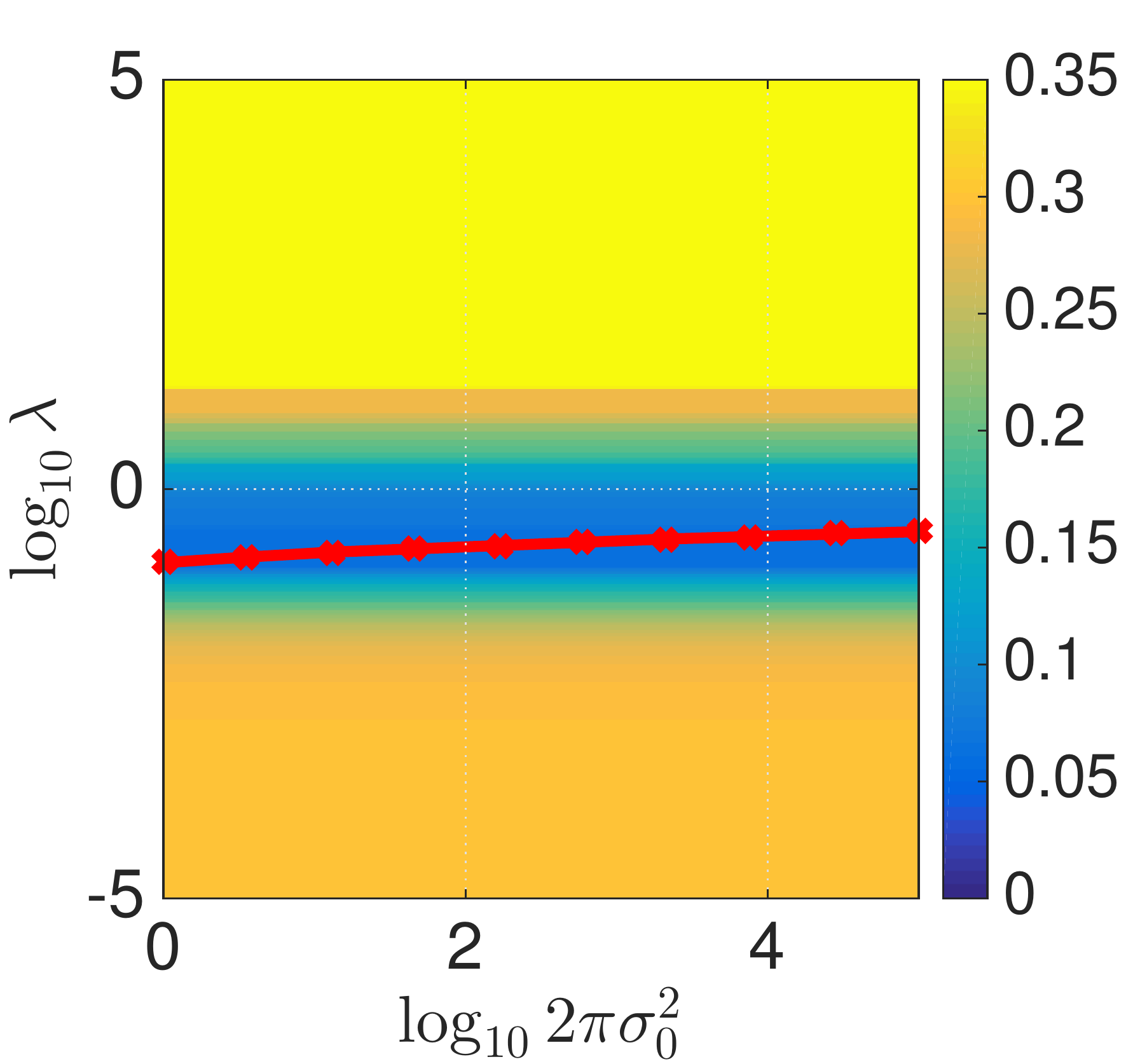}&
\includegraphics[width=.162\linewidth,clip=true,trim=.3cm 0cm .5cm .5cm]{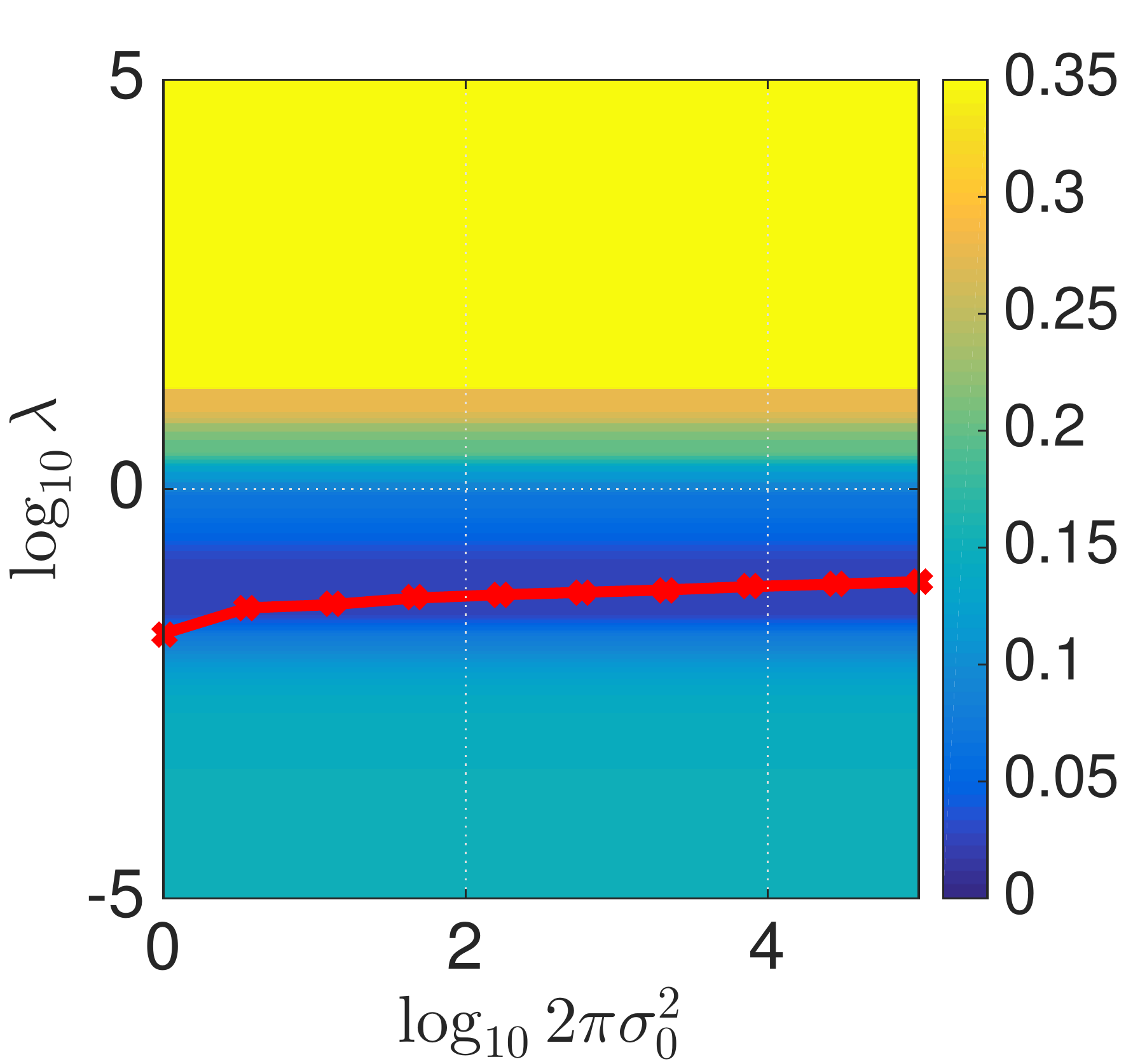}&
\includegraphics[width=.162\linewidth,clip=true,trim=.3cm 0cm .5cm .5cm]{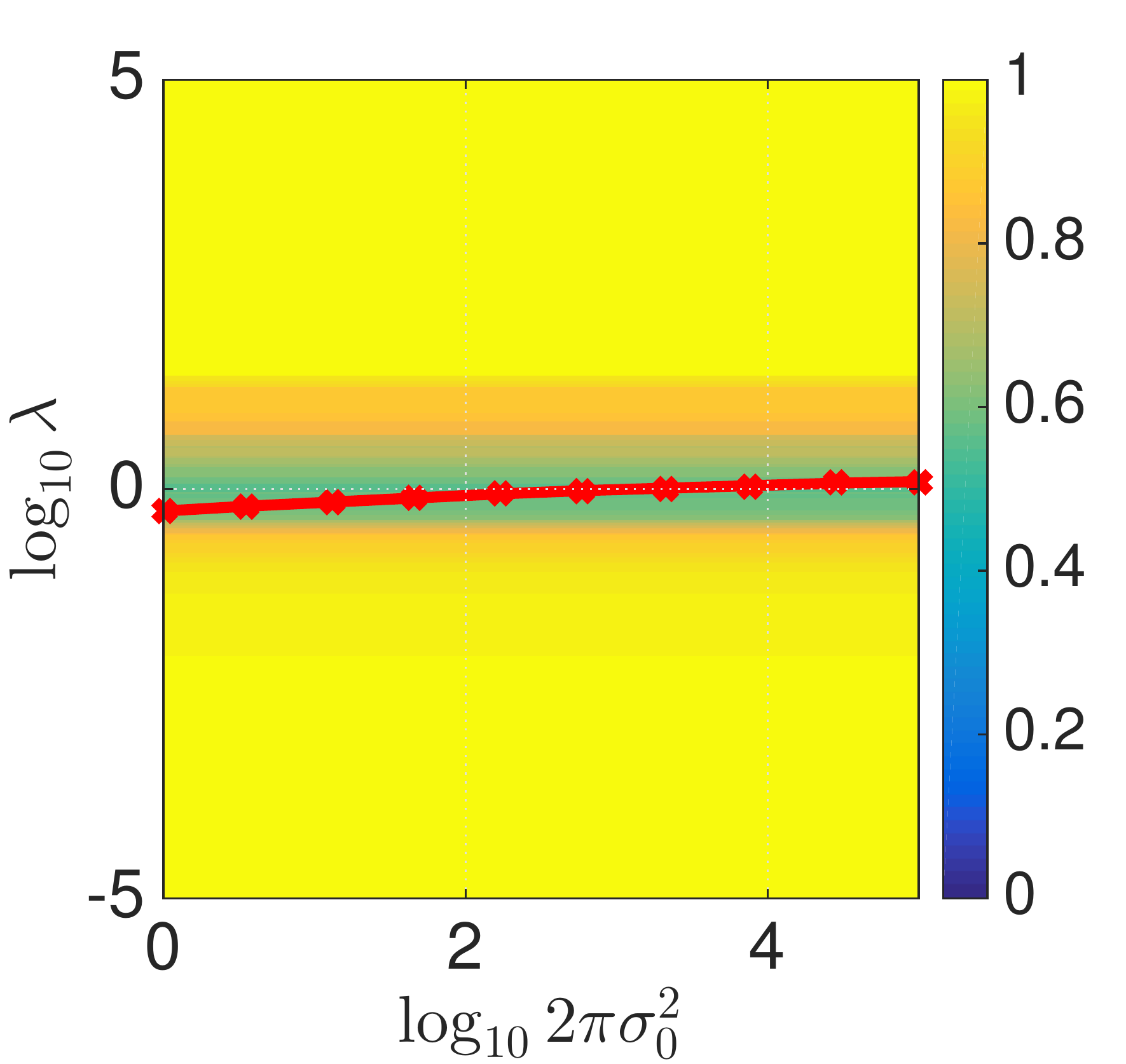}&
\includegraphics[width=.162\linewidth,clip=true,trim=.3cm 0cm .5cm .5cm]{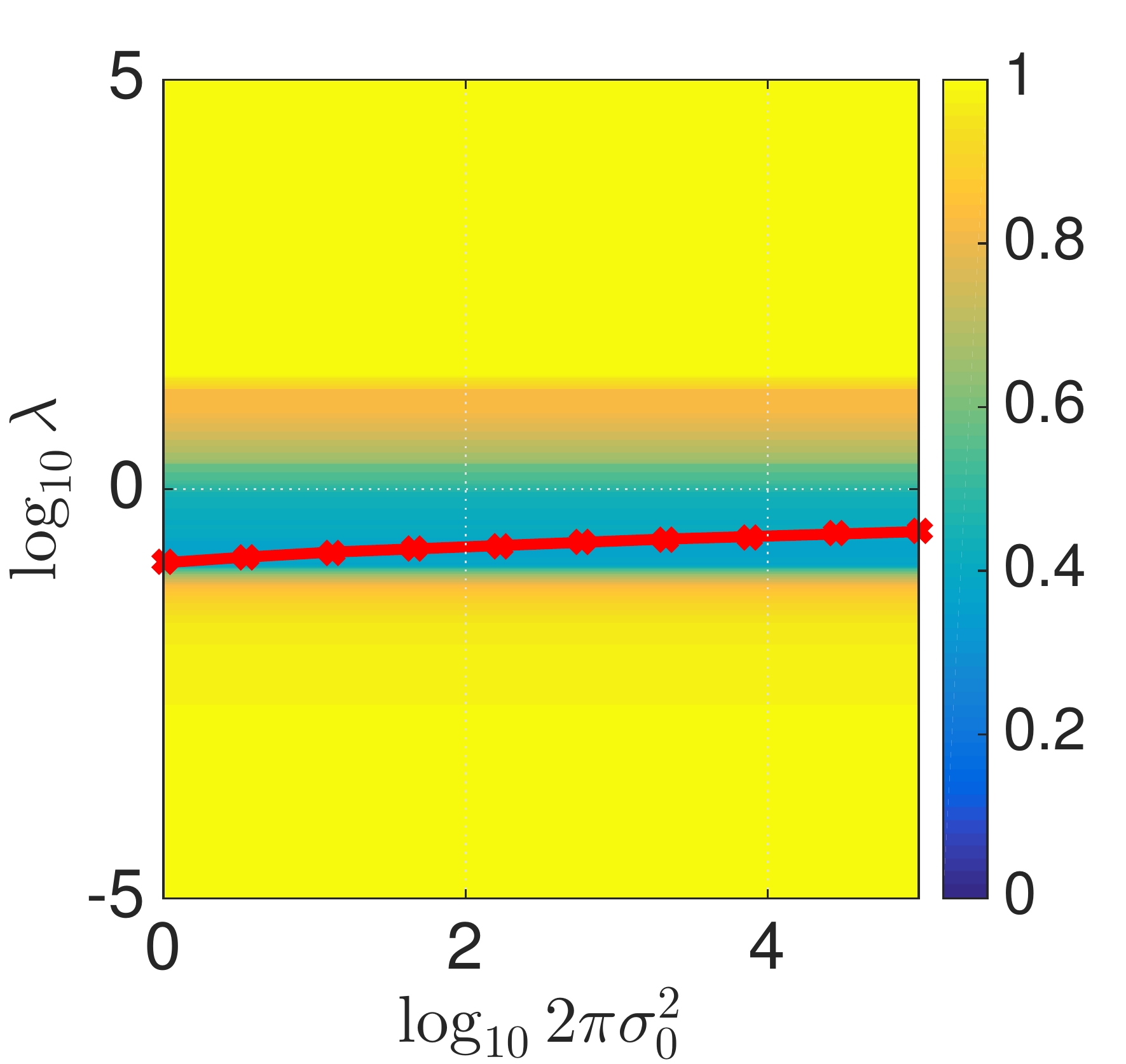}&
\includegraphics[width=.162\linewidth,clip=true,trim=.3cm 0cm .5cm .5cm]{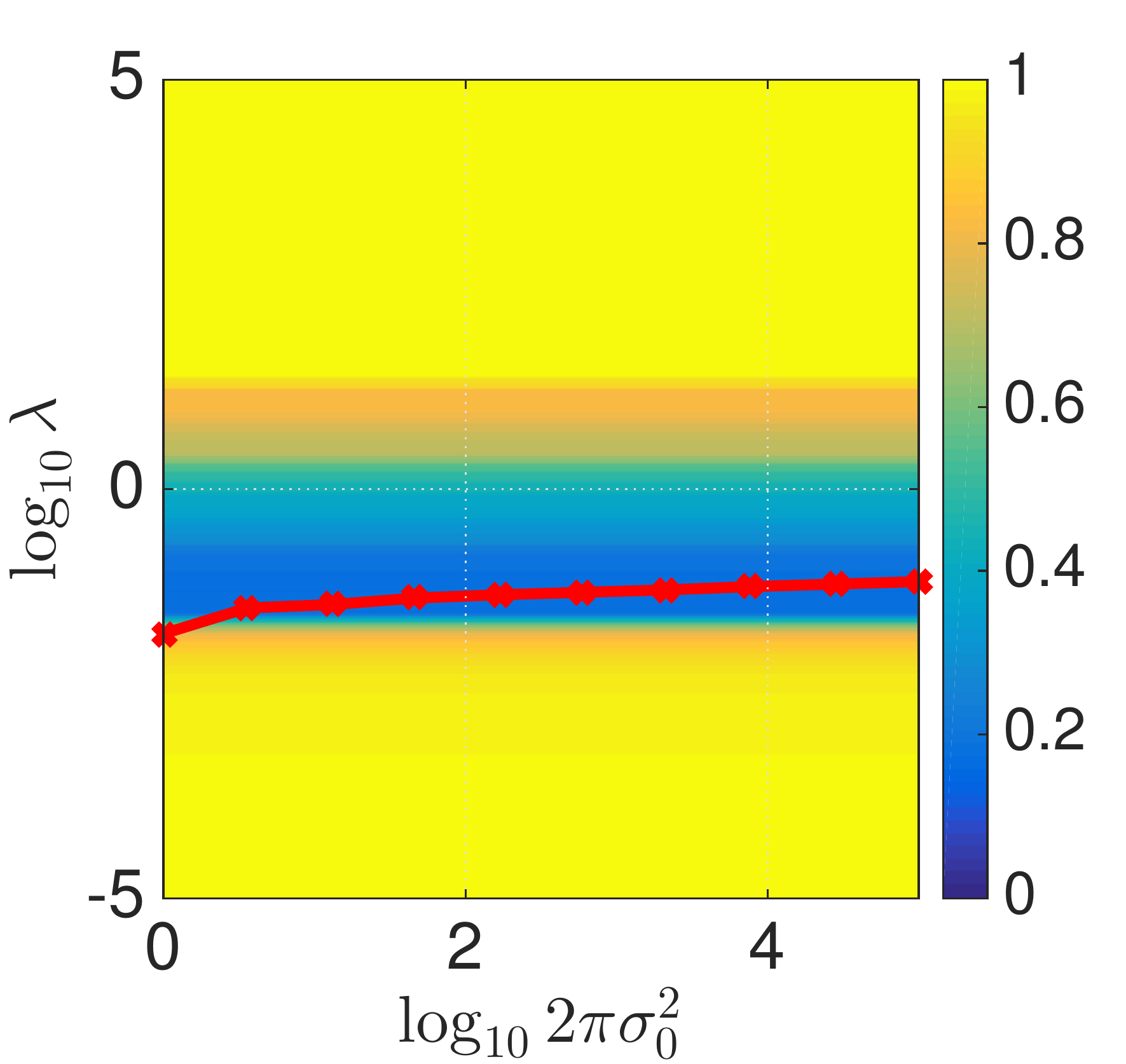}\\
\includegraphics[width=.162\linewidth,clip=true,trim=.3cm 0cm .5cm .5cm]{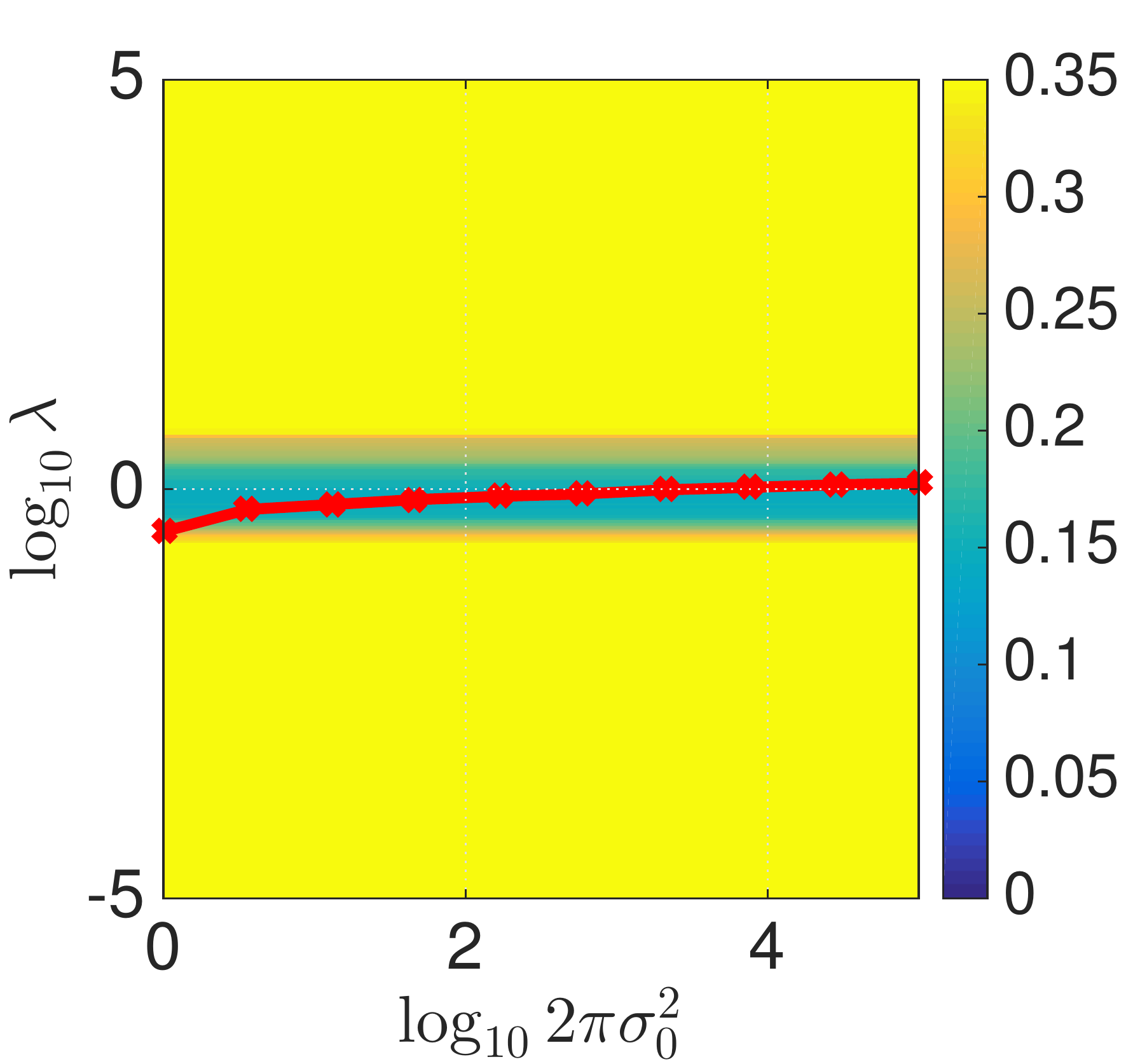}&
\includegraphics[width=.162\linewidth,clip=true,trim=.3cm 0cm .5cm .5cm]{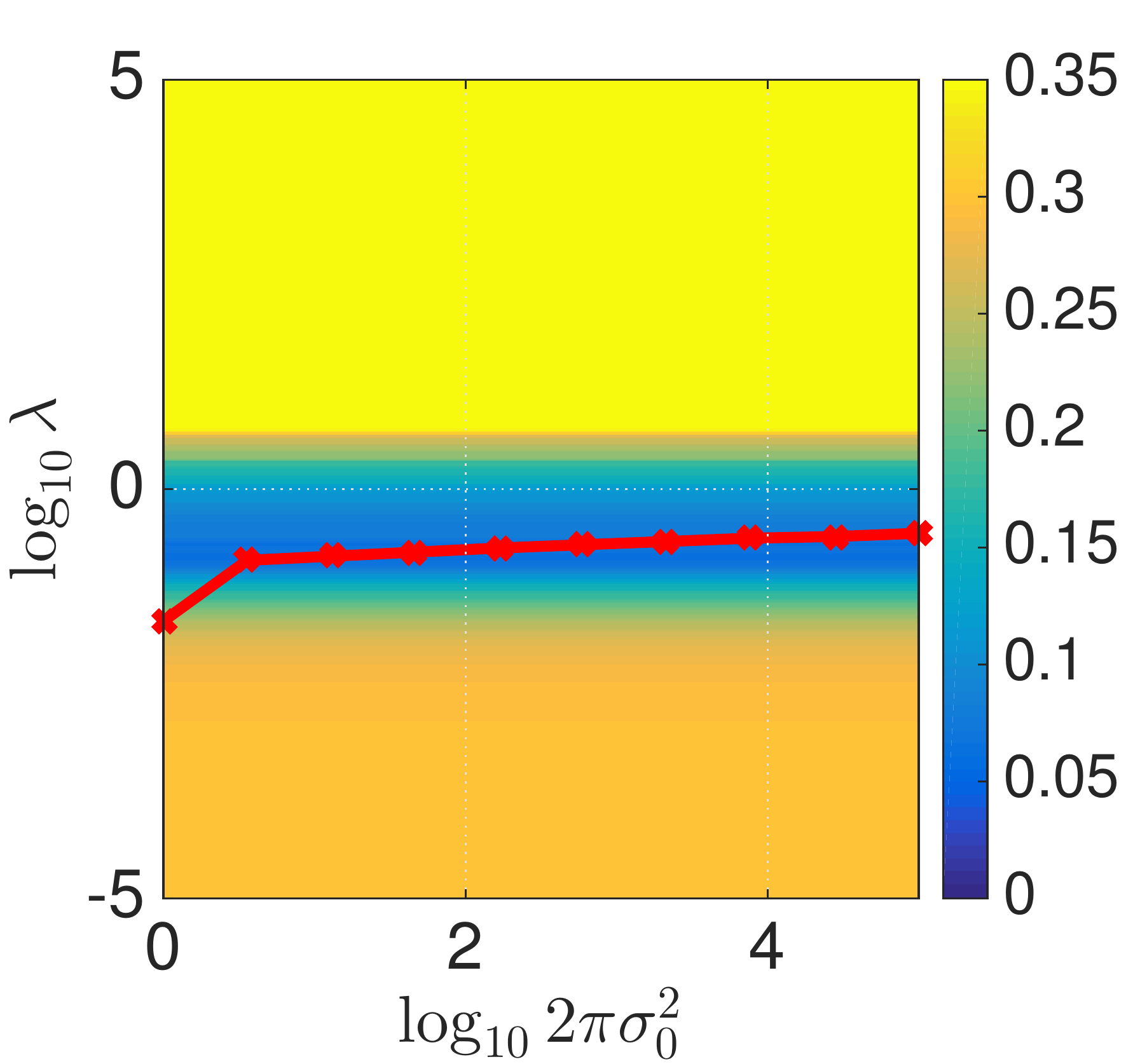}&
\includegraphics[width=.162\linewidth,clip=true,trim=.3cm 0cm .5cm .5cm]{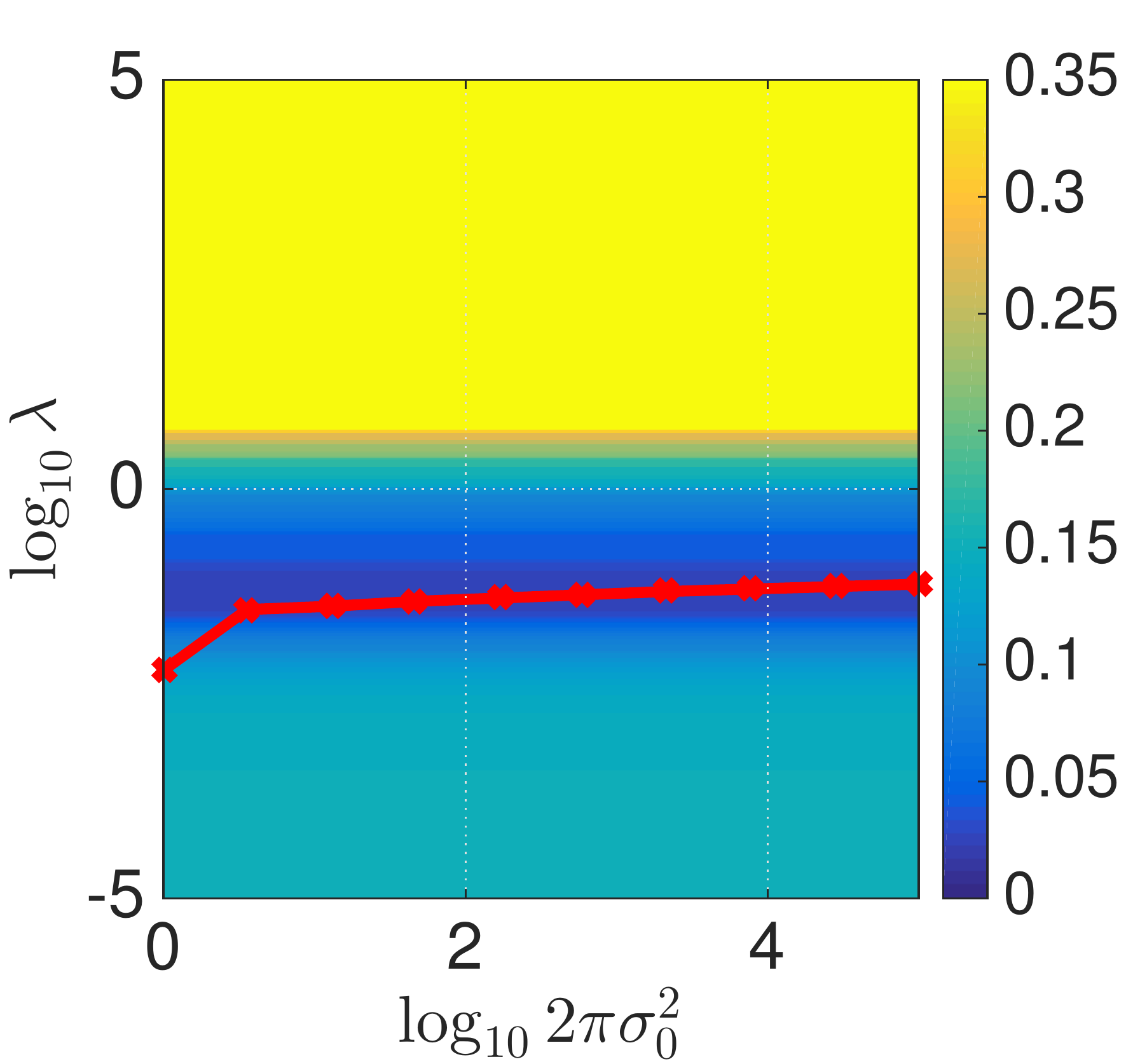}&
\includegraphics[width=.162\linewidth,clip=true,trim=.3cm 0cm .5cm .5cm]{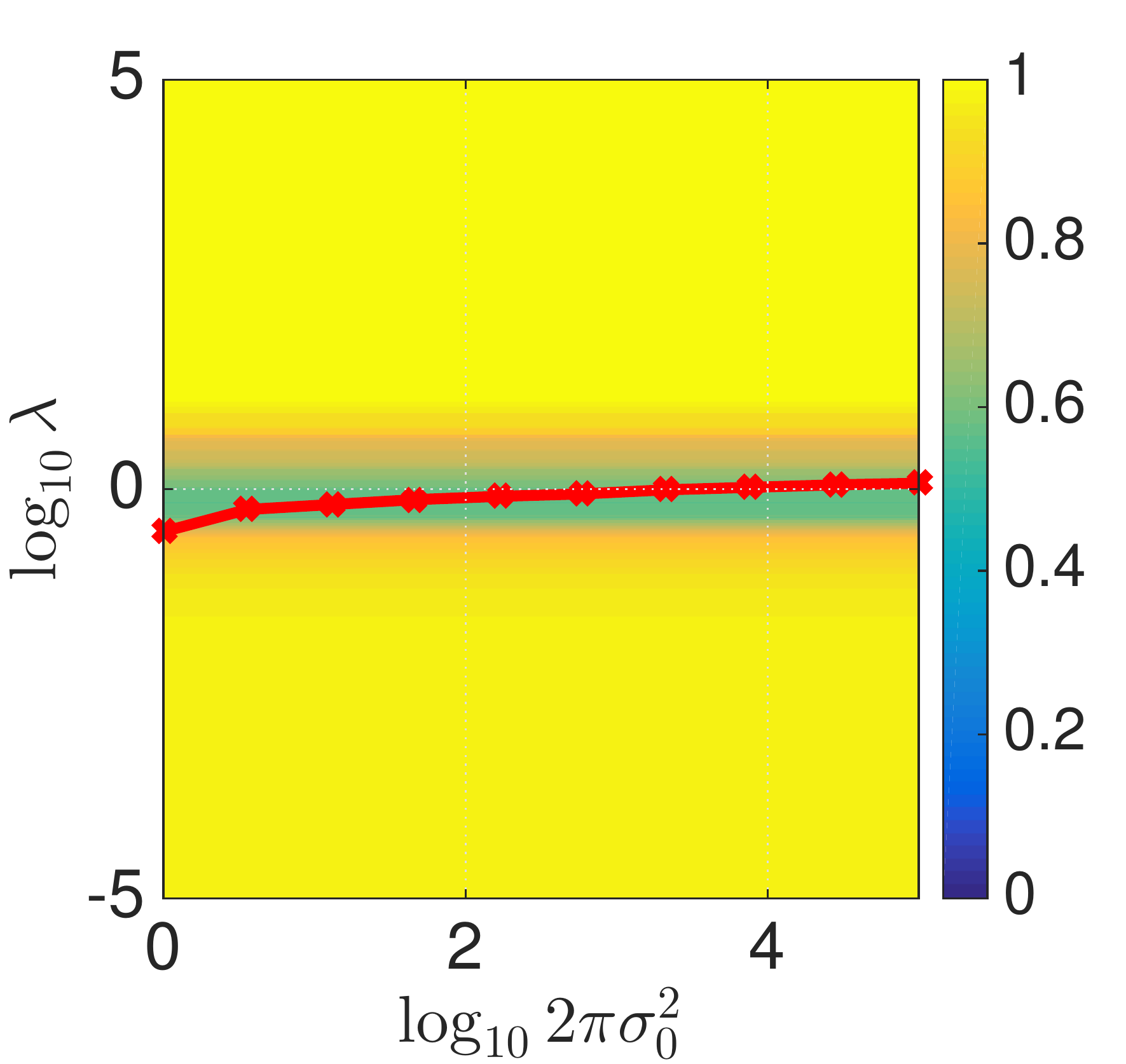}&
\includegraphics[width=.162\linewidth,clip=true,trim=.3cm 0cm .5cm .5cm]{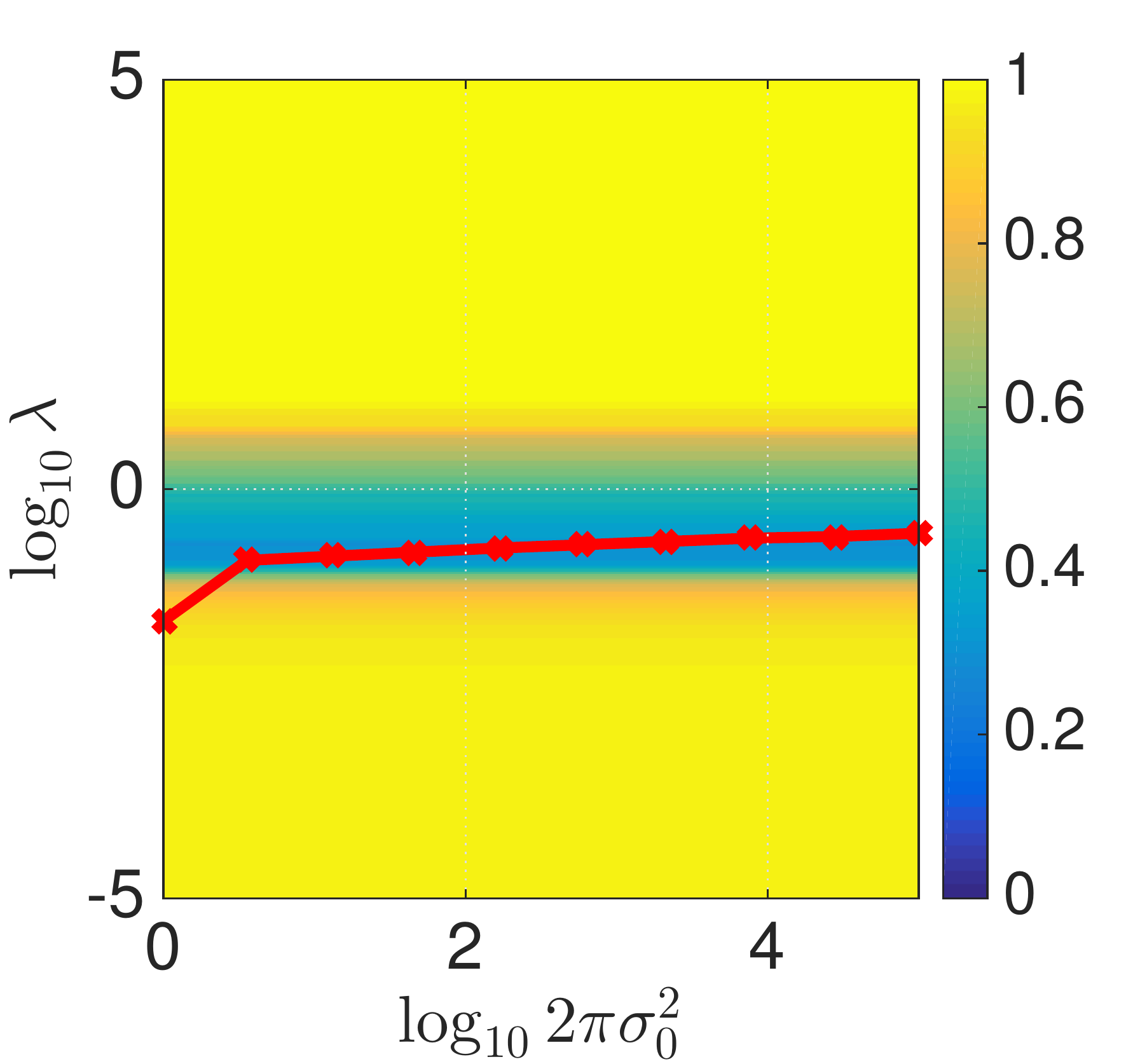}&
\includegraphics[width=.162\linewidth,clip=true,trim=.3cm 0cm .5cm .5cm]{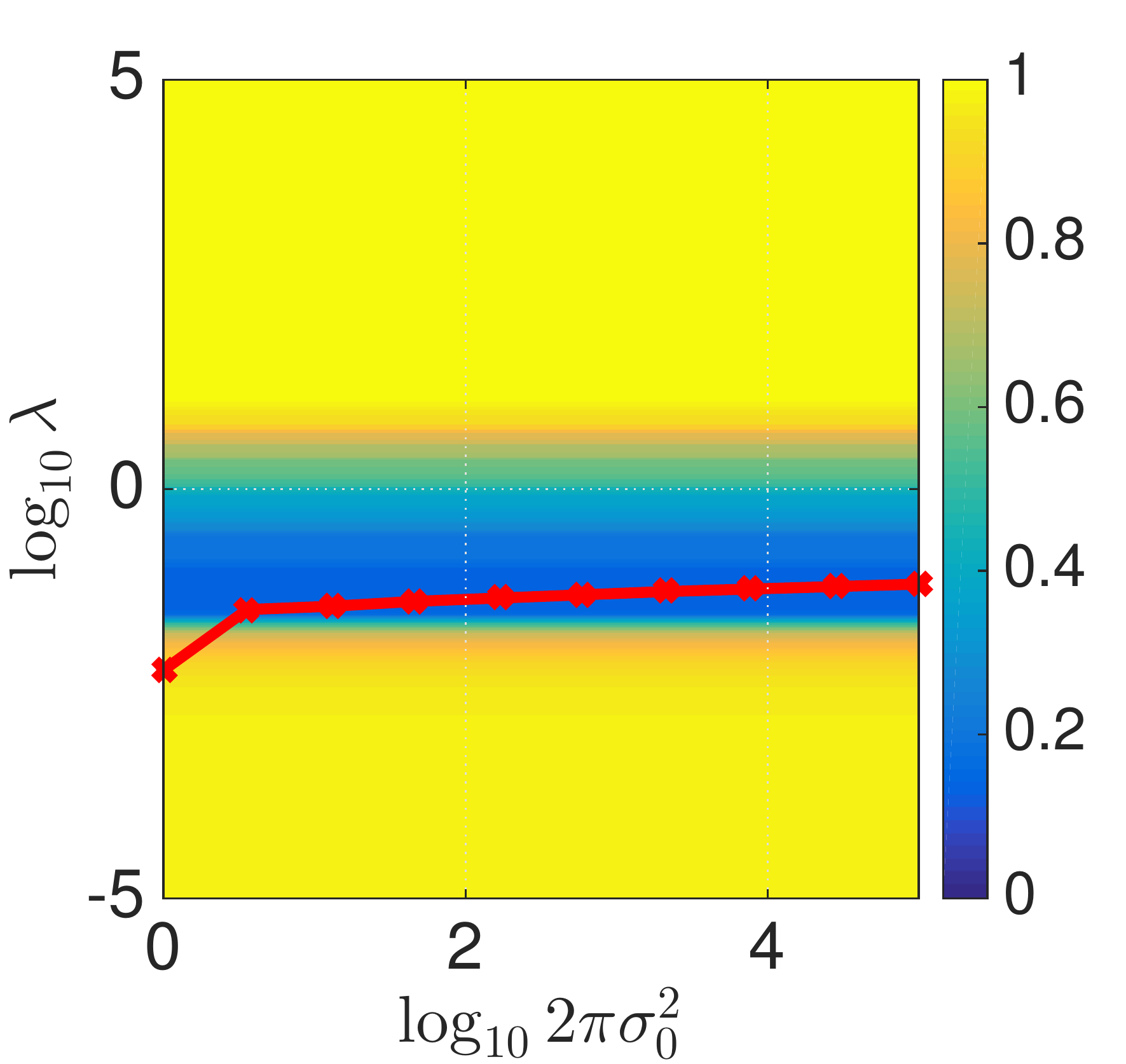}\\
\multicolumn{3}{c}{(a) relative MSE} &\multicolumn{3}{c}{(b) Jaccard error}
\end{tabular}
 \caption{\textbf{Estimation performance: RMSE and Jaccard error as functions of ${\lambda}$ and $\sigma_0^2$.}
The background displays the relative MSE (left) or Jaccard error (right) w.r.t. $\lambda$ and $\sigma_0^2$. We superimpose in red the estimate $\widehat{\lambda}$, which a priori explicitly depends on the choice of the hyperprior $\sigma_0^2$, is averaged over $50$ realizations and displayed in red as a function of $2\pi\sigma_0^2$.
 Choosing $\log 2 \pi\sigma_0^2 \in [0,5]$ leads to satisfactory estimation performance independently of $p$ and the ANR.
For each configuration $\overline{x}_{\max}-\overline{x}_{\min}=1$.
From top to bottom: $p=0.005$, $0.010$ and $0.015$.
From left to right: ${\rm ANR}=1$, $2$ and $4$.
\label{fig:sigma0}}
\end{figure*}

We finally investigate the impact of the choice of the hyperparameter $\sigma_0^2$ on the achieved solution, according to the discussion in Remark \ref{rem1}. Fig.~\ref{fig:sigma0dyn} displays $\widehat{\lambda}$ (red circle) as a function of $\sigma_0^2$, for different values of $\overline{x}_{\max} - \overline{x}_{\min}$.
It shows that using $\log 2 \pi\sigma_0^2 \in [0,5]$ systematically leads to satisfactory estimates that minimize the relative MSE (left) or Jaccard error (right).
This clearly indicates that $\sigma_0^2$ does not depend on data dynamics ($\overline{x}_{\max} - \overline{x}_{\min}$), which is what is expected from a hyperparameter.
Finally, to explore the potential dependencies on $p$ or ANR, we set $\overline{x}_{\max} - \overline{x}_{\min}=1$.
Fig.~\ref{fig:sigma0} shows that selecting any value of $\sigma_0$ such that $\log 2 \pi\sigma_0^2 \in [0,5]$ leads to satisfactory estimates minimizing the relative MSE (left) or Jaccard error (right), irrespectively of the actual values of $p$ or of the ANR.

\vskip\baselineskip

\section{Conclusions and perspectives}
\label{s:conc}
This contribution studied a change-point detection strategy based on the $\ell_2$-Potts model, whose performance depend crucially on the selection of a regularization parameter.
Using an equivalence between a variational formulation and a hierarchical Bayesian formulation of the change-point detection problem, the present contribution proposed and assessed an efficient automated selection of this regularization parameter.
It shows that estimation performance of the proposed procedure (evaluated in terms of global MSE  and Jaccard error) match satisfactorily those achieved with oracle solutions.
Moreover, when compared to fully Bayesian strategies, the proposed procedure achieved equivalent performance at significantly lower computational costs.
One of the advantages of the proposed approach is that it can be easily adapted to different additive noise degradations. For instance, for the $\ell_1$-Potts setup, that is the Laplacian noise assumption, the likelihood~\eqref{eq:likelihood} should be replaced by a Laplacian distribution with scale parameter $\sigma$ and the step \eqref{eq:sigmaest} should be replaced by $\widehat{\sigma}_\lambda = \|\boldsymbol{y}-\widehat{\boldsymbol{x}}_\lambda\|/N$ to be consistent. One difficulty that can be encountered for this kind of degradation is that we do not know the conjugate prior for the Laplace distribution. We have recently studied a related issue in a  conference paper \cite{Frecon_J_2017_p-ieee-icassp_bay_dca}. Future work could also aim to extend the present framework to Poisson noise.

\appendix

\section{Bayesian estimators\label{app:bayesianEstimator}}

The maximum a posteriori (MAP) or minimum mean squared error (MMSE) estimators associated with the joint posterior $f(\Vparam|\Vobs)$ in \eqref{eq:posteriorDistribution} can be approximated by using MCMC procedures that essentially rely on a partially collapsed Gibbs sampler  \cite{vanDyk2008} similar to the algorithm derived in \cite{Dobigeon_N_2007_j-ieee-tsp_joi_sma}. It consists in iteratively drawing samples (denoted $\cdot^{[t]}$) according to conditional posterior distributions that are associated with the joint posterior \eqref{eq:posteriorDistribution}. The resulting procedure, detailed in Algo.~\ref{algo:Gibbs}, provides a set of samples $ \boldsymbol{\vartheta} =\left\{\boldsymbol{r}^{[t]}, \boldsymbol{\mu}^{[t]},\sigma^{2[t]},\prob^{[t]}\right\}_{t=1}^{T_{\rm MC}}$ that are asymptotically distributed according to \eqref{eq:posteriorDistribution}. These samples can be used to approximate the MAP and MMSE estimators of the parameters of interest \cite{Marin2007}. The corresponding solutions are referred to as $\widehat{\boldsymbol{\xvar}}_{\rm MAP}$ and $\widehat{\boldsymbol{\xvar}}_{\rm MMSE}$ in Section~\ref{ss:bayes}.

\begin{algorithm}[H]
\caption{\label{algo:Gibbs} Piecewise constant Bayesian denoising}
\begin{algorithmic}[1]
\small
\REQUIRE Observed signal $\boldsymbol{y}\in\mathbb{R}^N$.\\
$\quad\;$ Hyperparameters $\Vhyper=\left\{\alpha_0,\alpha_1,\hmean,\hvar\right\}$.\\
\vspace{0.1cm}
\hspace{-0.6cm}\textbf{Iterations:}
\FOR{$t=1,\ldots,T_{\rm MC}$}
  \FOR{$\noobs=1,\ldots,\nbobs-1$}
	   \STATE Draw $\rup{\noobs}^{[t]} \sim f\left(\rup{\noobs}|\boldsymbol{y},\boldsymbol{r}_{\backslash \noobs},\prob,\sigma^2,\mu_0, \sigma^2_0\right)$
  \ENDFOR
  \FOR{$\noseg=1,\ldots,\sum_{i=1}^N r_i^{[t]}$}
      \STATE Draw $\mu_{\noseg}^{[t]} \sim f\left(\mu_k|\boldsymbol{y},\boldsymbol{r},\sigma^2,\mu_0,\sigma^2_0\right)$
  \ENDFOR
  \STATE Draw $\sigma^{2[t]} \sim f\left(\sigma^2|\boldsymbol{y},\boldsymbol{r},\boldsymbol{\mu}\right)$
	\STATE Draw $\prob^{[t]} \sim f\left(\prob|\boldsymbol{r},\alpha_0,\alpha_1\right)$
\ENDFOR
\vspace{0.1cm}
\ENSURE $ \boldsymbol{\vartheta}= \left\{\boldsymbol{r}^{[t]},\boldsymbol{\mu}^{[t]},\sigma^{2[t]},\prob^{[t]}\right\}_{t=1}^{T_{\rm MC}}$, $\widehat{\boldsymbol{\xvar}}_{\rm MAP}$ and $\widehat{\boldsymbol{\xvar}}_{\rm MMSE}$.
\end{algorithmic}
\end{algorithm}

\bibliographystyle{IEEEbib}
\bibliography{abbr,refs}

\end{document}

%% file: notations.tex
\def\bsL{{\boldsymbol{L}}}

\newcommand{\bTheta}{\boldsymbol{\Theta}}
\newcommand{\bPhi}{\boldsymbol{\Phi}}

\newcommand{\nbobs}{N}

\newcommand{\noobs}{i}
\newcommand{\noseg}{k}

\newcommand{\Vobs}{\mathbf{y}}

\newcommand{\rup}[1]{r_{#1}}

\newcommand{\prob}{p}

\newcommand{\hmean}{\mu_0}
\newcommand{\hvar}{\sigma^2_0}

\newcommand{\Vparam}{\bTheta}
\newcommand{\Vhyper}{\bPhi}